\documentclass{article}

\usepackage{microtype}
\usepackage{graphicx}
\usepackage{subfigure}
\usepackage{booktabs} 

\usepackage{hyperref}

\def\year{2021}\relax
\newif\ifisarxiv

\isarxivtrue

\ifisarxiv
\usepackage{fullpage}
\PassOptionsToPackage{numbers, compress}{natbib}
\else
\usepackage{icml2021}
\fi

\usepackage[utf8]{inputenc} 
\usepackage[T1]{fontenc}    
\usepackage{booktabs}       
\usepackage{amsfonts}       
\usepackage{nicefrac}       
\usepackage{microtype}      
\usepackage{soul}
\usepackage[utf8]{inputenc}
\usepackage{graphicx}
\usepackage{amsmath}
\usepackage{booktabs}
\usepackage{multirow}

\usepackage{microtype}
\usepackage{graphicx}
\usepackage{subfigure}
\usepackage{booktabs} 

\usepackage{amsmath,amsfonts,bm}

\usepackage{xcolor}
\usepackage{colortbl}

\def\eqref#1{equation~\ref{#1}}

\def\1{\bm{1}}

\DeclareMathAlphabet{\mathsfit}{\encodingdefault}{\sfdefault}{m}{sl}
\SetMathAlphabet{\mathsfit}{bold}{\encodingdefault}{\sfdefault}{bx}{n}

\def\sR{{\mathbb{R}}}

\usepackage{multirow}
\usepackage{paralist}

\usepackage{booktabs} 
\usepackage{multicol}
\usepackage{diagbox}

\usepackage{here}

\usepackage{amsmath,amssymb,amsfonts,amsbsy,amsfonts,latexsym}
\usepackage{multirow}
\usepackage{makecell}
\usepackage[labelfont=bf,textfont=it,belowskip=0pt,aboveskip=5pt,tableposition=top]{caption}
\usepackage{xcolor}
\usepackage{colortbl}

\definecolor{colorA}{RGB}{189,201,225}
\definecolor{colorB}{RGB}{103,169,207}
\definecolor{colorC}{RGB}{ 28,144,153}
\definecolor{colorD}{RGB}{  1,108, 89}

\newcolumntype{R}{>{\columncolor{gray!40}}r}
\newcolumntype{L}{>{\columncolor{gray!40}}l}
\newcolumntype{C}{>{\columncolor{gray!40}}c}

\usepackage{adjustbox}
\usepackage{amsthm}
\newtheorem{theorem}{Theorem}
\newtheorem{remark}{Remark}
\newtheorem{lemma}{Lemma}
\newtheorem{myDef}{Definition}

\usepackage{bbding}
\usepackage{tabularx,colortbl,xcolor}
\usepackage{multirow}
\usepackage[normalem]{ulem}
\useunder{\uline}{\ul}{}

\usepackage{enumitem}

\usepackage{xparse}

\DeclareGraphicsExtensions{.pdf,.png}
\usepackage{makecell}

\usepackage{longtable}
\usepackage{pgfplots}

\NewDocumentCommand{\var}{O{s} m O{}}{%
  \ensuremath{#1_{#2}^{#3}}
}
\usepackage{siunitx}

\newcommand{\commentout}[1]{}

\definecolor{light-gray}{gray}{0.80}

\newcommand\eref{Eq.~\ref}
\newcommand\fref{Fig.~\ref}
\newcommand\tref{Tab.~\ref}
\newcommand\sref{\textsection~\ref}

\newcommand\hc{ \rowcolor{orange!40}}

\def\0{{\bf 0}}

\usepackage{amsthm}

\usepackage{amsthm}

\newtheorem{mytheorem}{Theorem}
\newtheorem{assumption}[mytheorem]{Assumption}
\newtheorem{lemma}[mytheorem]{Lemma}
\newtheorem{remk}[mytheorem]{Remark}

\ifisarxiv
\usepackage{hyperref}
\else
\fi

\usepackage{chngcntr}
\usepackage{xspace}

\newcommand{\ssfl}{\textsc{SSFL}\xspace}

\ifisarxiv
\title{Improving Semi-supervised Federated Learning by Reducing the Gradient Diversity of Models}
\author{%
  Zhengming Zhang$^{1}$\footnote{equal contribution}~,Yaoqing Yang$^2$\samethanks, Zhewei Yao$^{2}$\samethanks~, Yujun Yan$^3$, \\ Joseph E. Gonzalez$^2$, Michael W. Mahoney$^{2,4}$ \\
  $^1$ Southeast University \\
  $^2$ University of California, Berkeley \\
  $^3$ University of Michigan, Ann Arbor\\
  $^4$ International Computer Science Institute\\
}
\date{}

\else
\icmltitlerunning{Improving Semi-supervised Federated Learning by Reducing the Gradient Diversity of Models}
\fi

\begin{document}

\ifisarxiv
\newcommand{\figsize}{0.42}
\newcommand{\figlen}{7.0}
\newcommand{\figwidth}{5.0}
\newcommand{\tabsize}{0.6}
\newcommand{\tabsizeThreeFour}{0.45}
\newcommand{\tabsizeFive}{0.5}
\fi

\newcommand*\samethanks[1][\value{footnote}]{\footnotemark[#1]}

\ifisarxiv
\title{Improving Semi-supervised Federated Learning by Reducing the Gradient Diversity of Models}
\maketitle

\fi

\vspace{-2mm}
\begin{abstract}
Federated learning (FL) is a promising way to use the computing power of mobile devices while maintaining the privacy of users. 
Current work in FL, however, makes the unrealistic assumption that the users have ground-truth labels on their devices, while also assuming that the server has neither data nor labels.
In this work, we consider the more realistic scenario where the users have only unlabeled data, while the server has some labeled data, and where the amount of labeled data is smaller than the amount of unlabeled data.
We call this learning problem semi-supervised federated learning (\ssfl). 
For \ssfl, we demonstrate that a critical issue that affects the test accuracy is the large \emph{gradient diversity} of the models from different users.
Based on this, we investigate several design choices.
First, we find that the so-called \emph{consistency regularization loss} (CRL), which is widely used in semi-supervised learning, performs reasonably well but has large gradient diversity.
Second, we find that Batch Normalization (BN) increases gradient diversity. Replacing BN with the recently-proposed Group Normalization (GN) can reduce gradient diversity and improve test accuracy. 
Third, we show that CRL combined with GN still has a large gradient diversity when the number of users is large. 
Based on these results, we propose a novel grouping-based model averaging method to replace the FedAvg averaging method.
Overall, our grouping-based averaging, combined with GN and CRL, achieves better test accuracy than not just a contemporary paper on \ssfl in the same settings (>10\%), but also four supervised FL algorithms.
\end{abstract}

\section{Introduction}
\label{sec:intro}

State-of-the-art machine learning models can benefit from the large amount of user data privately held on mobile devices, as well as the computing power locally available on these devices.
In response to this, federated learning (FL) has been proposed~\cite{konecny2016federated, Communication_Efficient_Learning}. 
In a typical FL pipeline, a server and some users jointly learn a model in multiple rounds. 
In each round, 
models are are updated locally (e.g., on users' devices) based on private user data, 
the server aggregates the updated models sent from the users, and 
the server then shares the aggregated model with the users for the next round.

In FL, it is commonly assumed that the data stored on the local devices are fully annotated with ground-truth labels, and that the server does not have any labeled data~\cite{konecny2016federated,Communication_Efficient_Learning,Federated_Optimization}. 
However, this assumption does not hold in practice. 
On the one hand, there is not a sufficient supply of labeled data on the users' side~\cite{FederatedSemisupervisedLearning}, as labeling data requires both time and domain knowledge~\cite{zhu2005semi,snow2008cheap}.
On the other hand, the server, which is often hosted by organizations, is more likely than a single user to acquire labeled data.
To give some concrete examples, consider two scenarios: cross-device FL (in which users are mobile devices) and cross-silo FL (in which users are organizations) \cite{kairouz2019advances}.
In a cross-device scenario, where a central server trains an object detector on images with the help of mobile users, the server can use a public dataset, e.g., \cite{lin2014microsoft}, to obtain labels, while the users often do not have images with ground-truth bounding boxes.
In a cross-silo scenario, where multiple medical institutes work together to diagnose a disease, the disease may be newly discovered by one medical institute, and so no labeled samples are present at other institutes~\cite{FederatedSemisupervisedLearning}.%
\footnote{There is some nuance here that the medical institute with ground-truth labels is physically different from the server, and the server itself does not have data. 
However, it will become apparent that this nuance does not affect the mathematical formulation considered in our paper because we can assume (virtually) that the server and the institute with labels are co-located and work together as a new server.} 
In these scenarios, the typical supervised FL setting is not appropriate. 

Motivated by these practical scenarios, we study the \emph{semi-supervised federated learning} (\ssfl) setting.
In \ssfl, users only have access to unlabeled data, while the server only has a small amount of labeled data.%
\footnote{In addition to our main setup, when users have unlabeled data only, we also compare our method to the state-of-the-art in another ``label-at-client'' scenario \cite{jeong2020federated}, when users have a limited amount of labeled data. Our method outperforms \cite{jeong2020federated} in this new scenario by a large margin (>10\%).} 
The goal is to train a model that can benefit from both labeled and unlabeled data.
In this context, our main contributions are the following.
\ifisarxiv
\else
\fi
\begin{enumerate}[noitemsep, nolistsep, labelindent=0pt, leftmargin=*]
\item \textbf{Demonstrating the importance of ``gradient diversity.''} 
We demonstrate the importance of reducing the gradient diversity \cite{Gradient_Diversity}, a notion which captures the dissimilarity between local gradient updates of users, in \ssfl. 
First, we show that the consistency regularization loss (CRL)~\cite{Fixmatch} can achieve reasonably good test accuracy, but it still has significantly larger gradient diversity than supervised FL. 
Then, we show that replacing the batch normalization (BN)~\cite{BN} in the model with group normalization (GN)~\cite{GroupNorm} can reduce gradient diversity and enhance test accuracy in the \ssfl setting. 
Finally, we propose a grouping-based model averaging technique to replace FedAvg \cite{Communication_Efficient_Learning}, to reduce gradient diversity further and to increase accuracy, especially when there are a large number of users.

\item \textbf{Proposing a strong baseline.} 
By proposing solutions to reduce gradient diversity, we obtain a strong \ssfl approach. 
Our method outperforms another \ssfl approach from a contemporary paper~\cite{jeong2020federated} in the same settings by 14.79\%-18.10\% in test accuracy.
Our method also achieves comparable or better accuracy than four existing supervised FL approaches that do not use GN or the grouping-based averaging. 
Specifically, our approach is 0.80\%/0.29\% better than EASGD/OverlapSGD~\cite{EASGD,Overlap}, despite having a lower communication frequency, and our approach is 14.44\%/11.14\% better than FedAvg/DataSharing~\cite{Communication_Efficient_Learning,FedNoniidData}, even when the degree of our non-iidness (in the sense of different distributions of classes at different users) is higher.

\item \textbf{Extensive empirical evaluation.}
We evaluate the proposed solution by varying different environmental factors and testing on multiple datasets. The environmental factors include different levels of non-iidness, the \emph{communication period} (i.e., the number of local update steps at each user between two communication rounds), the total amount of labeled data in the server, the number of users, 
and the number of users that communicate with the server in each communication round.
Interestingly, the problem of having large gradient diversity when the number of communicating users is large is discovered in one of these empirical evaluations (see Section \ref{subs:FixCk}).
\end{enumerate}
Overall, by formulating the \ssfl problem, analyzing the key limitation of large gradient diversity, selecting different design choices to reduce the gradient diversity, and thoroughly evaluating our design under different environmental factors, we provide a strong baseline for this \ssfl setting.
This strong baseline can achieve comparable or better accuracy than the state-of-the-art methods in both semi-supervised and supervised FL. 
The proposed method also only focuses on a few crucial components (e.g., normalization) that are easy to change in practice. 
To help the FL community reproduce our results, we have open-sourced our code.\footnote{https://github.com/jhcknzzm/SSFL-Benchmarking-Semi-supervised-Federated-Learning}

\section{Semi-supervised federated learning}
\label{sec:method}

\subsection{Basic setup}
\label{sec:ssfl_framework}
 
\begin{figure*}
\centering
\includegraphics[width=0.95\linewidth]{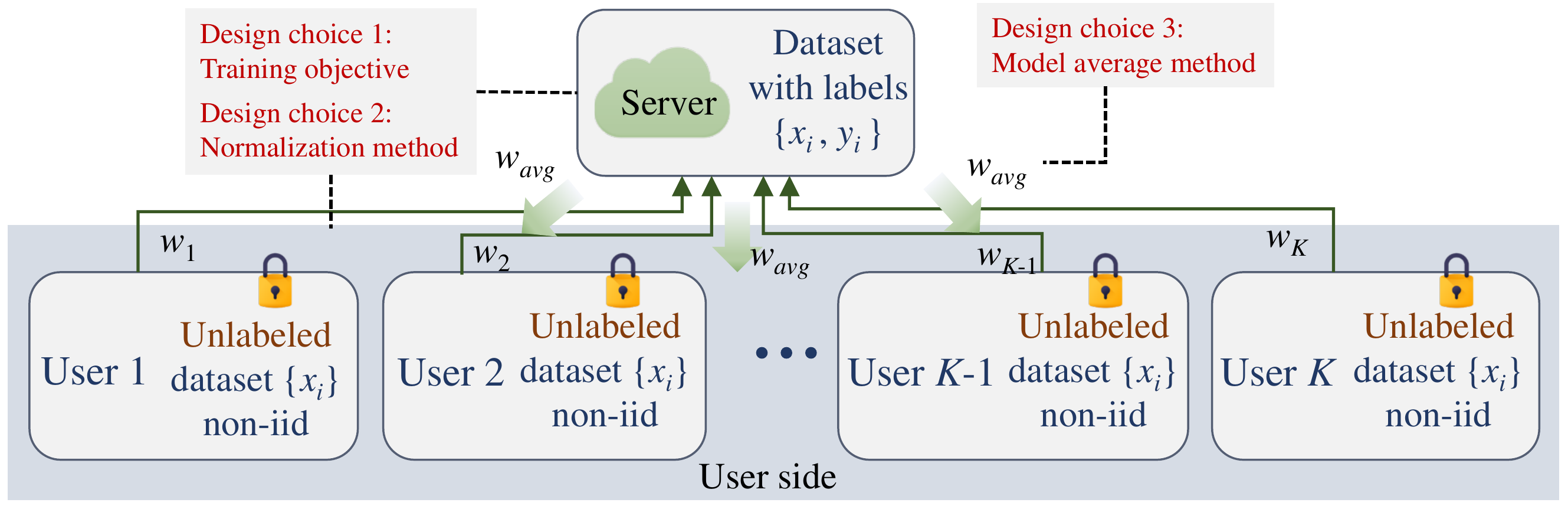}
\caption{\footnotesize 
Semi-supervised federated learning (\ssfl). 
Only the server has access to labeled data, i.e., the data stored in local users are unlabeled. 
Furthermore, the data distributions across different users are non-iid. 
}
\label{system_model}
\end{figure*}

In this subsection, we discuss the basic setup of \ssfl. There exist a cloud server and $K$ users/devices. Similar to the common FL setup~\cite{konecny2016federated}, the users and the server collaborate to train a model in multiple rounds by exchanging and updating model weights. 
For each round of communication, we allow the number of participating users connected to the server, which we denote as $C$, to be smaller than $K$, as is done commonly \cite{kairouz2019advances}. 
This is because, for example, some mobile devices only participate in the learning when being charged \cite{kairouz2019advances}.
Assuming $C\leq K$ for each round of communication can simulate this drop-and-reconnect case.

We denote the labeled dataset at the server as ${D_s}{\rm{ = }}\left\{ {\left( {{x_i},{y_i}} \right)} \right\}_{i = 1}^{{N_s}}$, and the unlabeled dataset stored at the $k$-th user as ${D_k}{\rm{ = }}\left\{ {{x_i}} \right\}_{i=1}^{{N_k}}$, for $k\in \{1,\ldots,K$\}. 
Here, $N_s$ ($N_k$) is the number of labeled (unlabeled) samples available at the server ($k$-th user). Also, similar to the standard FL setup, no raw data are exchanged between the server and the users. 
That is to say, the server can only use the dataset $D_s$, and the $k$-th user can only use the local dataset $D_k$. 
Note that the data distributions at different users are non-iid \cite{FedNoniidData,hsu2019measuring}. In this work, we consider image classification as a representative \ssfl task. 

We now describe the \ssfl training pipeline which can be slightly different from the standard FL setup.
Denote the local weights at the $k$-th user as $w_k$. 
Since the server has its own dataset $D_s$, unlike the standard FL setup, it also updates its own weights $w_s$. 
Denote the averaged weights at the server as $w_{avg}$ (which is different from $w_s$).
At round $t$, the server sends the averaged model weights $w_{avg}^t$ to the users.
Each user, upon receiving $w_{avg}^t$, locally updates its own model weights to $w_k^t$ and transmits $w_k^t$ to the server. 
At the same time, the server also has to update its own model weights from $w_{avg}^t$ to $w_s^t$ using the labeled dataset $D_s$.
Then, the server computes an averaged model $w_{avg}^{t+1}$ using all the received models, including its own model $w_s^t$. 
Finally, it proceeds to the next round and sends $w_{avg}^{t+1}$ to the users. 
Our basic \ssfl setup is illustrated in~\fref{system_model}. 

\subsection{Gradient diversity, and the ways to reduce it}
\label{Analysismetrics}

In this subsection, we present the definition of gradient diversity from \cite{Gradient_Diversity}. Then, we motivate several design choices to reduce gradient diversity in \ssfl. As we have discussed in the introduction, reducing the gradient diversity value is crucial for \ssfl.

\begin{myDef}[Metric for gradient diversity]
\label{def:weight_diversity}
The gradient diversity is defined as:
\begin{equation}
\small 
\Delta^t (w) = 
\sum\nolimits_{k \in \mathcal{C}_t} {\left\| \nabla w_k^t \right\|_2^2}/\left\| \sum\nolimits_{k \in \mathcal{C}_t} \nabla w_k^t  \right\|_2^2,
\end{equation}
where $\mathcal{C}_t$ denotes the set of participating users at round $t$, $w_k^t$ represents the model weights held by the $k$-th user at the beginning of round $t$,
and $\nabla w_k^t$ represents the gradient of $w_k^t$ evaluated on all data held by the $k$-th~user.
\end{myDef}

Gradient diversity measures the dissimilarity between the local gradient updates of users. 
In \ssfl, when gradient diversity is too large, the weights from different users are updated towards ``different directions,''
and it is thus problematic to directly average them, as is done in the common model averaging method known as FedAvg~\cite{Communication_Efficient_Learning}.
Gradient diversity gives a quantitative way to study this issue. 
We only include the users in $\mathcal{C}_t$, i.e., those who participate in this particular communication round, because only these weights are~averaged.

In what follows, we motivate three design choices, which are also shown in ~\fref{system_model}, that can affect gradient~diversity.

\begin{enumerate}[noitemsep, nolistsep, labelindent=0pt, leftmargin=*]
\item
{\bf{Training objective.}} 
Since there are no labels on the users' side, we have to choose appropriate loss functions carefully when updating local models at the users' side. 
\item
{\bf{Normalization.}}
Normalization (e.g., BN) has become standard in deep neural network models.
The \ssfl setting requires a careful choice of specific normalization methods.
\item
{\bf{Model averaging.}}
The way the server computes the aggregated model from the models that it receives is also a design choice.
We only consider ways to average the model weights, i.e., we do not consider model ensembling or distillation techniques, which can be time-consuming in multiple rounds \cite{OneShotFedLearning}.
\end{enumerate}

\subsection{Environmental factors for evaluation}\label{sec:other_factors}

In this subsection, we list some environmental factors that can affect the test accuracy of \ssfl algorithms. 
These factors are not controllable by the designer, and they are independent of the design choices listed in \sref{Analysismetrics}. 
However, these factors are helpful to evaluate different design choices, and they can potentially display the weakness of certain solutions.
The following factors are considered. 
\begin{enumerate}[noitemsep, nolistsep, labelindent=0pt, leftmargin=*]
\item 
{\bf{Non-iidness $R$}}: the non-iid metric of data distributions; see~Definition~\ref{def:R}.
\item 
{\bf{Communication period $T$}}: during two consecutive communications, the number of gradient update steps locally done by the users and the server.
\item 
{\bf{Server data number $N_s$}}: the number of labeled data in the server.
\item 
{\bf{User number $K$}}: the total number of users.
\item 
{\bf{Number of participating users $C$}}: during each communication round, the group of users who send their models to the server. 
\end{enumerate}

\textbf{Varying non-iidness to evaluate our solution.} 
Among the five environmental factors listed above, evaluating with different non-iidness requires special care, because we have to change the dataset to get different degrees of non-iidness.
Here, we follow convention and evaluate using synthesized non-iid datasets that have different \emph{class distribution skews} \cite{Overlap,FedNoniidData,hsu2019measuring,jeong2018communication,sattler2019robust}, e.g., a single user can have more data for one class or a couple of classes than others. 

To quantify the class distribution skew in our experiments, we use the average total variation distance in Definition \ref{def:R}. 
In the definition, the \emph{empirical} class distribution of the data $D_k$ at the $k$-th user is denoted by $P_k\in \sR^{d}$, where $d$ is the number of classes. Clearly, $\sum_{j=1}^dP_k[j]=1$, for all $1\leq k\leq K$. Recall that $K$ is the number of users/devices.

\begin{myDef}[Metric $R$ for non-iid level]
\label{def:R}
The non-iid metric $R$ to measure the class distribution skew is defined~as:
\begin{equation}
\small 
R = \frac{1}{K(K-1)/2}\sum\nolimits_{1\leq k < m \leq K} \|P_k - P_m\|_1 / 2,
\end{equation}
where $\|\cdot\|_1$ is the $L_1$ norm.
\end{myDef}
Here $\|P_k - P_m\|_1/2$ is the (normalized) total variation distance, which takes value in $[0,1]$, and $K(K-1)/2$ is the number of user pairs, i.e., it is the mean total variation distance averaged over pairs of users.
In particular, $0\leq R\leq 1$~\cite{basseville1989distance}.
When data are distributed in such a way that each user has a the same empirical class distribution which is uniform $P_k=[1/d,...,1/d], \forall k$, we have $R=0$; and in another extreme, when $K=d$ and each user only has samples from one class, we have $R=1$. 
\ifisarxiv
\begin{remk}[Different data sizes]\label{rem:data_size}
The metric $R$ in Definition \ref{def:R} does not explicitly consider the effect of different data sizes $N_k$'s at different users. 
We focus on the case when $N_k$'s are equal to each other, while slight difference may arise when the overall number of samples is not divisible by the number of users.
Notice that, we consider similar data sizes at different users, but we do not restrict ourselves to the case of uniform class distribution. Specifically, we have tested on datasets with non-uniform class sizes, e.g., SVHN dataset.
\end{remk}
\fi
We synthesize datasets with a specific $R$ value in $[0,1]$ to evaluate our \ssfl algorithm. The specific data synthesis and distribution procedures to achieve $R$ are relegated to~\sref{subsec:DataDistribute}.

\section{Reduce gradient diversity in \ssfl}
In this section, we study the three \ssfl design choices discussed in \sref{Analysismetrics}. 
We first present the details of the choices considered in this paper. 
Then, we use gradient diversity to analyze them.
\subsection{Design choice 1: training objective}\label{sec:ExperimentalevaluationoftheCRLtrainingobjective}

In this subsection, we present the training objective and focus on an existing semi-supervised loss called consistency regularization loss (CRL) \cite{Fixmatch}. 
In particular, the server loss $L_s$ and the user loss $L_k$ (of the $k$-th user) are defined as follows:
\begin{align}
L_s &= \frac{1}{N_s}\sum\nolimits_{(x_i, y_i)\in D_s} {l\left( {{y_i},{f_s}(\alpha ({x_i});{w_s})} \right)}, \label{eq:serverloss}\\
L_k &= \frac{1}{N_k}\sum\limits_{x_i\in D_k} \mathbf{1}_{ {\max ({{\bar y}_i}) \ge \tau }} l\left( {\arg\max(\overline y_i ),{f_k}(A({x_i});{w_k})} \right), \label{eq:Fixmatchuserloss}
\end{align}
where (1) $D_s$ is the set of $N_s$ labeled samples, (2) $D_k$ is the set of $N_k$ unlabeled samples owned by the $k$-th user, (3) $w_s$ ($w_k$) are the weights of server model $f_s$ ($k$-th user model $f_k$), (4) $l(\cdot, \cdot)$ is the cross-entropy loss, $\alpha(\cdot)$ and $A(\cdot)$ are two data augmentation functions which we will soon describe in Remark \ref{rem:data_augmentation}, (5) ${{\bar y}_i} = f_k(\alpha(x_i);w_k)$ is the prediction of the model $f_k$ on the augmented sample $\alpha(x_i)$, (6) $\1$ is the indicator function, (7) and $\tau$ is the threshold hyperparameter which helps decide which samples have high confidence to be trained, i.e., the term $\mathbf{1}_{\max (\bar y_i) \ge \tau}$. We refer to training with \eref{eq:serverloss} and \eref{eq:Fixmatchuserloss} as the \emph{CRL training objective}. 

\begin{remark}[Data augmentation]\label{rem:data_augmentation}

We now discuss the data augmentations in \eref{eq:serverloss} and \eref{eq:Fixmatchuserloss}.
In~\cite{Fixmatch}, the authors use two different types of data augmentations (DA): the standard flip-and-shift augmentation $\alpha(\cdot)$ (referred to as \emph{weak DA}); and the RandAugment \cite{RandAugment} ${A}(\cdot)$ (referred to as \emph{strong DA}). 
Here, the latter RandAugment uses two different augmentation methods (i.e., shift and crop) out of twelve possible augmentation methods (e.g., rotate, shift, solarize, etc.) for one image.
We refer the interested readers to~\cite{RandAugment} for a detailed explanation.
The key idea behind using two DAs (i.e., weak DA and strong DA) is that the predictions of the same image with two data augmentations should be similar to each other. 
Recall that, on the user side, the data have no labels.
Therefore, using this approach, we can use the pseudo-labels generated from weak DA samples to supervise strong DA samples, which is the loss between $\arg\max(\overline y_i )$ and ${f_k}(A({x_i});{w_k})$ in \eref{eq:Fixmatchuserloss}.
This is shown in \cite{Fixmatch} to boost the testing performance.
\end{remark}

\textbf{Other training objectives.}
To study the CRL training objective, we compare it to two other training objectives. One uses classical self-training similar to the way defined in~\cite{Selftraining}, which is also called ``pseudo-labeling'' in ~\cite{Fixmatch}:
\begin{equation}
\small 
\label{eq:userloss}
\begin{aligned}
L_k = \frac{1}{N_k}\sum\limits_{x_i\in D_k} & {\bf{1}}_{\max ({{\bar y}_i}) \ge \tau }
l\left( { \arg\max(\bar y_i),{f_k}(\alpha ({x_i});{w_k})} \right).
\end{aligned}
\end{equation}
This loss can be explained as replacing two augmentations $\alpha(\cdot)$ and $A(\cdot)$ in the CRL training objective \eref{eq:Fixmatchuserloss} with a single standard flip-and-shift augmentation $\alpha(\cdot)$. 
It is called self-training because the pseudo-labels obtained by applying $\arg\max$ to the model's output ${{\bar y}_i} = f_k(\alpha(x_i);w_k)$ are used to supervise the model's output $f_k(\alpha(x_i);w_k)$ itself.
We refer to~\eref{eq:userloss} as the \emph{self-training objective}.

The other training objective assumes that the users have (oracle) ground-truth labels, and it uses standard empirical risk minimization for the user loss, e.g., used in~\cite{Communication_Efficient_Learning}:
\begin{equation}
\small 
\label{eq:supervisedloss}
\begin{aligned}
L_k &= \frac{1}{N_k}\sum\nolimits_{x_i\in D_k} l\left( y_i,f_k(\alpha (x_i);w_k) \right),
\end{aligned}
\end{equation}
where $y_i$ is the (oracle) ground-truth label of $x_i$. We refer to \eref{eq:supervisedloss} as the \emph{supervised training objective}.

\subsection{Design choice 2: normalization method}

In this subsection, we describe the next design choice regarding the normalization method. 
Recent works~\cite{adafed,NonIIDDataQuagmire} find that in supervised FL with non-iid data distributions, the performance of group normalization (GN) is usually much better than that of batch normalization (BN). 
In contrast to BN, which normalizes the feature maps over the batch, height, and width dimensions, GN normalizes the feature maps over the channel, height, and width dimensions. 
We conjecture that the improvement of applying GN in FL is due to the reduced gradient diversity, and we thus empirically evaluate the effects of these two different normalization~methods.

\subsection{Design choice 3: model averaging}

In this subsection, we study model averaging methods. We focus on a novel grouping-based averaging method.
The main idea is to divide the $C$ communication users in each round into $S>1$ groups and then perform the average group-wise.
Specifically, after collecting all $C$ model weights from the communication users, 
the server randomly divides them into $S$ equal-sized groups $\{G_i^t\}_{i=1}^{S}$, and updates the averaged weights according~to:
\begin{equation}
\small 
\label{eq:hierarchicalavg}
\left\{ \begin{array}{l}
w_{avg,i}^{t+1} = \left( w_s^t + \sum\nolimits_{k\in {G_i^t}} w_k^t  \right)/(|G_i^t| + 1),~\forall i\in \{1\ldots S\} \\
w_{avg}^{t+1} = \sum\nolimits_{i = 1}^S w_{avg,i}^{t+1} /S.
\end{array} \right.
\end{equation}
In the equation above, $w_{avg,i}^{t+1}$ represents the averaged weights in each group, and $w_{avg}^{t+1}$ is the average of these averaged weights. 
After computing $w_{avg,i}^{t+1}$ and $w_{avg}^{t+1}$, the server broadcasts $w_{avg,i}^{t+1}$ to the user group $G_i^t$, and it uses $w_{avg}^{t+1}$ for the training (updates) done by the server itself on the labeled data. 
It is worth noting that the groups $\{G_i^t\}_{i=1}^{S}$ change with $t$ because the set of participating users $\mathcal{C}_t$ change with time. We compare the grouping-based averaging method to FedAvg:
\begin{equation}
\small 
\label{eq:fedavg_def}
w_{avg}^{t+1} \stackrel{FedAvg}{=} \left(w_s^t + \sum\nolimits_{k\in \mathcal{C}_t} w_k^t\right)/(C+1),
\end{equation}
where $\mathcal{C}_t$ denotes the set of participating users with size $C$ in each round.

\subsection{Comparing different \ssfl methods}

In this subsection, we study different design choices by studying five different methods:
\begin{itemize}[noitemsep, nolistsep, labelindent=0pt, leftmargin=*]
    \item \emph{CRL with BN} uses \eref{eq:serverloss} and \eref{eq:Fixmatchuserloss} as the training objective. It uses BN as the normalization method and FedAvg in \eref{eq:fedavg_def} as the model averaging method.
    \item \emph{Self-training} uses \eref{eq:userloss} as the training objective. It also uses BN and FedAvg.
    \item \emph{Supervised training} uses \eref{eq:supervisedloss} as the training objective. It also uses BN and FedAvg.
    \item \emph{CRL with GN} uses \eref{eq:serverloss} and \eref{eq:Fixmatchuserloss} as the training objective. It replaces BN with GN, and it uses FedAvg.
    \item \emph{Grouping-based} uses the same CRL training objective and GN, as in CRL with GN, but it uses the grouping-based averaging method in \eref{eq:hierarchicalavg} instead of FedAvg.
\end{itemize}

We compare CRL with BN to self-training and supervised training to show where the CRL training objective stands compared to both semi-supervised and supervised algorithms.
We compare CRL with BN to CRL with GN to show which normalization method is better.
Further, we compare CRL with GN with the grouping-based method to show which model averaging method is better.
The grouping-based solution combines CRL, GN, and our grouping-based averaging method. This solution is our main algorithm.

We use ResNet-18~\cite{ResNet} on Cifar-10. Here, for the environmental factors in~\sref{sec:other_factors}, we set $T = 16$, $K = 10$, $C = 10$, and $N_s = 1000$. We compare under two $R$ values, with $R=0.4$ referred to as the non-iid case, and $R=0$ referred to as the iid case.
The threshold $\tau$ used in~\eref{eq:serverloss} and~\eref{eq:Fixmatchuserloss} is chosen to be $0.95$, the same as in~\cite{Fixmatch}. 

\begin{figure}
    \centering
    \includegraphics[width=\figsize\linewidth]{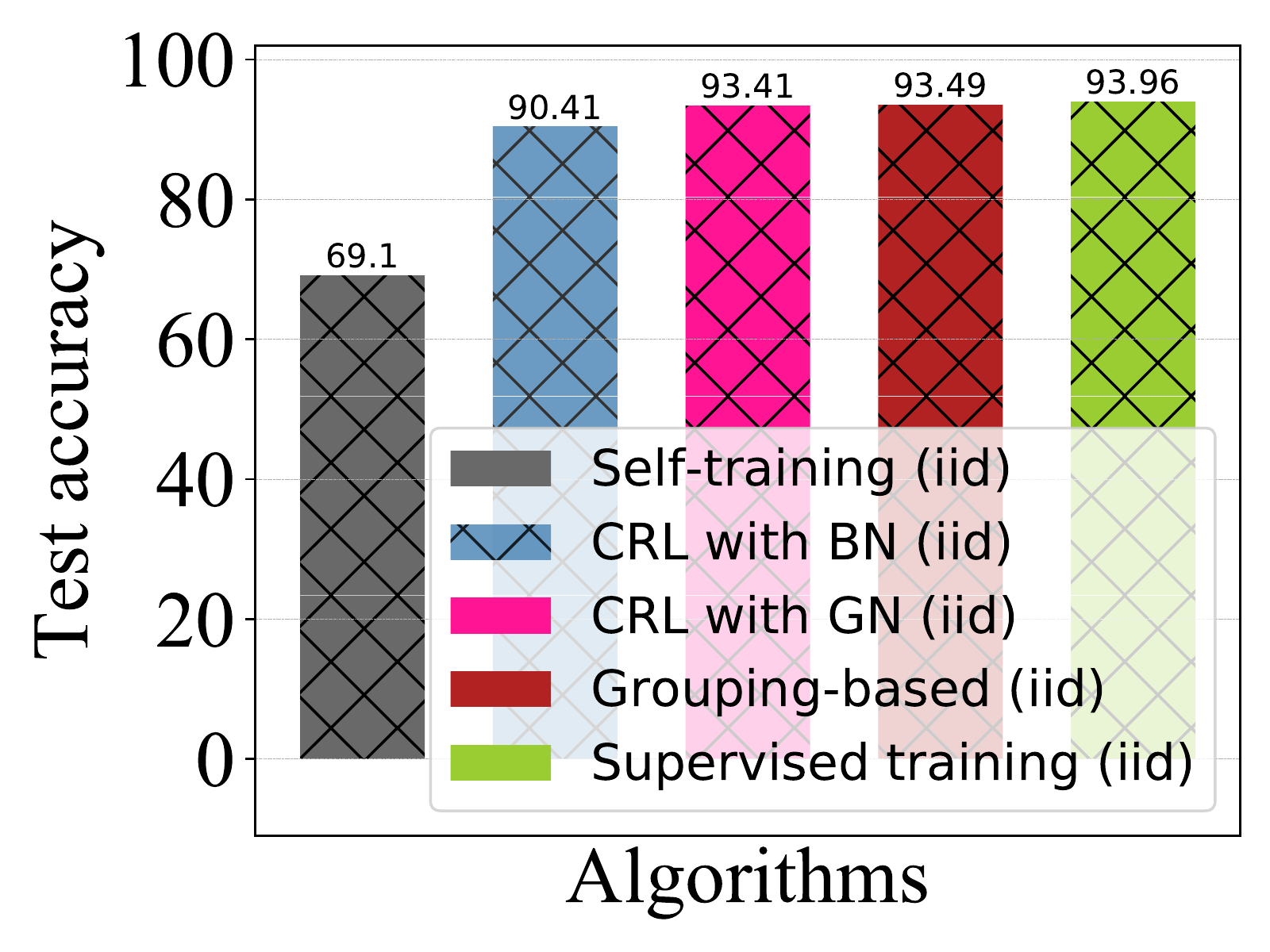}
    \includegraphics[width=\figsize\linewidth]{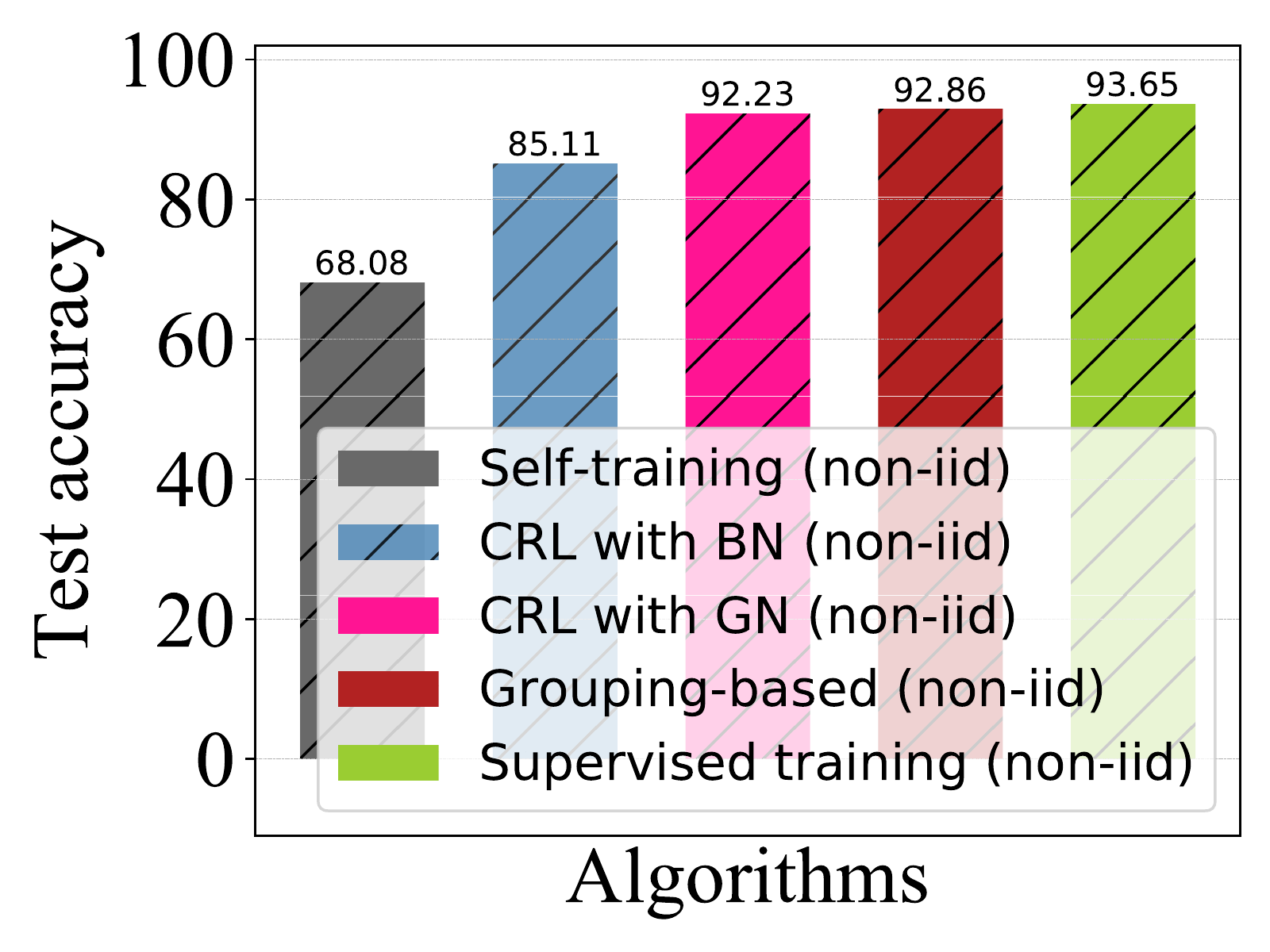}
    \caption{\footnotesize
    (Left) Test accuracy of different methods in the iid setting ($R=0.0$) on Cifar-10.
    (Right) Test accuracy of different methods in the non-iid setting ($R=0.4$) on Cifar-10.
    }
    \label{fig:TestAccComparisonbetweendifferentmethods}
\end{figure}

See Figure \ref{fig:TestAccComparisonbetweendifferentmethods} for the results. From the test accuracy results, we have the following observations.
\begin{itemize}
    \item When restricted to either the iid or the non-iid case, CRL improves significantly over self-training, but it cannot achieve the accuracy of supervised training.
    \item By comparing CRL with BN to CRL with GN, we show that GN improves the test accuracy.
    \item By comparing CRL with GN to the grouping-based method, we show that the grouping-based averaging improves the test accuracy compared to FedAvg.
\end{itemize}

\begin{figure}[ht]
    \centering
    \includegraphics[width=\figsize\linewidth]{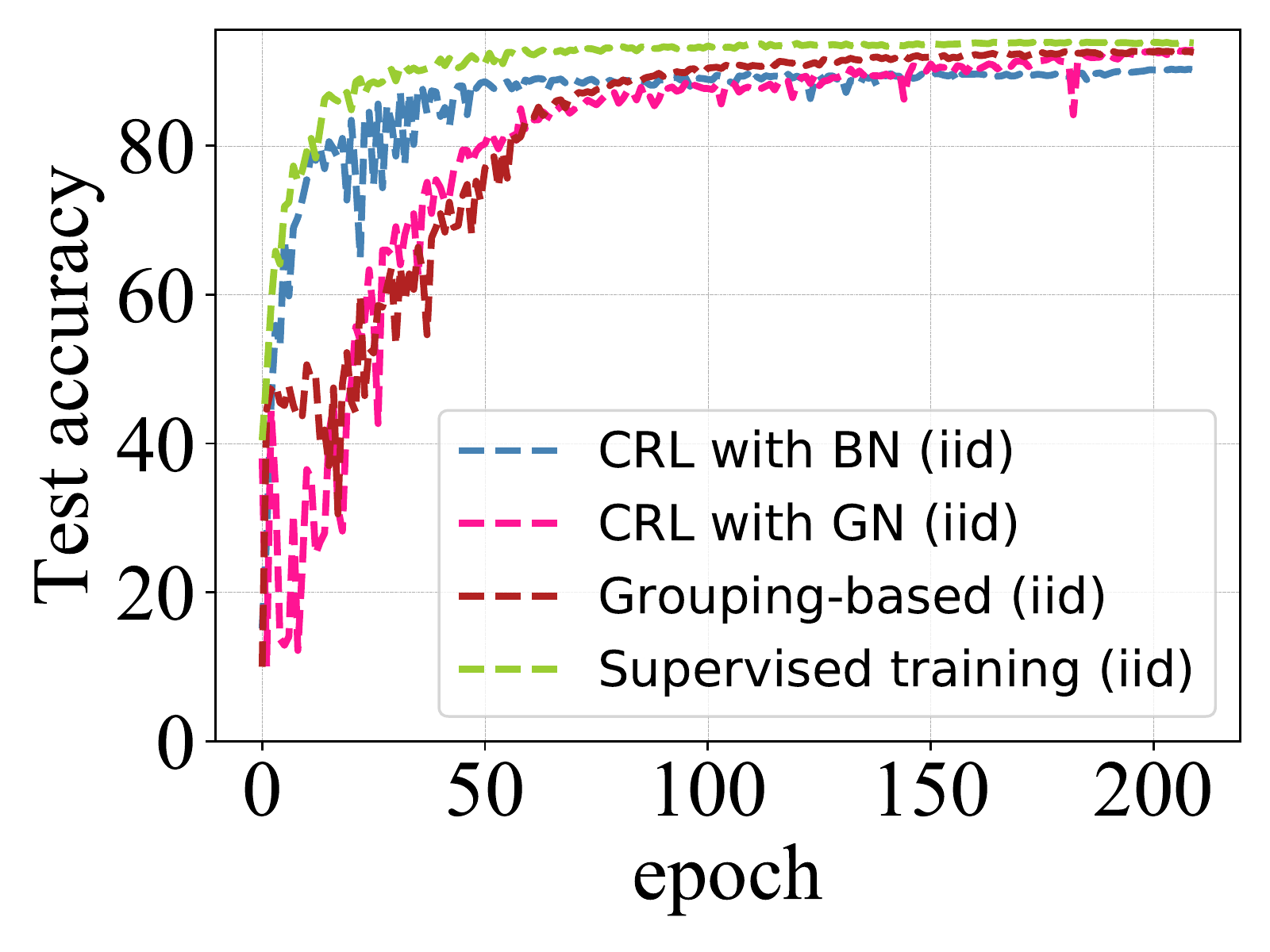}
    \includegraphics[width=\figsize\linewidth]{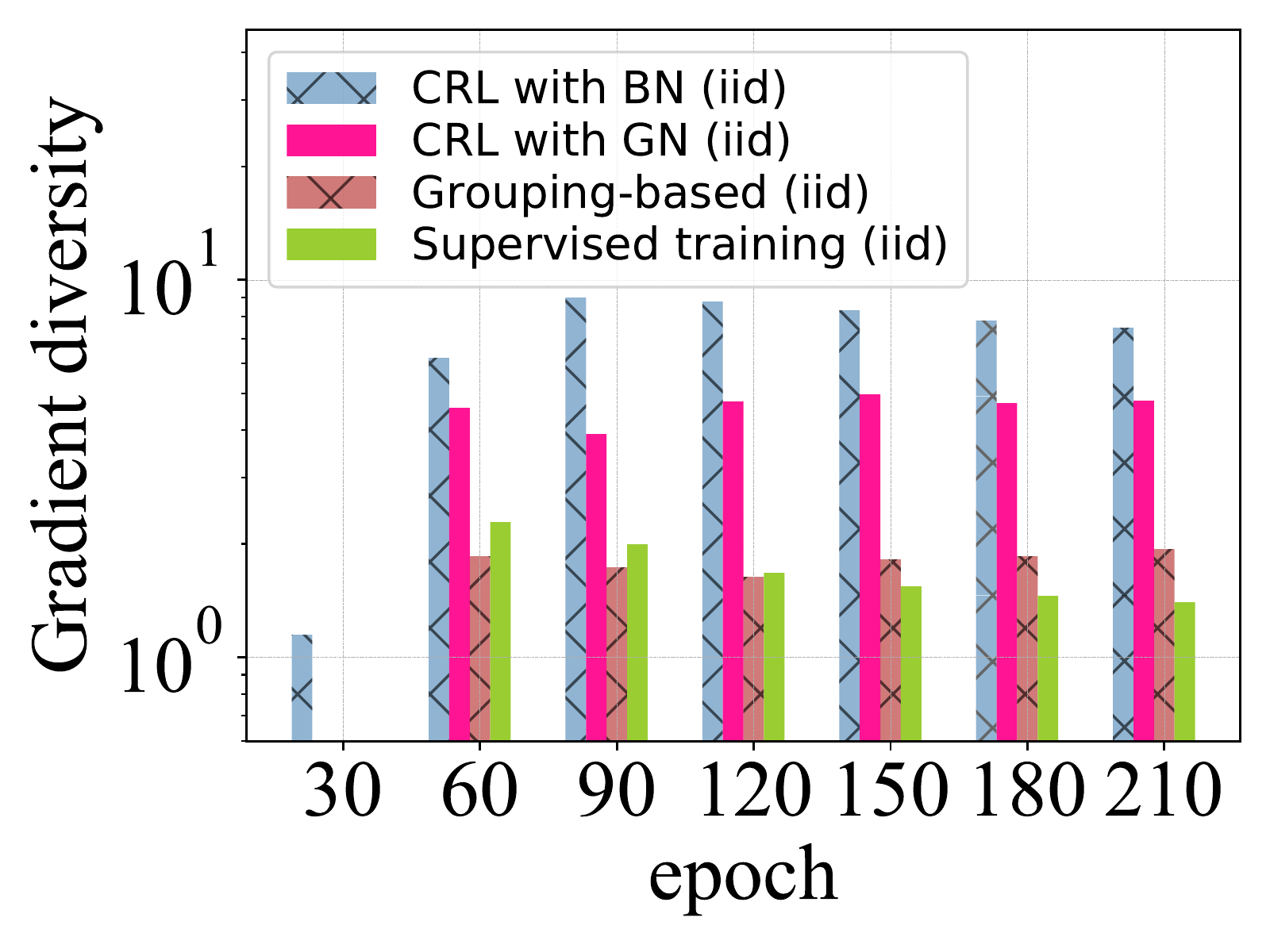}
    \caption{\footnotesize
    (Left) Convergence curves of different methods on Cifar-10 in the iid setting ($R=0$).
    (Right) The corresponding gradient diversity during training.
    }
    \label{fig:TrainingCurveandGradDiversityIID}
\end{figure}

\begin{figure}[ht]
    \centering
    \includegraphics[width=\figsize\linewidth]{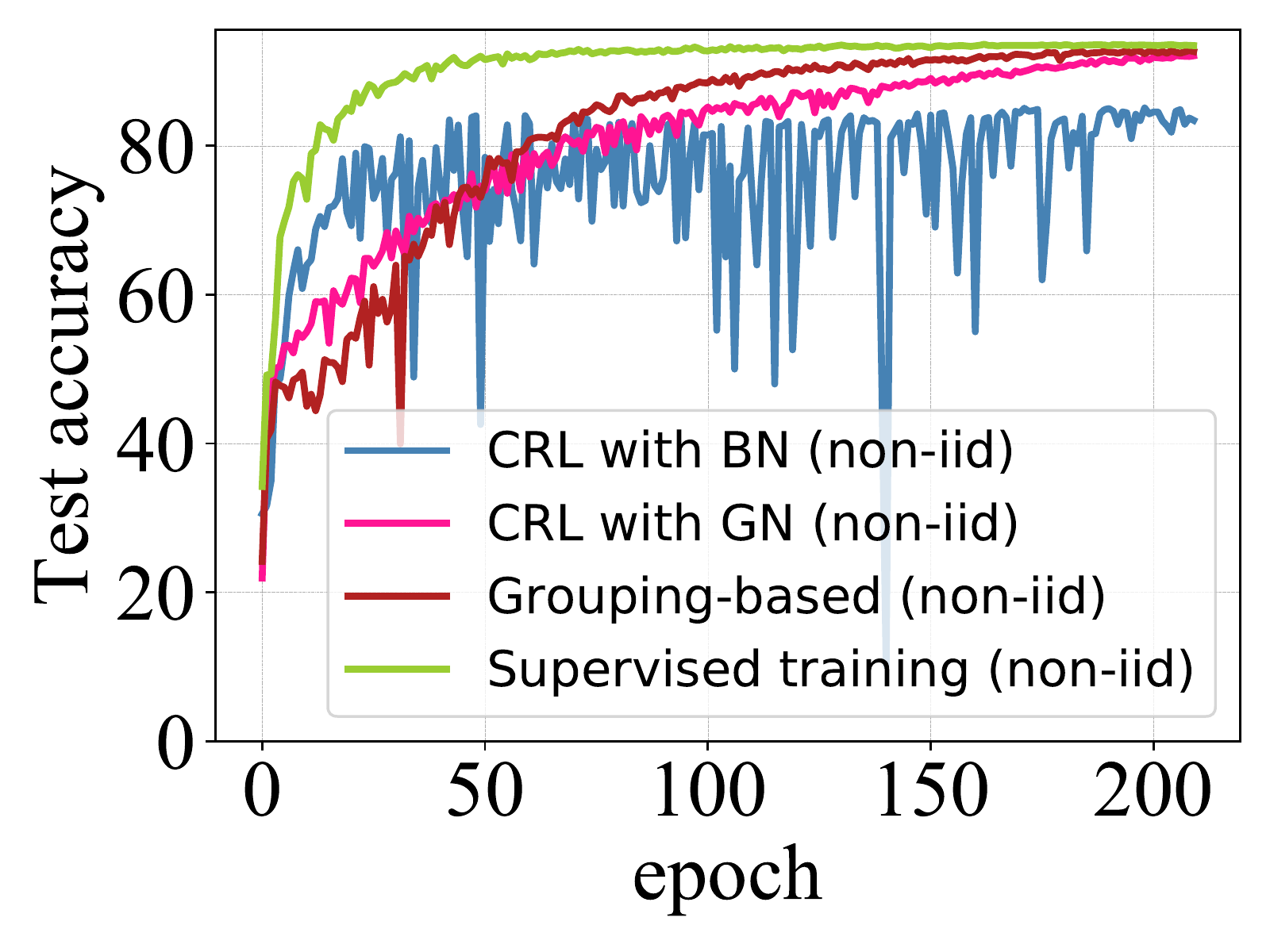}
    \includegraphics[width=\figsize\linewidth]{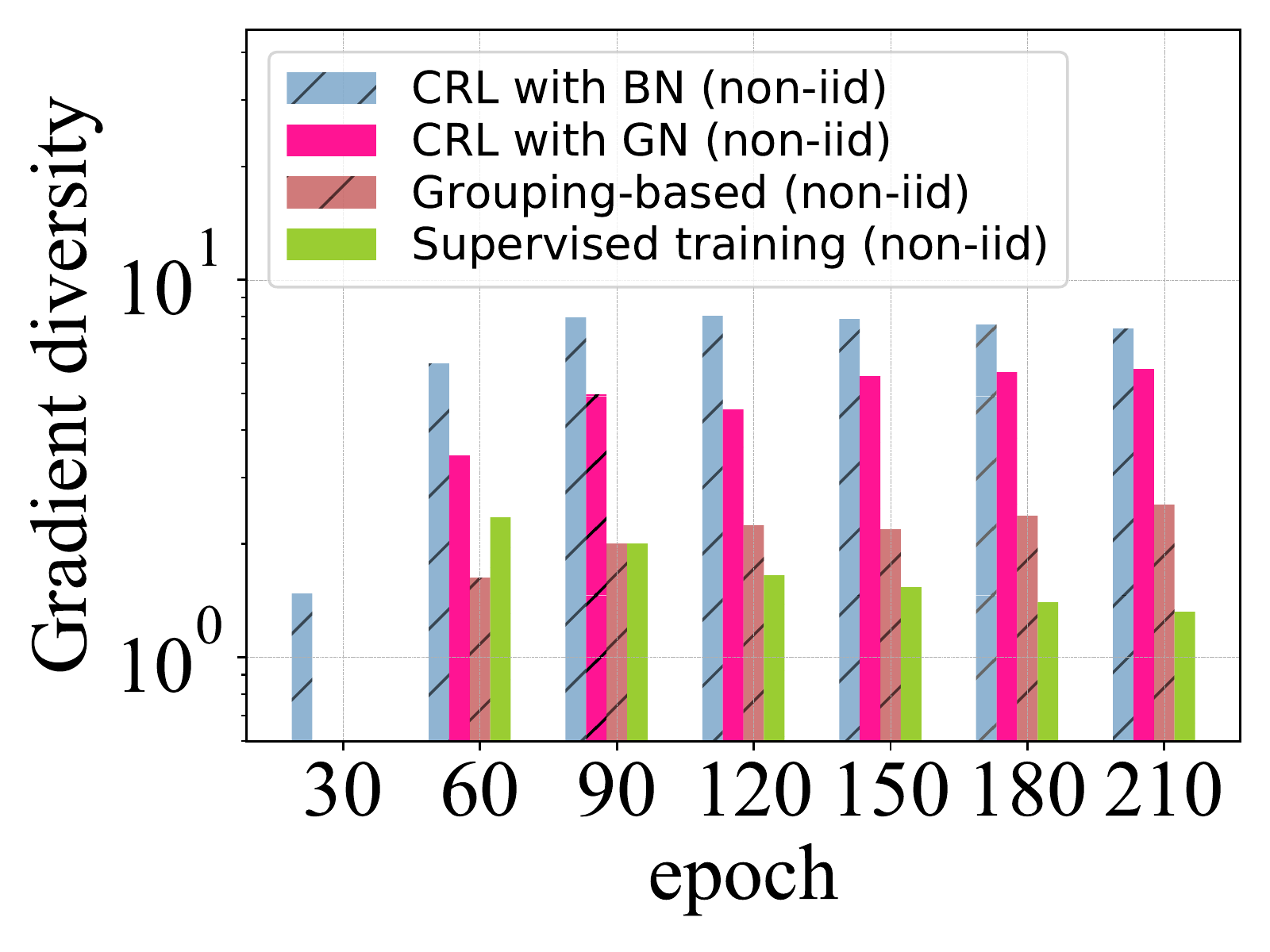}
    \caption{\footnotesize
    (Left) Convergence curves of different methods on Cifar-10 in the non-iid setting ($R=0.4$).
    (Right) The corresponding gradient diversity during training.
    }
    \label{fig:TrainingCurveandGradDiversityNONIID}
\end{figure}

\subsection{Gradient diversity analysis of different methods.}
In this subsection, we use the gradient diversity in Definition \ref{def:weight_diversity} to analyze different design choices. See \fref{fig:TrainingCurveandGradDiversityIID} and \fref{fig:TrainingCurveandGradDiversityNONIID}. The left plot shows the convergence curves. The right plot shows the gradient diversity values. Then, we parse the results:
\begin{itemize}[noitemsep, nolistsep, labelindent=0pt, leftmargin=*]
    \item When restricted to either the iid or the non-iid case, GN reduces gradient diversity compared to BN.
    \item Similarly, when restricted to either the iid or the non-iid case, grouping-based averaging reduces gradient diversity compared to FedAvg (see the comparison to CRL with GN, which uses FedAvg).
    \item The grouping-based method has a comparable gradient diversity value to supervised training.
\end{itemize}

\section{Results under different factors}
\label{sec:Results}

In this section, we extensively evaluate our grouping-based \ssfl solution, i.e., CRL objective combined with GN and grouping-based model averaging. We vary the environmental factors mentioned in \sref{sec:other_factors}, which include the non-iidness $R$, the communication period $T$, the number of labeled data $N_s$ in the server, the user number $K$ and the number of participating users $C$.
All the environmental factors used in this section are reported in~\tref{tab:ExperimentParameters}.

\textbf{Experiment settings.} We consider three datasets, Cifar-10, SVHN, and EMNIST in our empirical evaluation. 
We use ResNet-18 as the training model on both Cifar-10 and SVHN datasets; and we use the same CNN model as~\cite{adafed} on EMNIST.
See Appendix~\ref{sec:experiment_details} for more details. 

\begin{figure}
    \centering
    \includegraphics[width=0.32\linewidth]{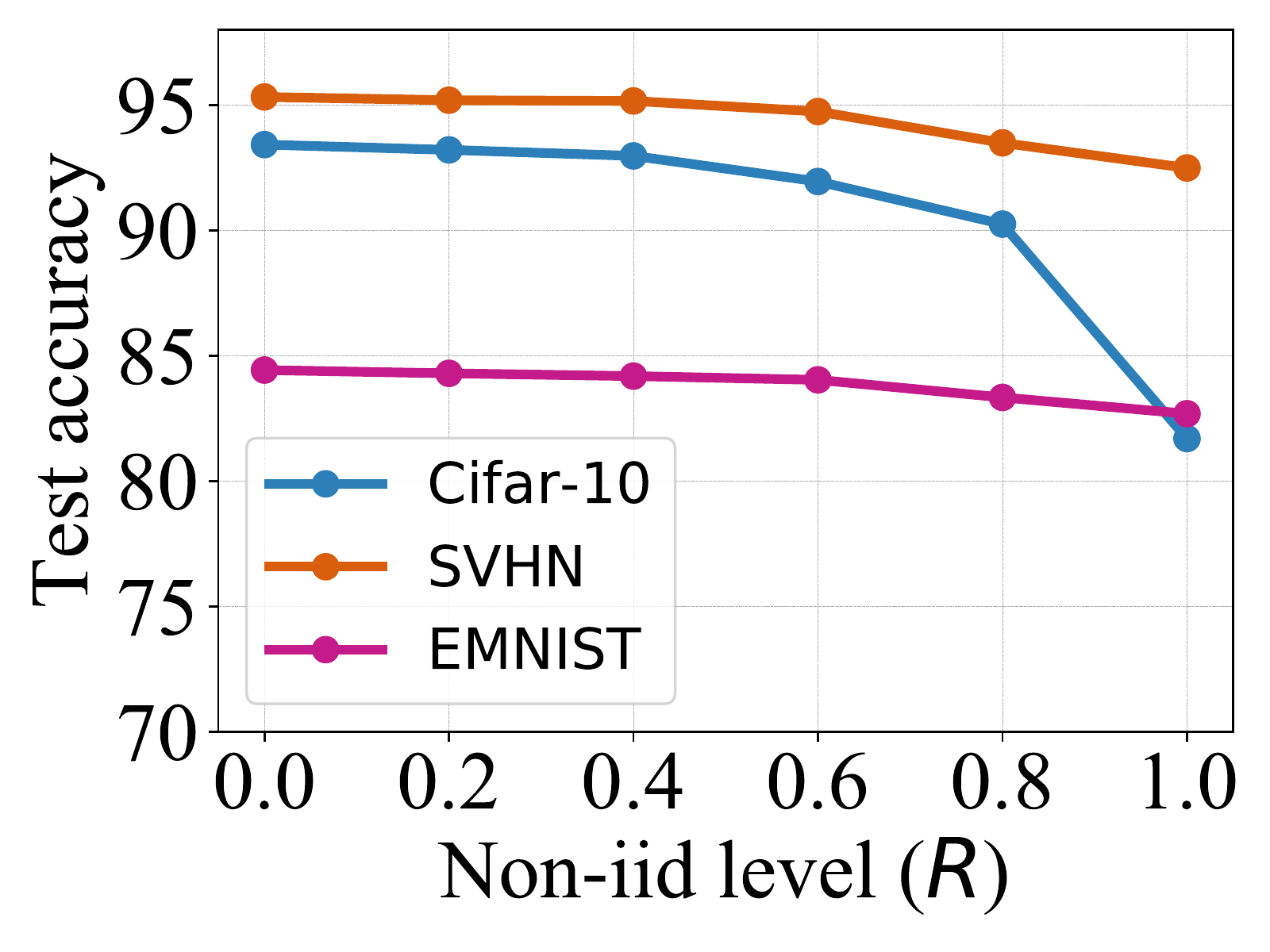}
    \includegraphics[width=0.32\linewidth]{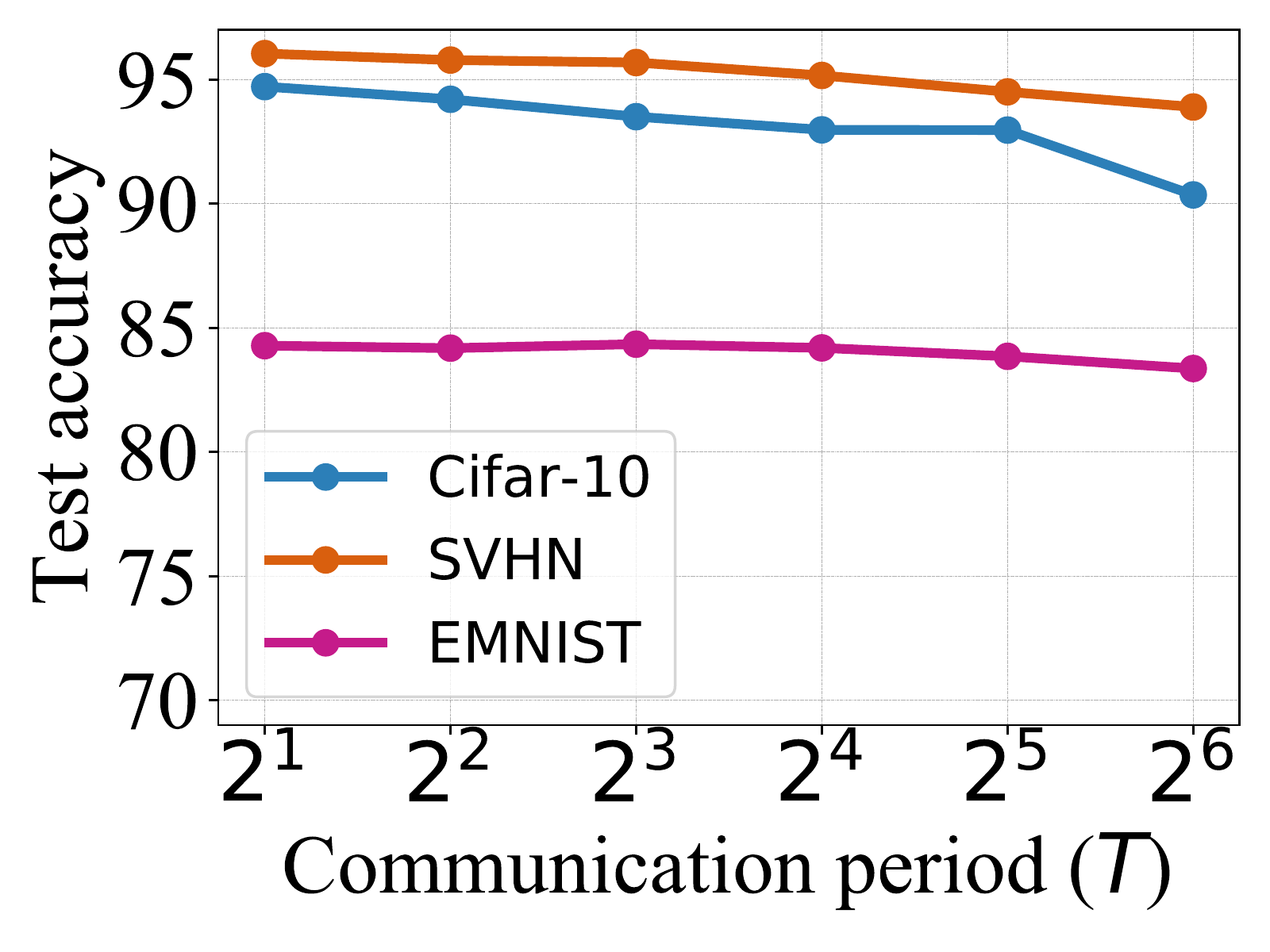}
    \includegraphics[width=0.32\linewidth]{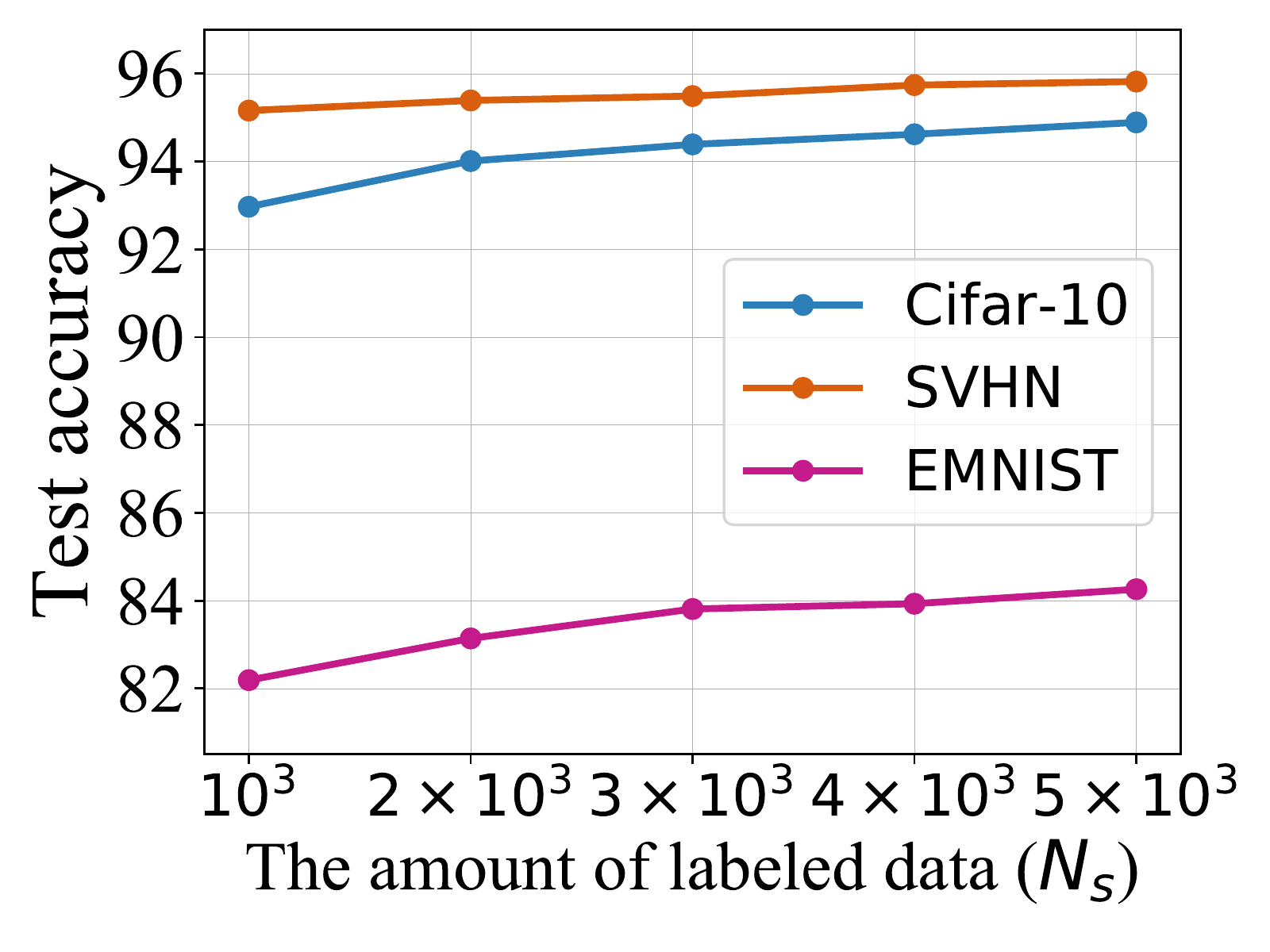}
    \caption{\footnotesize
    (Left) Comparison between different non-iid levels ($R$) on Cifar-10, SVHN and EMNIST.
    (Middle) Accuracy versus communication period $T$.
    (Right) Accuracy versus labeled data points in the server ($N_s$). 
    }
    \label{fig:impact_r_ns_cp}
\end{figure}

\subsection{Impact of $R$, $T$, and $N_s$}
\label{sec:impact_r_cp_ns}

In this subsection, we study the effect of the first three environmental factors.
First, we illustrate the effect of the non-iid level $R$ (defined in~Definition \ref{def:R}). 
For Cifar-10, SVHN and EMNIST, the experiment parameters are reported respectively in rows 1-3 of~\tref{tab:ExperimentParameters}. For each experiment, we fix all the parameters only except the non-iidness parameter $R$.
The results are shown in the left of~\fref{fig:impact_r_ns_cp}. 
When $R=0$, each user has the same empirical class distribution which is uniform. When $R=1$, each user only has a single class of data.
As can be seen, the accuracy decreases as the non-iid level $R$ increases (from 93.42\% to 81.7\% on Cifar-10, from 95.32\% to 92.49\% on SVHN, and from 84.43\% to 82.69\% on EMNIST.)
This is in accord with our intuition that iid data distribution typically leads to the best result. 

We also illustrate the effect of the communication period $T$ on Cifar-10, SVHN, and EMNIST. 
For these three datasets, the experiment parameters are reported in rows 4-6 of~\tref{tab:ExperimentParameters}.
In these experiments, we again only vary the communication period $T$ while holding all the remaining parameters fixed.
The middle of \fref{fig:impact_r_ns_cp} presents our results. 
Increasing $T$ (i.e., communicating less frequently) leads to a worse generalization performance. 
This is explainable since the local model can overfit when $T$ is large. 
In addition, the convergence behaviors on Cifar-10 at different $T$ can be found in~\fref{fig:cpCifar10}. 

Then, we investigate the impact of the number of labeled samples $N_s$ in the server.
For the experiments on three datasets, the experiment parameters are shown in rows 7-9 of~\tref{tab:ExperimentParameters}.
The results are shown in the right part of \fref{fig:impact_r_ns_cp}.
We notice that increasing the amount of labeled data in the server can improve the final generalization performance. 
For example, with 5000 labeled samples, the test accuracy values on all the three datasets are higher as compared to 1000 labeled data, e.g., for Cifar-10 the improvement is 1.92\%, for SVHN the improvement is 0.66\%, and for EMNIST the improvement is 2.07\%.
These results are reasonable since the increase in the amount of labeled data can make the model trained by the server more accurate, which helps the users obtain more accurate pseudo-labels. 
In the extreme case where the server has the entire labeled training dataset, the situation degrades to a supervised learning setting.

\subsection{Impact of $C$ and $K$, and the effectiveness of grouping-based average when $C$ is large}
\label{subs:FixCk}

\begin{table}[H]
\begin{minipage}[t]{1.0\linewidth}
    \caption{\footnotesize
    Accuracy versus amount of communicating users $C$  on Cifar-10 and SVHN. 
    Here, ``$^*$'' means we train SVHN for $E=120$ epochs instead of $E=40$ epochs for normal SVHN~training.
    }
    \centering
    \begin{adjustbox}{width=\tabsize\textwidth,center} 
    \begin{tabular}[t]{lcccccccccccccccccc}
    \toprule
 Dataset &  $K=10$, $C=10$& $K=20$, $C=20$ & $K=30$, $C=30$ \\
  
  \hc Cifar-10 
               & 92.86$\%$ & 92.93$\%$ & 92.12$\%$ \\
  
   SVHN  
        & 95.49$\%$  & 94.99$\%$  & 78.77$\%$ (94.93\%$^*$)\\
  \midrule
   &  $K=10$, $C=10$ & $K=20$, $C=10$ & $K=30$, $C=10$\\
  
 \hc  Cifar-10   & 92.86$\%$ & 93.19$\%$ & 92.84$\%$ \\
  
   SVHN   & 95.49$\%$  & 95.43$\%$  & 93.56$\%$ \\
    \bottomrule
    \end{tabular}
    \label{tab:CommunicationVolumeResults_on_cifar_svhn}
    \end{adjustbox}
\end{minipage}\hfill
\end{table}

\begin{table}[H]
\begin{minipage}[t]{1.0\linewidth}
\newcommand{\tabincell}[2]{\begin{tabular}{@{}#1@{}}#2\end{tabular}}
\caption{\footnotesize  
Accuracy versus the number of communicating users $C$ on EMNIST dataset}
  \centering
  \begin{adjustbox}{width=\tabsize\textwidth,center} 
  \begin{tabular}{lccccccccccccc}
\toprule
 Dataset & \tabincell{c}{$K=47$\\ $C=10$} & \tabincell{c}{$K=47$\\ $C=30$} & \tabincell{c}{$K=47$\\ $C=47$}\\
 \midrule
\hc  EMNIST (FedAvg) & 83.07$\%$ & 79.05$\%$ & 65.48$\%$\\

  EMNIST (Grouping-based)  & 84.43$\%$ & 83.12$\%$ & 82.95$\%$\\
\bottomrule
  \end{tabular}
  \label{tab:CommunicationVolumeResults_on_emnist}
  \end{adjustbox}
\end{minipage}\hfill
\end{table}

In this subsection, we analyze the remaining two environmental factors $C$ and $K$. Again, we change one specific environmental factor while holding all the other factors fixed. The settings of the environmental factors for the experiments in this subsection are reported in rows 10-18 of~\tref{tab:ExperimentParameters}. 

\begin{figure}
    \centering
    \includegraphics[width=0.32\linewidth]{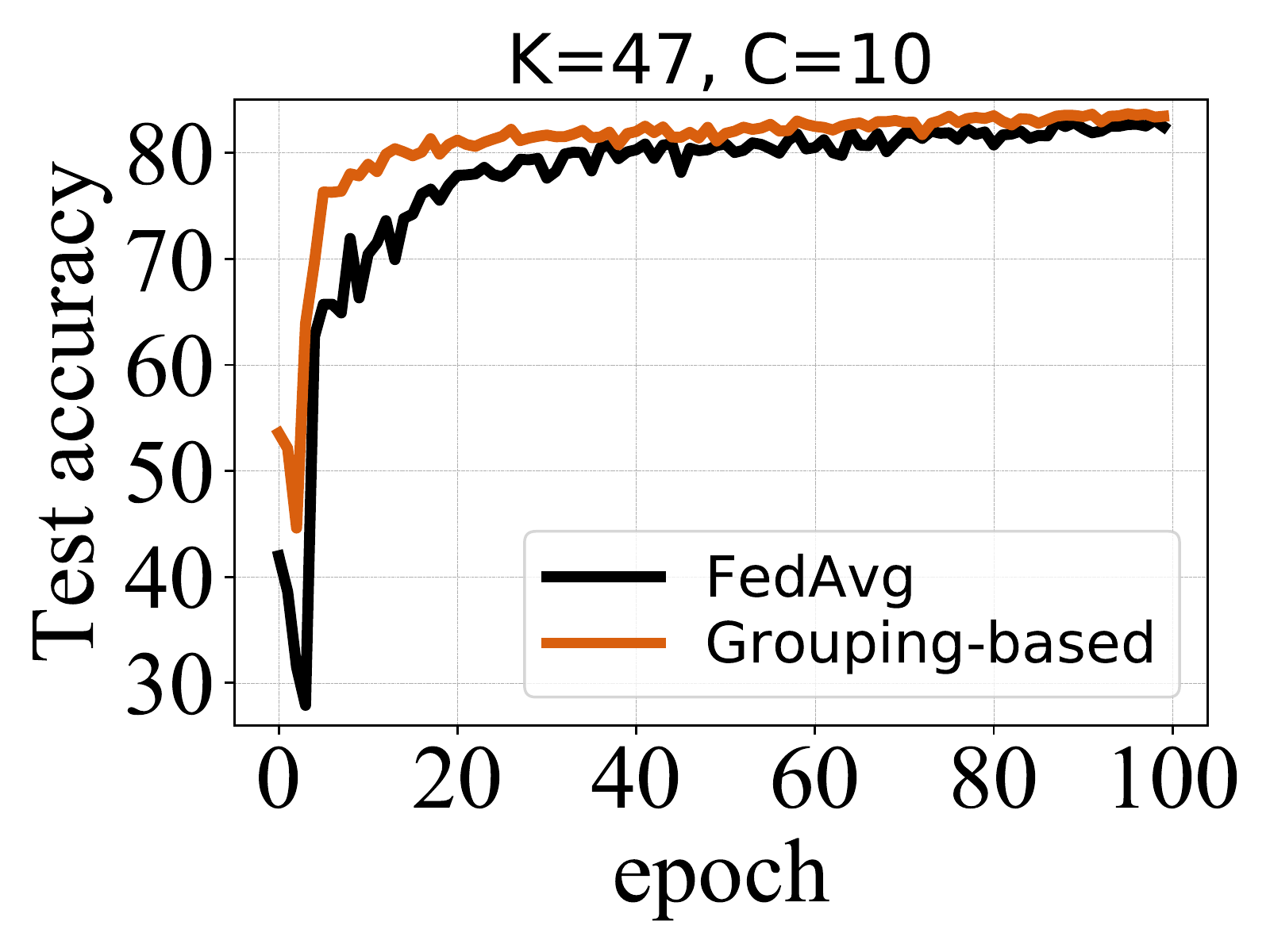}
    \includegraphics[width=0.32\linewidth]{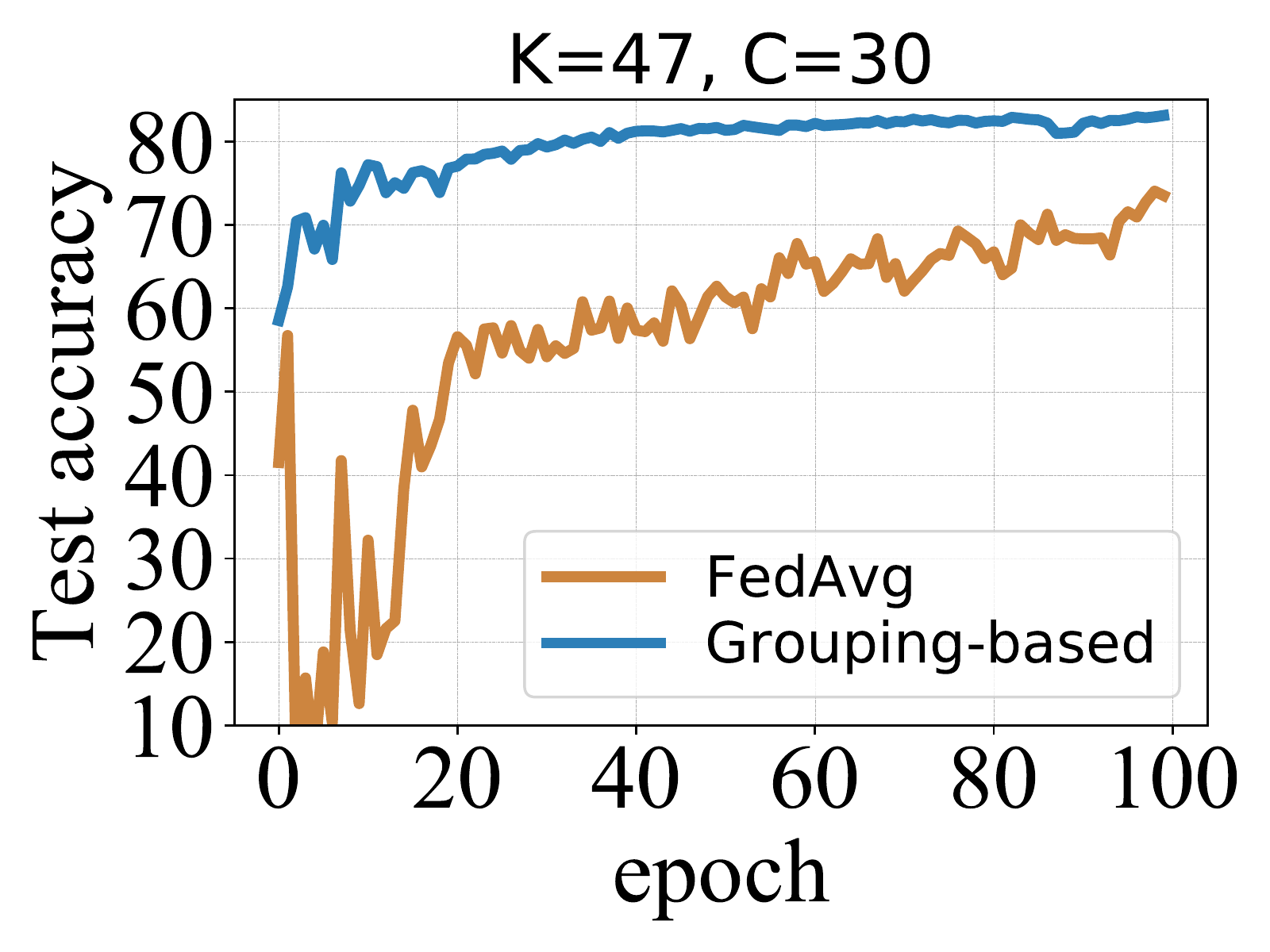}
    \includegraphics[width=0.32\linewidth]{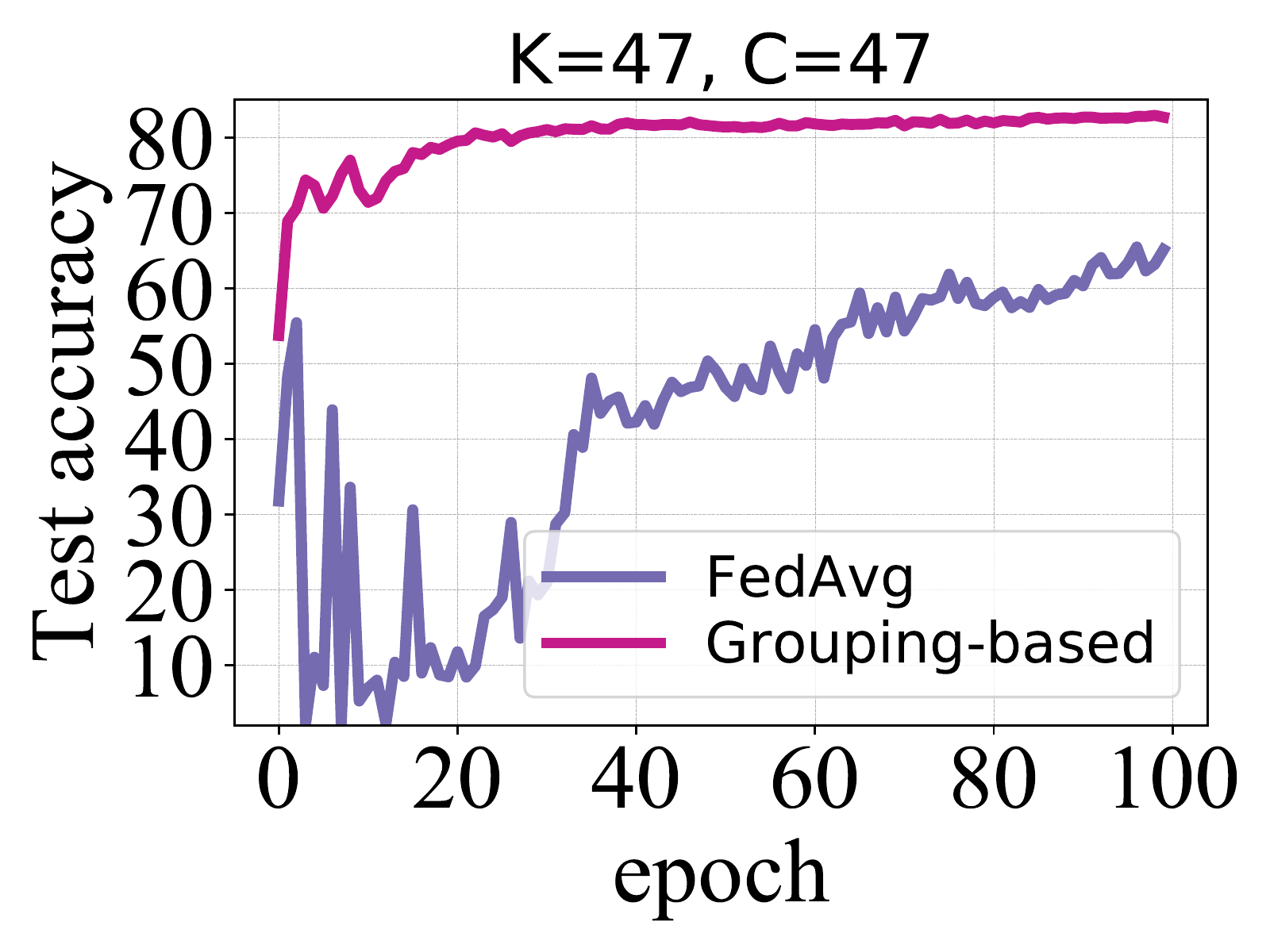}
     \includegraphics[width=0.32\linewidth]{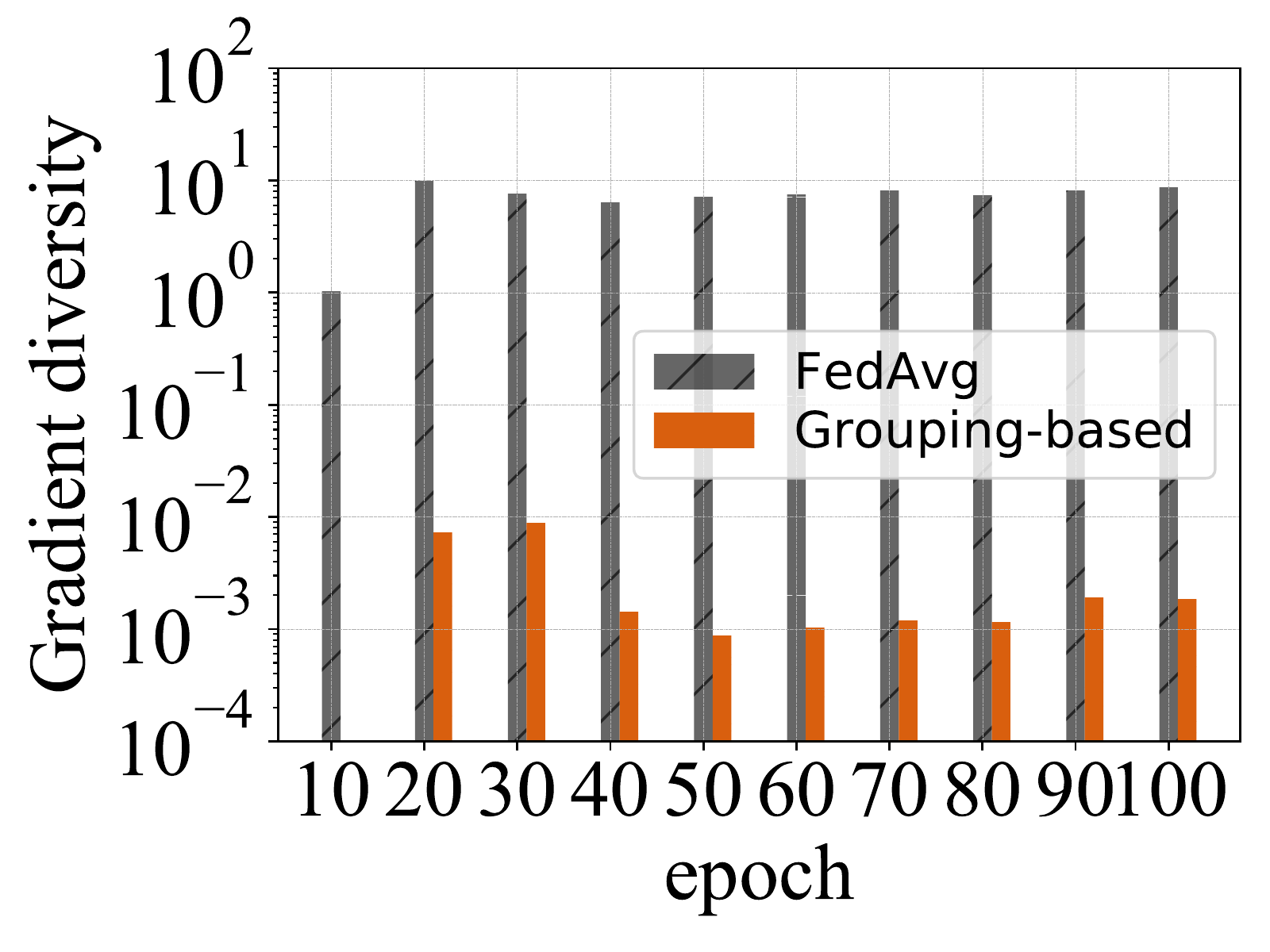}
    \includegraphics[width=0.32\linewidth]{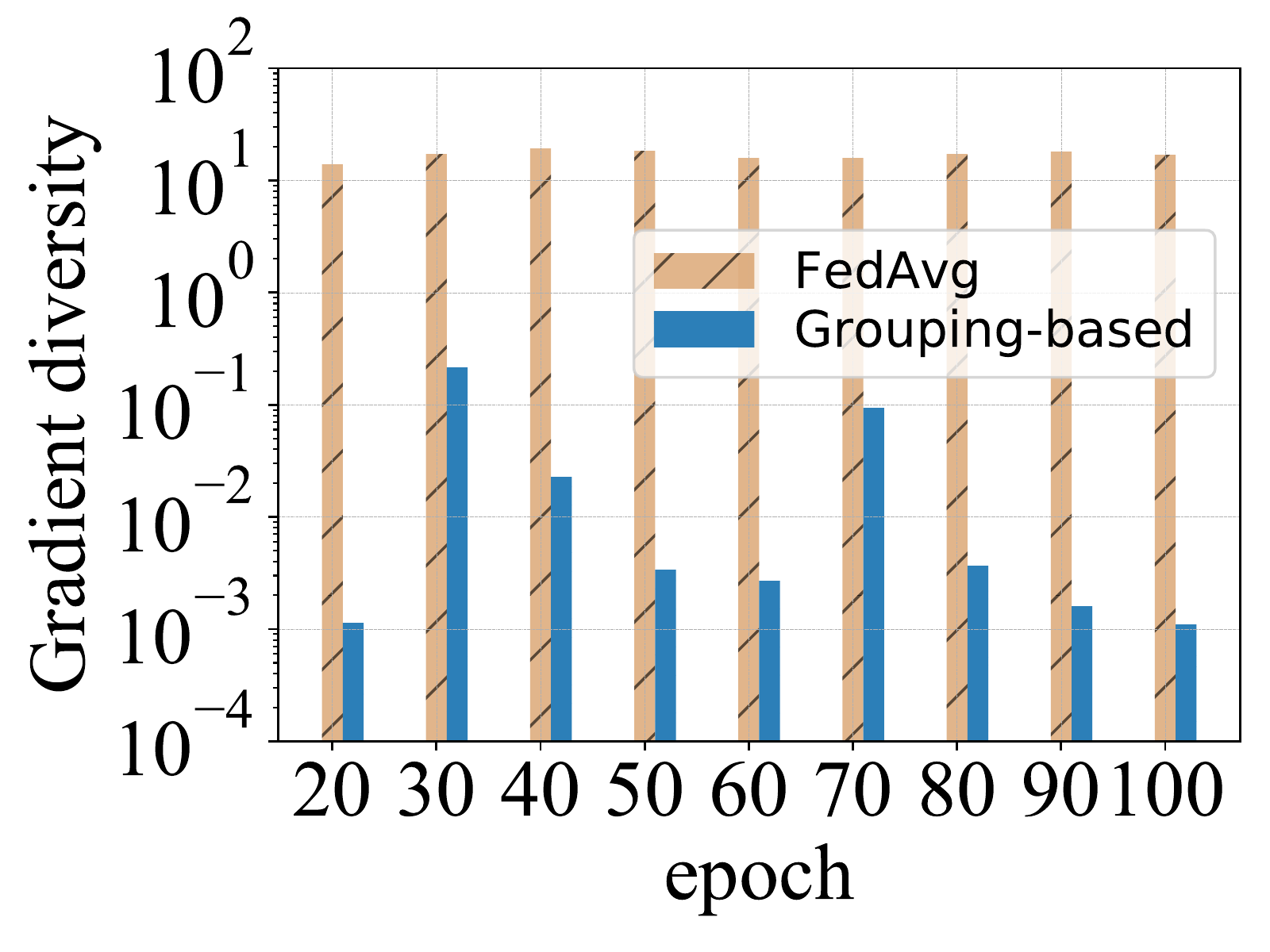}
    \includegraphics[width=0.32\linewidth]{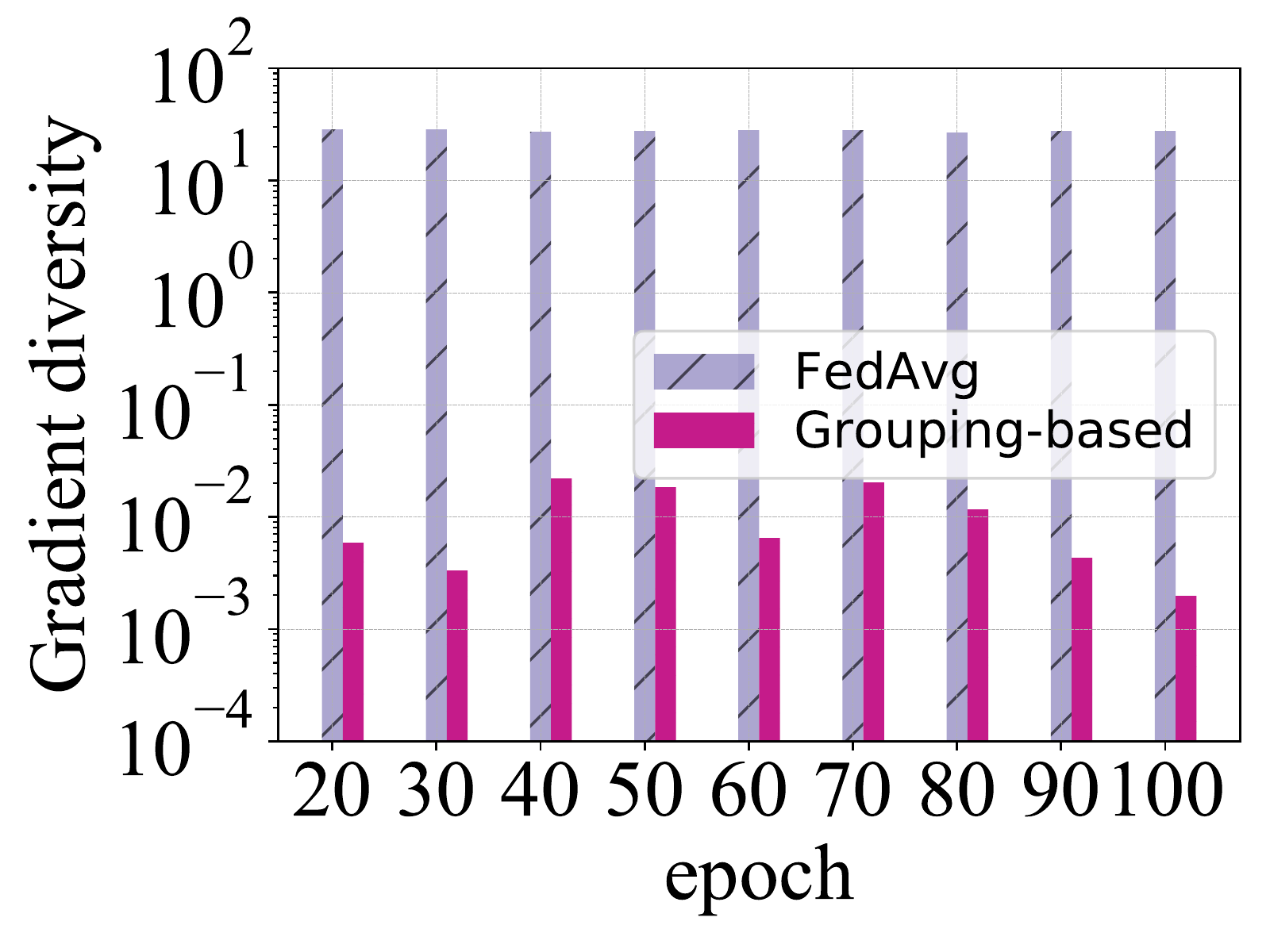}
    \caption{\footnotesize
    (Top) Convergence curves of FedAvg method on EMNIST when (left) $C=10$, (middle) $C=30$ and (right) $C=47$. (Bottom) The corresponding results on gradient diversity.
    }
    \label{fig:AccanddiversitiesonEMNIST}
\end{figure}

The results of Cifar-10 and SVHN are shown in~\tref{tab:CommunicationVolumeResults_on_cifar_svhn}, and the result of EMNIST is presented in~\tref{tab:CommunicationVolumeResults_on_emnist}. 
On the top of~\tref{tab:CommunicationVolumeResults_on_cifar_svhn}, we set $C=K$ and increase $K$.
At the bottom of~\tref{tab:CommunicationVolumeResults_on_cifar_svhn}, we show the result with fixed $C=10$ and various $K$ (from 10 to 30). 

As can be seen from the top of~\tref{tab:CommunicationVolumeResults_on_cifar_svhn}, increasing the number of users $K$ has a marginal effect ($<$1\%) on the accuracy, from $K=10$ to $K=30$. 
One notable thing here is that with $K=30$, if we train 40 epochs on SVHN, the accuracy is 78.77\%, which is 16.72\% lower than $K=10$. 
If we increase the training epochs from 40 to 120 for $K=30$ on SVHN, the final accuracy is 94.93\%. One can refer to~\fref{KSVHN} for the convergence curve of this experiment. 

Counterintuitively, when comparing the results at the bottom of~\tref{tab:CommunicationVolumeResults_on_cifar_svhn} to the results on the top, the results when $C<K$ are consistently better than when $C=K$. 
Particularly, the $K=30$, $C=10$ case outperforms $C=30$ by 0.72\% on Cifar-10 and by 14.79\% on SVHN, respectively. 

Similar to the above results of Cifar-10 and SVHN, from the results of EMNIST shown in~\tref{tab:CommunicationVolumeResults_on_emnist}, one can clearly see that when $K$ is large, a large $C$ decreases the performance significantly if one does not use the grouping-based average. Particularly, the $K=C=47$ case is lower than $C=10$ by 17.59\%. However, this reduction in accuracy can be mitigated if we use the grouping-based method, which is only 
1.48\%. 

We proceed to study why the grouping-based averaging performs significantly better than FedAvg for the particular case when $C$ is large. See \fref{fig:AccanddiversitiesonEMNIST}. 
From the results, we can see that the large number of communicating users causes large gradient diversity. We can also see that grouping-based averaging can reduce gradient diversity and increase accuracy. See appendix~\sref{sec:Weightdiversityexperimentson Cifar10} for additional analysis on gradient diversity. We also conduct additional experiments to study the user participation rate $C/K$ in \sref{sxn:impact_c_over_k}.

\subsection{Comparing with other supervised/semi-supervised results}\label{sec:compare_jeong}

In this subsection, we compare our grouping-based method with other FL algorithms, in both semi-supervised and supervised settings. 
First, in the semi-supervised setting, we conduct the experiment on Cifar-10 
with exactly the same setting as a recent SSFL paper \cite{jeong2020federated}. 
For the Cifar-10 data, according to Table 1 in \cite{jeong2020federated}, we set $N_s=5000$, $K=100$, $C=5$, and $R=0$ (which is the iid case) or $R=1$ (which is the most difficult non-iid case).
From \tref{JeongCifar10}, one can see that our grouping-based solution outperforms the method proposed in~\cite{jeong2020federated} by a large margin.
We notice that the results in \cite{jeong2020federated} are presented in two different settings including the \emph{labels-at-server} setting and the \emph{labels-at-client} setting. 
The first setting is the same as our paper, i.e., only the server has labeled data, while the users have unlabeled data.
In this setting, $N_s=5000$ labeled data are own by the server. 
The second setting is different but it is straightforward to apply our grouping-based solution. 
In this setting, $N_s=5000$ labeled data are distributed to 100 users. In each round, $C=5$ users are random selected to communicate with the server.
See Appendix \ref{sec:UersSideSemi} for the details of adapting our solution to the label-at-the-client setting.

\begin{table}
\caption{\footnotesize
Comparing with \cite{jeong2020federated} in exactly the same setting on Cifar-10. The model in \cite{jeong2020federated} is ResNet-9.
}
\centering
\begin{tabular}[t]{lcccccccccccccccccc}
\toprule
      & FedMatch & Ours\\
    \midrule
   \hc  Labels-at-client (iid) &  53.51\%  &   {\bf 71.61\%}\\
   Labels-at-client (non-iid) &  54.26\%  &  {\bf 69.05\%}\\
   \hc  Labels-at-server (iid) &  46.81\%  &   {\bf 63.32\%}\\
   Labels-at-server (non-iid) &  47.11\%  &  {\bf 63.24\%}\\
    \bottomrule
\end{tabular}
\label{JeongCifar10}
\end{table}

\begin{table}
    \caption{\footnotesize
   Comparison with supervised FL. Here, ``$^*$'' is calculated according to the setting in DataSharing.
    }
    \label{ComparisonwithsupervisedFL}
    \centering
    \begin{adjustbox}{width=\tabsizeThreeFour\linewidth} 
    \begin{tabular}[t]{lcccccccccccccccccccccccccccc}
   \toprule
  Method  & Test accuracy \\
\midrule
{Supervised FedAvg } & 78.52$\%$ ($R=0.29$)\\
\hc {DataSharing }  & 81.82$\%$ ($R=0.29^*$) \\
 {Grouping-based (ours)}  & {\bf 92.96$\%$ ($R=0.4$)}\\
\bottomrule
    \end{tabular}
    \end{adjustbox}
\end{table}

We also compare our solution with supervised FL methods in~\tref{ComparisonwithsupervisedFL}. 
We choose two supervised FL methods for comparison: Supervised FedAvg~\cite{Communication_Efficient_Learning} and DataSharing~\cite{FedNoniidData}. 
We set $K=10$, $C=10$ and $T=32$, and we use ResNet-18 to be the model for training. 
The non-iid setting of DataSharing~\cite{FedNoniidData} corresponds to the scenario where we set $R=0.29$. 
For our solutions, we set $N_s=1000$ and $R=0.4$. The detailed experimental parameters of different methods can be seen from rows 22-25 of~\tref{tab:ExperimentParameters}. 
Larger $R$ means a higher non-iid level and thus a more difficult scenario (which we have experimentally demonstrated in \fref{fig:impact_r_ns_cp}). From~\tref{ComparisonwithsupervisedFL} we see that the performance of our method ($R=0.4$) on Cifar-10 is still better than Supervised FedAvg ($R=0.29$) and DataSharing methods ($R=0.29$) even when the scenario of $R=0.4$ is more difficult. 

We also compare our method with EASGD~\cite{EASGD} and OverlapSGD~\cite{Overlap} which are communication efficient algorithms under supervised settings. 
We use the same parameters in their papers, i.e., $K = 16$, $R = 0.4$, $C = 16$ and $N_s = 1000$ on Cifar-10. See rows 26-29 of~\tref{tab:ExperimentParameters} for the details.
The results are shown in~\tref{SupervisedResultsOverlapSGD}. 
We see that our result has better accuracy than both EASGD and OverlapSGD. 
Particularly, even with $T=32$ (larger $T$ means a harder scenario; see \fref{fig:impact_r_ns_cp}), our method has 0.80\% or 0.29\% better performance, as compared to EASGD or OverlapSGD in the setting of $T=2$, respectively.
Note that both EASGD and OverlapSGD are supervised algorithms, which means they have all the data labels.

There are other important issues we consider. 
For example, we evaluate the performance of our grouping-based method in fully supervised FL in~\sref{sec:GroupingbasedinSFL}, and we show that the improvement in this scenario is limited. Thus, our method is more suitable in the \ssfl setting.
We also compare our grouping-based solution with FixMatch~\cite{Fixmatch} in~\sref{sec:GroupingbasedandFixMatch}).
We show that our results are comparable to FixMatch even if FixMatch uses centralized setting.
Moreover, we extend our current methods to the setting where users can have both labeled and unlabeled samples in~\sref{sec:UersSideSemi}, i.e., the labels-at-client case.
We also provide additional results on another semi-supervised dataset STL-10, see~\sref{sec:PerformanceSTL10}.
Finally, since for federated learning, we are interested in studying what happens when the number of users is large, we provide an additional experiment with the number of users set to 470 in~\sref{sec:EMNISTLargeK}. We show that the grouping-based method outperforms FedAvg in this setting as well.

\begin{table}
    \caption{\footnotesize
    Comparison with two other supervised FL algorithms EASGD and OverlapSGD on Cifar-10.
    }
    \label{SupervisedResultsOverlapSGD}
    \centering
    \begin{adjustbox}{width=\tabsizeFive\linewidth} 
    \begin{tabular}[t]{lcccccccccccccccccccccccccccc}
   \toprule
  Method  & $T=2$ & $T=8$ & $T=32$\\
\midrule
{EASGD} & 91.12$\%$ & 88.88$\%$ & $-$\\
\hc {OverlapSGD} & 91.63$\%$ &91.45$\%$  & $-$\\
 {Grouping-based (ours)} & {\bf 94.22$\%$}& {\bf 93.58$\%$} &  {\bf 91.92$\%$}\\
\bottomrule
    \end{tabular}
    \end{adjustbox}
\end{table}

\section{Related work}

\ifisarxiv

{\bf{Federated learning.}} 
Federated learning (FL)~\cite{konecny2016federated, Communication_Efficient_Learning,FederatedSemisupervisedLearning,ExploitingUnlabeledData,FedNoniidData,FederatedAdversarialDomainAdaptation,ReducingClientResource, ATOMO, OneShotFedLearning,TernaryCompression} is a decentralized computing framework that enables multiple users to learn a shared model while potentially protecting the privacy of users (although recent work~\cite{AttacksFL} shows this may not be the case). 
Federated Averaging (FedAvg)~\cite{Communication_Efficient_Learning}, which is the most popular FL algorithm, shows good performance when the data distribution across users is iid.
However, in the non-iid case, the performance can significantly degrade. 
In fact, dealing with non-iid distributions is deemed by many to be one of the most critical challenges in FL~\cite{FedNoniidData,FederatedAdversarialDomainAdaptation,FederatedDistillationandAugmentation}. 
In~\cite{FedNoniidData}, a data-sharing method is proposed to improve the final accuracy. 
However, sharing massive data among all users requires both large storage space as well as stable connections between users and the server. 
Importantly, all of these methods require the data stored by the local users to come with ground-truth labels (in order to perform model updates locally). 
The FL problem in the semi-supervised setting, when users do not have labels, however, is ``relatively ignored'' and has ``little prior arts,'' as mentioned in a recent survey paper~\cite{FederatedSemisupervisedLearning}.

In addition to the challenge of the non-iidness of the data distribution and the need for local ground truth labels, communication efficiency is another critical problem in FL~\cite{ATOMO,PracticalSecureAggregation,FedProx,TwoStreamFederatedLearning,mcmahan2019communication}. 
One way to relieve the communication burden of FL is to increase the period (the number of local gradient descent iterations) between consecutive communication stages. 
However, when this communication period increases, the diversity between different models increases, and the fusion of these models by the server may lead to accuracy degradation. 
To handle this problem, \cite{FedProx} proposes FedProx, which adds a proximal term in the user local loss function to restrict the update distance between the local model and the global model. 
Other work considers gradient compression and model compression to reduce the communication cost~\cite{ATOMO,TernaryCompression,mcmahan2019communication}. 
For example, \cite{ATOMO} proposes atomic sparsification of stochastic gradients, which leads to significantly faster distributed training. 

{\bf{Semi-supervised Learning.}} 
Semi-supervised learning (SSL) is a classical problem when only a small fraction of data is labeled~\cite{IntroductiontoSemiSupervisedLearning,Fixmatch, Selftraining,Tritraining, NIPSPImodel, NIPSMeanteachers, MixMatch, ReMixMatch}. SSL includes many impactful algorithms. For example, self-training \cite{yarowsky1995unsupervised} uses the
model's own predictions on unlabeled data to supervise the training of the same model. 
Co-training \cite{blum1998combining} trains two models in parallel using two set of conditionally independent features, and let the two models supervise each other.
Tritraining \cite{Tritraining} first trains three classifiers using bootstrap. Then, each classifier is trained on samples agreed by the other two classifiers.
Graph-based SSL \cite{zhu2003semi} propagates labels on a graph generated by the similarity between different samples. 

In recent years, the problem of SSL in the context of deep neural networks has been extensively studied. In \cite{MixMatch}, a specific consistency regularization is used: the average predictions on several augmented views of a single unlabeled sample is sharpened (using temperature scaling) and used to supervise the different predictions. Mixup \cite{MixUp} is further applied as a traditional regularization approach. Unsupervised data augmentation (UDA) \cite{xie2019unsupervised} applies AutoAugment \cite{cubuk2018autoaugment} to generate data-dependent augmentations to improve the performance.
In~\cite{Selftraining}, a self-training method is introduced, which improves the state-of-the-art accuracy on ImageNet~\cite{ImageNet}, even compared to supervised learning~\cite{ResNet,EfficientNet}.
In~\cite{Fixmatch}, a simplified SSL loss is proposed which directly uses pseudo-labeling to provide consistency regularization on augmented samples. 

{\bf{SSFL.}}
Regarding the motivation of \ssfl, a recent survey paper \cite{FederatedSemisupervisedLearning} raises the practical concern that users may not have ground-truth labels. 
Regarding the problem formulation, \cite{jeong2020federated} is the most relevant. It uses a consistency loss to achieve the agreement between users, which aligns with the intuition in our method to reduce gradient diversity. 
The setting of \cite{ExploitingUnlabeledData} is also similar to ours but focuses on the label-at-client scenario.
Apart from these two, there are several other contemporary papers that consider different settings.
For example, the paper \cite{itahara2020distillation} considers using shared unlabeled data for distillation-based message exchanging. 
The paper \cite{OneShotFedLearning} assumes that the unlabeled data is held by the server.
The paper \cite{kang2020fedmvt} focuses on the ``vertical'' FL setting in which the data is partitioned from the feature dimension.
Two other papers \cite{zhao2020semi,yang2020federated} use \ssfl in specific professional fields.
Another paper~\cite{papernot2016semi} studies semi-supervised private aggregation of an ensemble of teacher models trained on separate subsets of the whole dataset, which is not in the FL setting but is closely related.

\fi

\section{Conclusions}
\label{sec:conclusions}
We studied the semi-supervised federated learning (\ssfl) setting in which most samples are unlabeled. Based on the observations of large gradient diversity, we proposed to use GN and a novel grouping-based model averaging method. We conducted extensive evaluations in various scenarios to evaluate our solution. The results showed that our \ssfl method achieves better test accuracy even when compared to existing semi-supervised or supervised FL algorithms.

We emphasize that our solution can be extended to other FL scenarios, such as standard supervised FL (see \sref{sec:GroupingbasedinSFL}) and the label-at-client FL~\cite{FederatedSemisupervisedLearning} (see \sref{sec:UersSideSemi}).
Another challenging scenario worth mentioning is where there is a significant mismatch between the user data distributions and the distribution at the server, in which case the label supervision from the server may conflict with the information provided by users. 
We envision that techniques from unsupervised domain adaptation \cite{long2016unsupervised} are useful to address this problem. In addition, personalization \cite{Personalization,mansour2020three} is important for \ssfl because it can mitigate the mismatch between the data distributions at the server and at the users' side.
Although our work focuses on empirical analysis, it is meaningful future work to explore the theory behind the new \ssfl setting, e.g., by advancing recent theoretical results in non-iid FL \cite{li2019convergence,FedNoniidData} and combining with analysis of particular data augmentation schemes such as CRL.

\ifisarxiv
\subsection*{Acknowledgments}
We would like to thank Jianyu Wang and Daniel Rothchild for their valuable feedback.
Michael W. Mahoney would like to acknowledge the UC Berkeley CLTC, ARO, IARPA (contract W911NF20C0035), NSF, and ONR for providing partial support of this work. 
Joseph E. Gonzalez would like to acknowledge supports from NSF CISE Expeditions Award CCF-1730628 and gifts from Alibaba, Amazon Web Services, Ant Group, Ericsson, Facebook, Futurewei, Google, Intel, Microsoft, Nvidia, Scotiabank, Splunk and VMware.
Our conclusions do not necessarily reflect the position or the policy of our sponsors, and no official endorsement should be~inferred.
\else
\fi

\vspace{10mm}

\bibliography{ref.bib}
\ifisarxiv
\bibliographystyle{ieeetr}
\fi

\vspace{10mm}

\appendix

\counterwithin{figure}{section}
\counterwithin{table}{section}

\begin{center}
{\Large\textbf{Appendix}}
\end{center}

\section{Additional details on our empirical evaluation.}
\label{sec:experiment_details}

\subsection{Datasets} 
{{Cifar-10}}~\cite{Cifar} consists of images with 3 channels, each of size 32$\times$32 pixels. Each pixel is represented by an unsigned int8.
This dataset consists of 60000 color images from 10 classes, with 6000 images in each class. There are 50000 training images and 10000 test images. 
\noindent{SVHN}~\cite{SVHN} is obtained from images of house numbers in Google Street View images. 
It has 99289 digits from 10 classes. 
There are 73257 digits for training and 26032 digits for testing. 
All digits have been resized to a fixed resolution of 32$\times$32 pixels. 
\noindent{EMNIST}~\cite{EMNIST} is a set of handwritten digits which have been resized to a fixed resolution of 28$\times$28 pixels. 
It is an unbalanced dataset that has 814255 digits from 62 classes, including A-Z, a-z, and 0-9. 
However, since the uppercase and lowercase of some handwritten letters are difficult to distinguish, for these letters, following~\cite{EMNIST}, we combine the uppercase and lowercase classes into a new class. 
There are 15 merged letters in total, including [C, I, J, K, L, M, O, P, S, U, V, W, X, Y, Z].  
Thus, there are 47 classes left. 
To ensure every user has almost the same amount of data, following~\cite{EMNIST}, we truncate the training dataset to have 2400 data points per class and drop the rest. 
Notice that the above pre-processing step to truncate the dataset and the letter-merging step are both used in \cite{EMNIST} to improve the generalization accuracy.
In particular, there are 112800 digits for training, and we use the full test dataset (18800 digits) for testing.

\subsection{Data assignment procedures for a particular value of $R$} 
\label{subsec:DataDistribute}
For a dataset with $d$ classes, to synthesize a data assignment procedure such that the non-iid level is $R$ (see Definition \ref{def:R}), we follow the procedures below to distribute the data. 

\begin{enumerate}
\item[1.] ({\bf Server data})
We assign $N_s$ labeled training samples from each of the $d$ classes to the server. 
Recall that $N_s$ is the total number of samples at the server. Thus, the class distribution at the server is uniform.\footnote{Note that this step requires each class to have more than $N_s/d$ samples, which is satisfied in all of our experiments.}

\item[2.] ({\bf User data})
Let $n_j$ denote the number of samples left in class $j$ after distributing the data to the server.
The empirical distribution across different classes for the remaining data is $Q=[q_1, \dots, q_d]$ such that $\sum_{j=1}^d q_j = 1$, where $q_i=\frac{n_i}{\sum_{j=1}^d n_j}$. 
For a specific data assignment to the users, we use the \emph{main class} of a user to refer to the class with the maximum number of samples at the user. 
Let $m_j$ denote the number of users whose main class is $j$, and let $\mathcal{U}_j = \{\text{User}_{j1}, \text{User}_{j2}, \dots, \text{User}_{jm_j}$\} denote the set of such users. 
The $m_j$'s should be chosen to satisfy the constraint $\sum_{j=1}^d m_j = K$ in which $K$ is the total number of users and $d$ is the number of classes.

We use $\text{User}_{j1} \in \mathcal{U}_j$ as an example to illustrate the data assignment (the same for other users): 
\begin{enumerate}
    \item We first assign $n_jR/m_j$ unlabeled training samples for the main class $j$. 
    \item After that, for each class $i\in \{1, \dots, d\}$, we assign $(1-R) n_i q_j /m_j$ unlabeled samples, in which $i$ can be equal to the main class $j$.
\end{enumerate}
\end{enumerate}

According to the above data assignment procedures, for a user who belongs to $\mathcal{U}_j $(e.g., $\text{User}_{j1}$), the number of unlabeled data for each class $i$ is:
\begin{equation}
\label{eq:Nij}
N_{j,i} = \left\{ \begin{array}{l}
{n_i}R/{m_j} + {n_i}{q_j}(1 - R)/{m_j},~~\text{for }i = j\\
{n_i}{q_j}(1 - R)/{m_j},~~\text{for }i \ne j.
\end{array} \right.
\end{equation} The total number of unlabeled data held by this user is $\sum\nolimits_{i = 1}^d {{N_{j,i}}}  = {n_j}/m_j$.  The empirical distribution $P_j=[q_{j,1},\dots,q_{j,d}]$ across different classes can be calculated,
\begin{equation}
q_{j,i} = \left\{ \begin{array}{l}
R + {q_j}(1 - R),~~\text{for }i = j\\
{q_i}(1 - R),~~\text{for }i \ne j.
\end{array} \right.
\label{eq:Qij}
\end{equation}
One can see that, for the two users whose \emph{main} classes are $j$ and $k$ ($j \ne k$), we have ${\left\| {{P_j} - {P_k}} \right\|_1} = 2R$. 
In this case, based on Definition~\ref{def:R}, the total distance  is $R$.

\subsection{Optimizer} 
\label{sec:optimizer}

For all our computations, the optimizer we use is SGD with momentum. 
For the learning rate schedule, we use the cosine learning rate decay~\cite{cosine}, shown in the~\eref{lr} below, which is a commonly used schedule in SSL \cite{Fixmatch}:
\begin{equation}
\small 
{\gamma _t} = \gamma  \times \max \left\{ {\cos \left( {\pi  \times c \times \frac{{t - eM/B}}{{EM/B - eM/B}}}, \right),\varepsilon } \right\},
\label{lr}
\end{equation}
where $\gamma$ is the base learning rate,
$c$ is the periodic coefficient,
$E$ is the number of training epochs, 
$B$ is the batch size,
$t$ is the current iteration, 
$M$ is the number of training samples used in one epoch, 
$e$ is the number of epochs for warmup~\cite{warmup}, 
and $\varepsilon$ is a small constant. 
The hyperparameters used for different datasets can be found in~\tref{Optimizer}.

We believe that if one further tunes those hyperparameters, the performance of our grouping-based methods can improve further. 

\begin{table*}
\caption{
Optimizer hyperparameters used on different datasets. }

  \centering
  \begin{tabular}{lccccccccc}
\toprule
  Dataset & $E$ & $M$ & $\gamma$ & $e$ & $\varepsilon$ & weight decay & momentum & $c$ & $B$\\
\midrule
  Cifar-10 & 300 & 65536 & 0.146 & 5 & 1e-4 & 1e-4 & 0.9  & 2.3 & 64\\

 \hc SVHN & 40 & 65536 & 0.146 & 5 & 1e-4 & 1e-4 & 0.9 & 2.3  & 64\\

  EMNIST & 100 & 65536 & 0.03 & 0 & 1e-4 & 1e-4 & 0.9 & 0.4375 & 64\\
\bottomrule
  \end{tabular}
  \label{Optimizer}
\end{table*}

\subsection{Computing infrastructure} 
All experiments use NVIDIA GPU servers as computing nodes. Each server contains 8 TITAN V GPUs and the servers are internetworked via commodity 1 Gbps Ethernet. In our experiments, we set the random seed to 2019 to partition the training data according to~\sref{subsec:DataDistribute} across all nodes. The experiments are implemented in PyTorch and Gloo communication backend, and we generate the weights of the neural networks with random seed 1. Thus, we can simulate broadcasting the same initialized machine learning model to all participants, and the results are reproducible.

\subsection{Experiment parameters} 

We include all the environmental parameters used in \sref{sec:Results}. See \tref{tab:ExperimentParameters}. 
We have defined these parameters in \sref{sec:other_factors}. They include the non-iidness $R$, the communication period $T$, the number of labeled data $N_s$ held by the server, the user number $K$, the number of participating users $C$, and the number of groups $S$ for our grouping-based method.
We have also included the experiment title, the dataset used in the experiment, and the FL method that we test.

\begin{table*}
\begin{minipage}[t]{1.0\linewidth}

    \caption{
    Parameter settings of empirical experiments in~\sref{sec:Results}. 
    Note that R represents ``Non-iidness'', T represnets ``Communication Period'', $M_s$ represents ``server data number'', C represents ``Communicated users'', K represents ``User number'', and $S$ represents ``Group number''.
    }
    \centering
    \begin{adjustbox}{width=0.995\textwidth,center} 
    \setlength\tabcolsep{3.pt}
    \begin{tabular}[t]{lccccccccccccccccccc}
    \toprule
 Experiment title & Row-id & Dataset & Method & $R$ &  $T$ &  $N_s$ &  $C$ &  $K$ &   $S$\\
  \midrule
  \hc \multirow{3}*{\makecell[c]{Impact of $R$\\~\fref{fig:impact_r_ns_cp} (left)}} & 1 & Cifar10 & Grouping-based & $\{0.0,0.1,\cdots,1.0\}$ & 16 & $10^3$ &  $10$ & $10$ & $-$\\
  & 2 & SVHN & Grouping-based & $\{0.0,0.1,\cdots,1.0\}$ & 16 & $10^3$ &  $10$ & $10$ & $-$\\
  & 3  & EMNIST & Grouping-based & $\{0.0,0.1,\cdots,1.0\}$ & 16 & $4.7\times10^3$ &  $10$ & $47$ & $-$\\
    \midrule
  \hc \multirow{3}*{\makecell[c]{Impact of $T$\\~\fref{fig:impact_r_ns_cp} (middle)}} & 4 & Cifar10 & Grouping-based & $0.4$ & $\{2,2^2,\cdots,2^5\}$ & $10^3$ &  $10$ & $10$ & $-$\\
 & 5 & SVHN & Grouping-based & $0.4$ & $\{2,2^2,\cdots,2^5\}$ & $10^3$ &  $10$ & $10$ & $-$\\
 & 6  & EMNIST & Grouping-based & $0.4$ & $\{2,2^2,\cdots,2^5\}$ & $4.7\times10^3$ &  $10$ & $47$ & $-$\\
    \midrule
   \hc \multirow{3}*{\makecell[c]{Impact of $N_s$\\~\fref{fig:impact_r_ns_cp} (right)}} & 7 & Cifar10 & Grouping-based & $0.4$ & $16$ & $\{10^3,2\times10^3,\cdots,5\times10^3\}$ &  $10$ & $10$ & $-$\\
 & 8 & SVHN & Grouping-based & $0.4$ & $16$ & $\{10^3,2\times10^3,\cdots,5\times10^3\}$ &  $10$ & $10$ & $-$\\
 & 9 & EMNIST & Grouping-based & $0.4$ & $16$ &  $\{10^3,2\times10^3,\cdots,5\times10^3\}$ & $10$ & $47$ & $-$\\
    \midrule
    \hc \multirow{2}*{\makecell[c]{Impact of $C$\\~\tref{tab:CommunicationVolumeResults_on_cifar_svhn} and~\tref{tab:CommunicationVolumeResults_on_emnist}}} & 10  & Cifar10 & Grouping-based & $0.4$ & $16$ & $10^3$ &  $10$ & $10$ & $-$\\
 & 11 & Cifar10 & Grouping-based & $0.4$ & $16$ & $10^3$ &  $\{10,20\}$ & $20$ & $-$\\
 & 12 & Cifar10 & Grouping-based & $0.4$ & $16$ & $10^3$ &  $\{10,30\}$ & $30$  & $-$\\
 & 13  & SVHN & Grouping-based & $0.4$ & $16$ & $10^3$ &  $10$ & $10$ & $-$\\
 & 14  & SVHN & Grouping-based & $0.4$ & $16$ & $10^3$ &  $\{10,20\}$ & $20$ & $-$\\
 & 15  & SVHN & Grouping-based & $0.4$ & $16$ & $10^3$ &  $\{10,30\}$ & $30$ & $-$\\
 & 16  & EMNIST & Grouping-based & $0.4$ & $16$ &  $4.7\times10^3$ & $\{10,30,47\}$ & $47$ & $-$\\
    \midrule
   \hc \multirow{2}*{\makecell[c]{Impact of $K$\\~\tref{tab:CommunicationVolumeResults_on_cifar_svhn}}} & 17 & Cifar10 & Grouping-based & $0.4$ & $16$ & $10^3$ &  $10$ & $\{10,20,30\}$ & $-$\\
 & 18  & SVHN & Grouping-based & $0.4$ & $16$ & $10^3$ &  $10$ & $\{10,20,30\}$ & $-$\\
  \midrule
   \hc \multirow{3}*{\makecell[c]{FedAvg vs.\\ Grouping-based\\~\tref{fig:Groupingmethod}}}& 19 & Cifar10 & FedAvg/Grouping-based & $0.4$ & $16$ & $10^3$ &  $10$ & $10$ & $-$/$2$\\
   & 20 & SVHN & FedAvg/Grouping-based & $0.4$ & $16$ & $10^3$ &  $20$ & $20$ & $-$/$2$ \\
  & 21 & EMNIST & FedAvg/Grouping-based & $0.4$ & $16$ & $4.7\times10^3$ &  $47$ & $47$ & $-$/$5$ \\
  \midrule
   \hc \multirow{4}*{\makecell[c]{Comparison with \\ supervised
FL\\~\tref{ComparisonwithsupervisedFL}}} & 22 & Cifar10 & Supervised FedAvg & $0.29$ & $32$ & $-$ &  $10$ & $10$ & $-$\\
   & 23 & Cifar10 & DataSharing & $0.29$ & $32$ & $-$ &  $10$ & $10$ & $-$\\
  & 24 & Cifar10 & FedAvg & $0.4$ & $32$ & $10^3$ &  $10$ & $10$ & $-$\\
  & 25 & Cifar10 & Grouping-based & $0.4$ & $32$ & $10^3$ &  $10$ & $10$ & $2$\\
  \midrule
   \hc \multirow{4}*{\makecell[c]{Comparison with \\EASGD and \\OverlapSGD\\~\tref{SupervisedResultsOverlapSGD}}} & 26 & Cifar10 & EASGD & $0.4$ & $\{2,8\}$ & $-$ &  $16$ & $16$ & $-$\\
   & 27 & Cifar10 & OverlapSGD & $0.4$ & $\{2,8\}$ & $-$ &  $16$ & $16$ & $-$\\
  & 28 & Cifar10 & FedAvg & $0.4$ & $\{2,8,32\}$ & $10^3$ &  $16$ & $16$ & $-$\\
  & 29 & Cifar10 & Grouping-based & $0.4$ & $\{2,8,32\}$ & $10^3$ &  $16$ & $16$ & $2$\\
  \midrule
   \hc \multirow{2}*{\makecell[c]{Impact of the ratio $\eta$\\~\tref{tab:DifferentProportions}}}& 30 & Cifar10 & FedAvg/Grouping-based & $0.4$ & $16$ & $10^3$ &  $\{3,6,10,30\}$ & $30$ & $-$\\
   & 31 & SVHN & FedAvg/Grouping-based & $0.4$ & $16$ & $10^3$ &  $\{3,6,10,30\}$ & $30$& $-$ \\
   \midrule
   \hc \multirow{3}*{\makecell[c]{Fully supervised \\FL~\tref{SFLComparison}}}& 32 & EMNIST & Grouping-based & $0.4$ & $16$ & $4.7\times10^3$ &  $10$ & $10$ & $2$\\
   & 33 & EMNIST & Grouping-based & $0.4$ & $16$ & $4.7\times10^3$ &  $20$ & $20$ & $2$\\
   & 34 & EMNIST & Grouping-based & $0.4$ & $16$ & $4.7\times10^3$ &  $47$ & $47$ & $5$\\
   \midrule
   \hc \multirow{3}*{\makecell[c]{Grouping-based vs. \\FixMatch~\tref{FixMatchComparison}}}& 35 & Cifar10 & Grouping-based & $0.4$ & $16$ & $4\times10^3$ &  $10$ & $10$ & $2$\\
   & 36 & SVHN & Grouping-based & $0.4$ & $16$ & $10^3$ &  $20$ & $20$ & $2$\\
   & 37 & EMNIST & Grouping-based & $0.4$ & $16$ & $4.7\times10^3$ &  $47$ & $47$ & $5$\\
   & 38 & Cifar10 & FixMatch & $0.0$ & $1$ & $4\times10^3$ &  $1$ & $1$ & $-$\\
   & 39 & SVHN & FixMatch & $0.0$ & $1$ & $10^3$ &  $1$ & $1$ & $-$\\
   & 40 & EMNIST & FixMatch & $0.0$ & $1$ & $4.7\times10^3$ &  $1$ & $1$ & $-$\\
   \midrule
   \hc \multirow{1}*{\makecell[c]{Users have labeled samples~\tref{UersSideSemi}}}& 
   41 & EMNIST & FedAvg & $\{0.4,0.6\}$ & $16$ & $4.7\times10^3$ &  $47$ & $47$ & $-$\\
    \midrule
   \hc \multirow{4}*{\makecell[c]{Performance on STL-10\\~\tref{STL10}}}& 
   42 &  STL-10 & Self-training & $0.0$ & $16$ & $10^3$ &  $2$ & $10$ & $-$\\
   & 43 &STL-10 & CRL with BN & $0.0$ & $16$ & $10^3$ &  $2$ & $10$ & $-$\\
   & 44 &STL-10 & FedAvg & $0.0$ & $16$ & $10^3$ &  $2$ & $10$ & $-$\\
   & 45  &STL-10 & Grouping-based & $0.0$ & $16$ & $10^3$ &  $2$ & $10$ & $2$\\
    \midrule
    \hc \multirow{3}*{\makecell[c]{Performance on EMNIST with \\large user number~\tref{EMNSITLargeUserNumber}}}& 
   46 & EMNIST & FedAvg/Grouping-based & $0.4$ & $16$ & $4.7\times10^3$ &  $10$ & $47$ & $-$/$2$\\
   & 47 &EMNIST & FedAvg/Grouping-based & $0.4$ & $16$ & $4.7\times10^3$ &  $20$ & $47$ & $-$/$2$\\
   & 48 &EMNIST & FedAvg/Grouping-based & $0.4$ & $16$ & $4.7\times10^3$ &  $30$ & $47$ & $-$/$2$\\
    \midrule
    \hc \multirow{2}*{\makecell[c]{Comparison with~\cite{jeong2020federated} on \\
    Cifar10 Tab. \ref{JeongCifar10}}} & 
  49 &  Cifar10 & FedMatch& $\{0,1\}$ & $100$ & $5\times10^3$ &  $5$ & $100$ & $-$\\
  & 50  & Cifar10 & Grouping-based & $\{0,1\}$ & $100$ & $5\times10^3$ &  $5$ & $100$ & $2$\\
    \bottomrule
    \end{tabular}
    \label{tab:ExperimentParameters}
    \end{adjustbox}
\end{minipage}\hfill
\end{table*}

\section{Convergence speed of different communication period $T$ on Cifar-10}
\label{sec:Converge_at_different_cp}

\begin{figure}
\centering
\resizebox{\figlen cm}{\figwidth cm}{\includegraphics{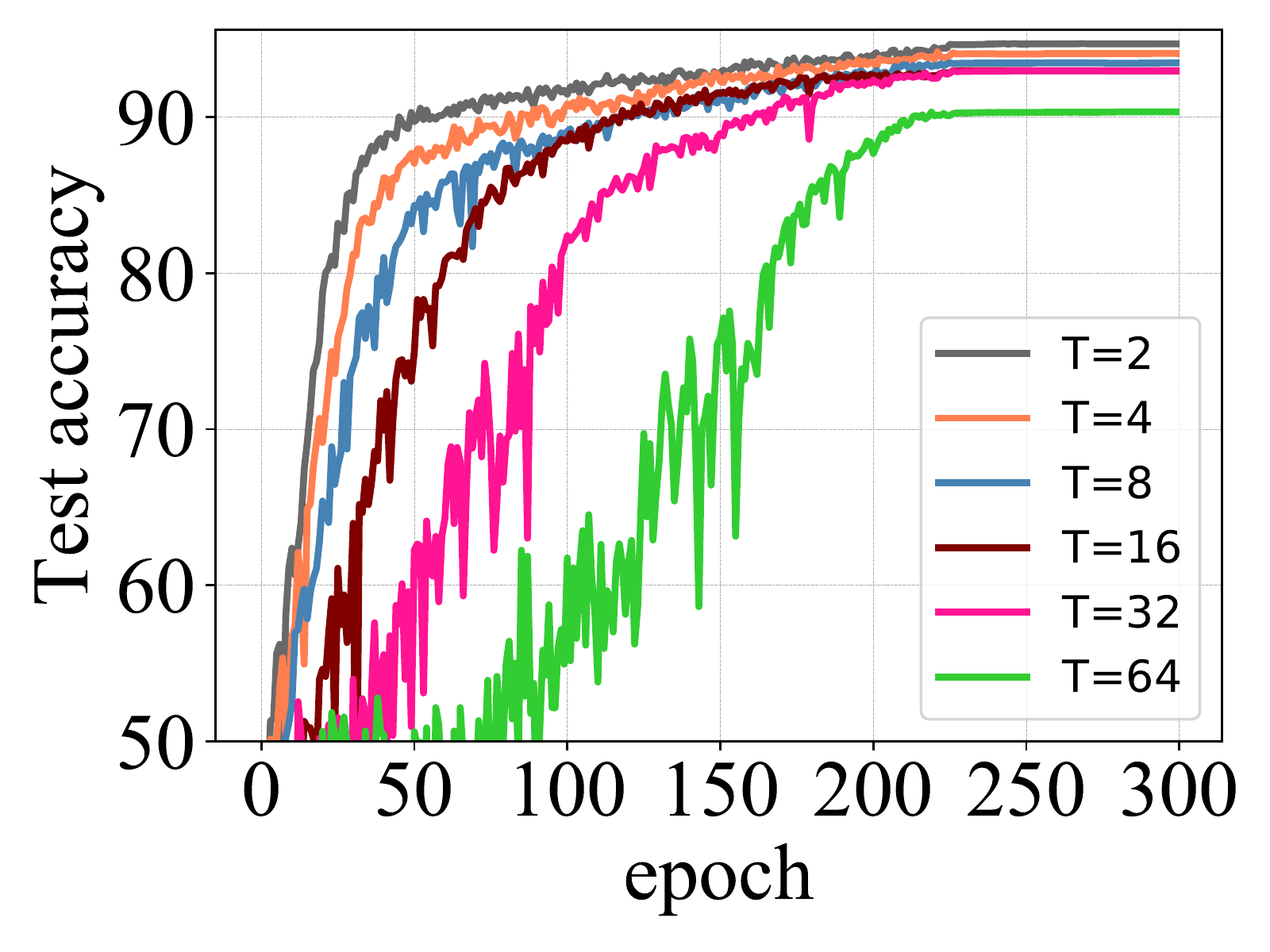}}
\caption{ 
Convergence curves of our group-based solution
of different
communication periods on Cifar-10.}
\label{fig:cpCifar10}
\end{figure}

In this section, we show the test accuracy curves of training with different communication period $T$ on Cifar-10, which provides more details to the experimental results discussed in~\sref{sec:impact_r_cp_ns}.
See~\fref{fig:cpCifar10}. 
The experimental settings are also presented in~\sref{sec:impact_r_cp_ns}.
From~\fref{fig:cpCifar10}, we can see that when $T$ is small, our grouping based method, which combines the CRL training objective and GN, converges quickly.

\section{The effect of training epochs on SVHN}
\label{sec:emnist_training_epoch_effect}

\begin{figure}
\centering
\resizebox{\figlen cm}{\figwidth cm}
{\includegraphics{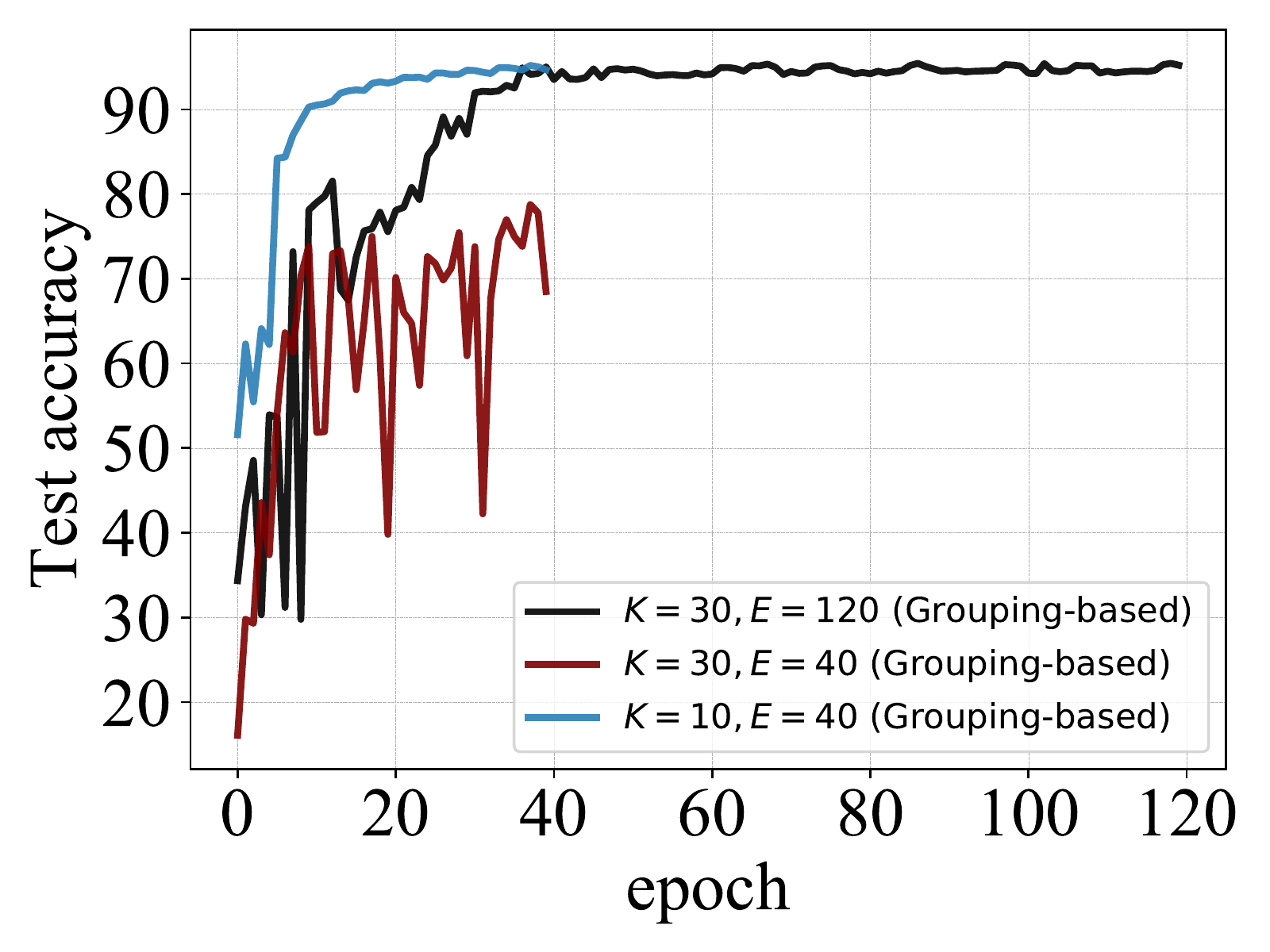}}
\caption{ 
Convergence curves of our grouping-based solution for different $K$ and $E$ on SVHN.}
\label{KSVHN}
\end{figure}

In this section, we study the effect of training epochs on SVHN when $K=C=30$, which provides more details to the experiments in~\tref{tab:CommunicationVolumeResults_on_cifar_svhn}, \sref{subs:FixCk}.
As shown in~\tref{tab:CommunicationVolumeResults_on_cifar_svhn}, the accuracy of our grouping-based method on SVHN with $K=C=30$ and $E=40$ is much lower than that of $K=C=10$ and $E=40$. 
Increasing the number of training epochs to $E=120$ can significantly improve the performance of $K=C=30$. 
The comparison is shown in~\fref{KSVHN}. 

\section{Ablation study on gradient diversity }
\label{sec:Weightdiversityexperimentson Cifar10}

In this section, we provide more ablation study, by measuring gradient diversity with multiple alternative definitions.
Recall that gradient diversity \eref{def:weight_diversity} measures the dissimilarity between concurrent gradient updates of different users. As we have discussed extensively in the main paper, reducing gradient diversity is the key to improving \ssfl methods. Thus, we study gradient diversity by selecting several different ways to compute.

First, we can remove the square operation in \eref{def:weight_diversity}, and we only use the $\ell_2$-norm to measure gradient diversity:
\begin{equation}
\small 
\Delta_1^t (w) = 
\sum\nolimits_{k \in \mathcal{C}_t} {\left\| \nabla w_k^t \right\|_2}/\left\| \sum\nolimits_{k \in \mathcal{C}_t} \nabla w_k^t  \right\|_2.
\label{graddiversityType2}
\end{equation}
Second, we can also replace the $\ell_2$-norm with the $\ell_1$-norm, which leads to the following two alternatives with/without the square operation:
\begin{equation}
\small 
\Delta_2^t (w) = 
\sum\nolimits_{k \in \mathcal{C}_t} {\left\| \nabla w_k^t \right\|_1^2}/\left\| \sum\nolimits_{k \in \mathcal{C}_t} \nabla w_k^t  \right\|_1^2.
\label{graddiversityType3}
\end{equation}

\begin{equation}
\small 
\Delta_3^t (w) = 
\sum\nolimits_{k \in \mathcal{C}_t} {\left\| \nabla w_k^t \right\|_1}/\left\| \sum\nolimits_{k \in \mathcal{C}_t} \nabla w_k^t  \right\|_1.
\label{graddiversityType4}
\end{equation}

Third, we can change the set $\mathcal{C}_t$ in the computation.
Note that in all of the definitions above, we calculated gradient diversity only using the gradients from the users. Therefore, the set $\mathcal{C}_t$ only contains users. However, we can also include the server gradient in the calculation of gradient diversity.

Fourth, we can change the way of computing each individual gradient. We notice that in FL, the local gradient updates are not aggregated directly. Instead, sequential gradient updates are applied to each user. Then, the updated weights from the users are averaged.
Thus, instead of calculating the diversity of gradient $\nabla w_k^t$ evaluated on all the user data, we can define $\nabla w_k^t$ as the difference between the model before and after local gradient updates, i.e. 
\begin{equation}
\nabla w_k^t=\left\{ \begin{array}{l}
w_k^t-w_{avg}^{t - 1}, \rm{for\ FedAvg,}\\
w_k^t-w_{avg,i}^{t - 1}, \rm{for\ \text{grouping-based}},
\end{array} \right.
\label{CumulativeGradient}
\end{equation}
where $i$ is the index of group to which user $k$ belongs; see \eref{eq:hierarchicalavg}. It can be seen that \eref{CumulativeGradient} is the cumulative change in weights after the local gradient updates. 
We can substitute the above-defined gradient \eref{CumulativeGradient} into \eref{def:weight_diversity} and \eref{graddiversityType2}-\eref{graddiversityType4} to calculate gradient diversity.

Thus, we can either perform the square operation or not, either use the $\ell_2$-norm or the $\ell_1$-norm, either include the server or not in $\mathcal{C}_t$, and either using the cumulative gradient updates \eref{CumulativeGradient} or not. 
In total, we have $2\times2\times 2\times2 =16$ different ways of measuring gradient diversity.

\begin{figure*}
    \centering
    \includegraphics[width=0.32\linewidth]{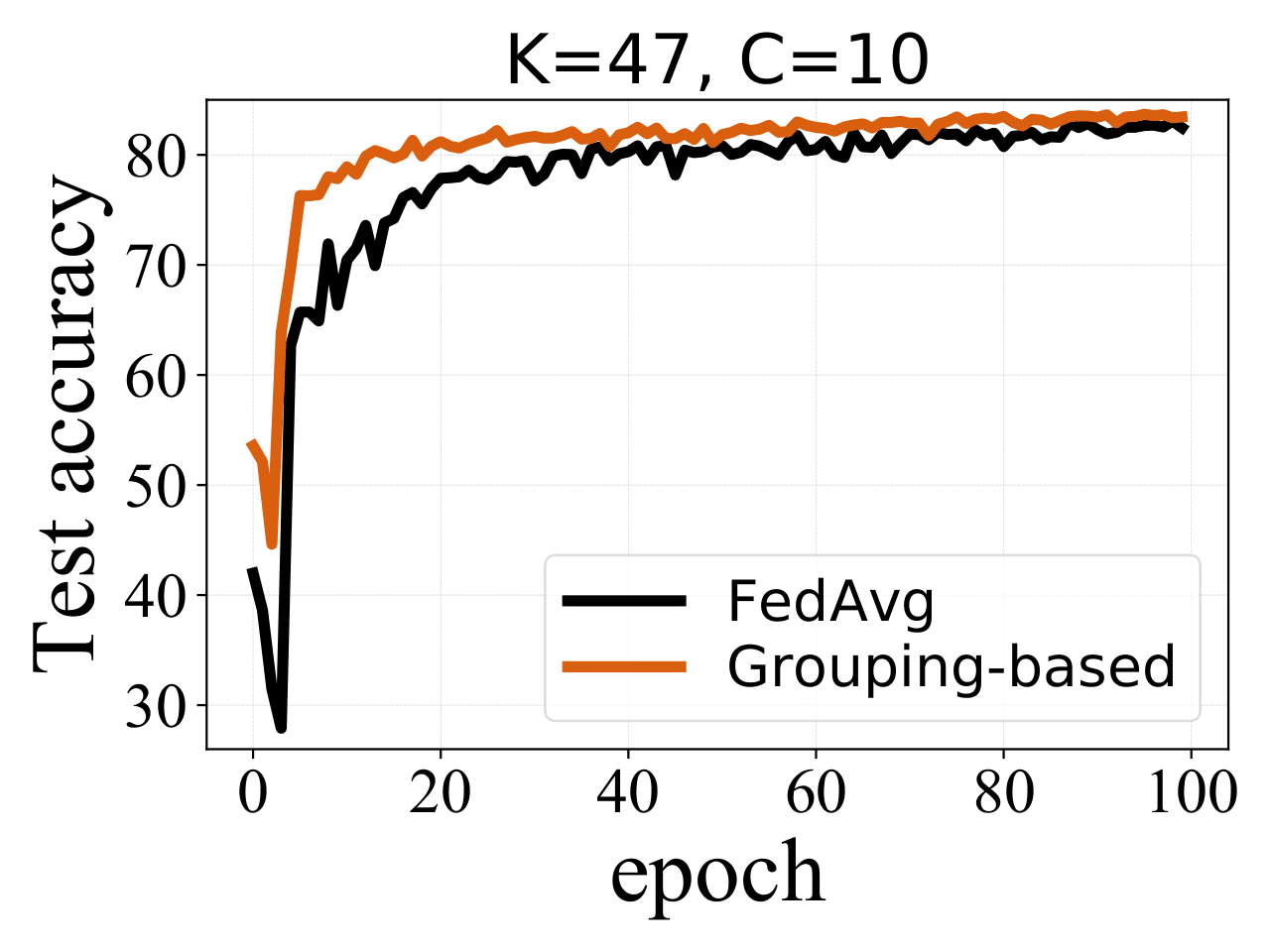}
    \includegraphics[width=0.32\linewidth]{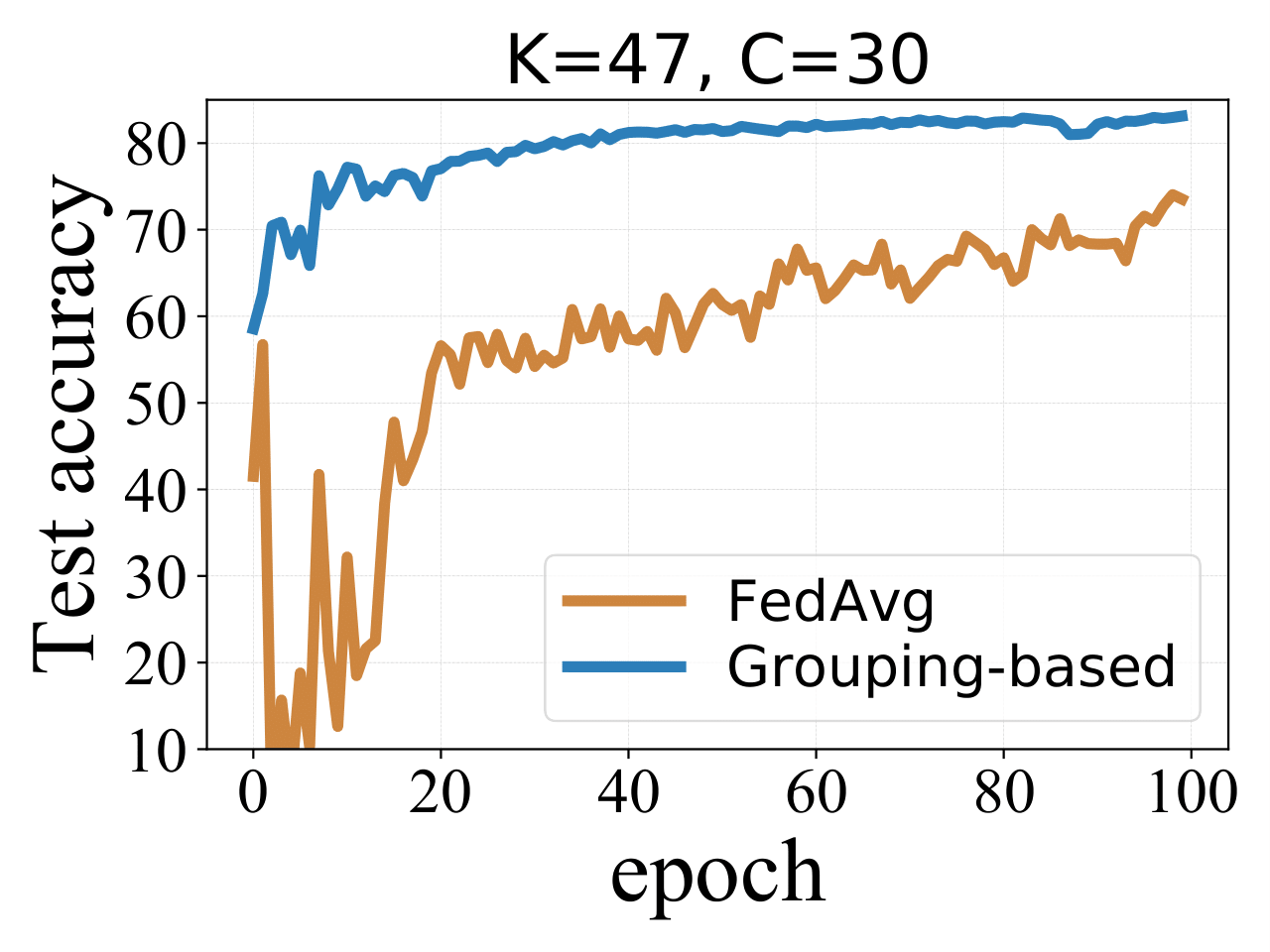}
    \includegraphics[width=0.32\linewidth]{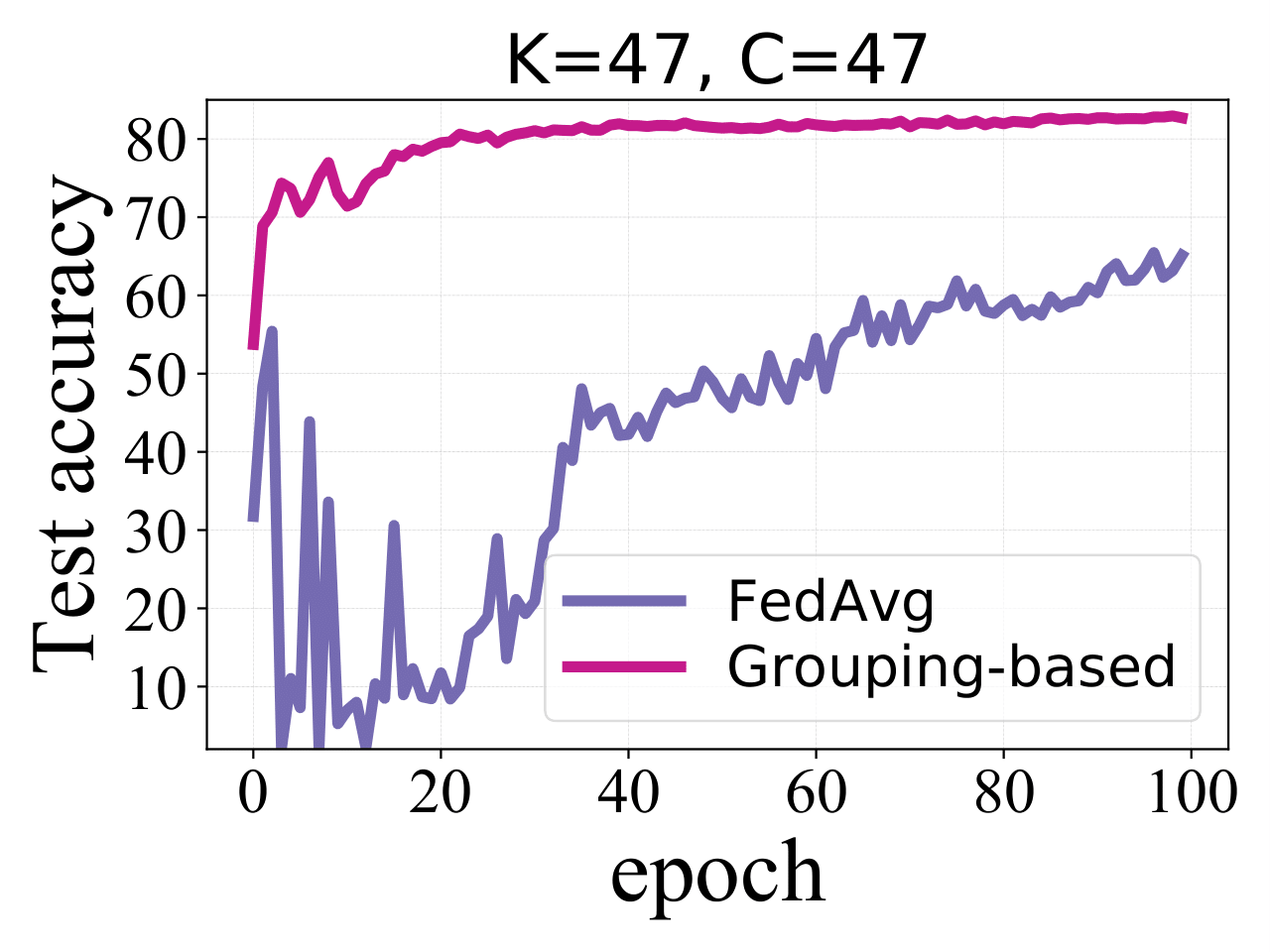}
    \includegraphics[width=0.32\linewidth]{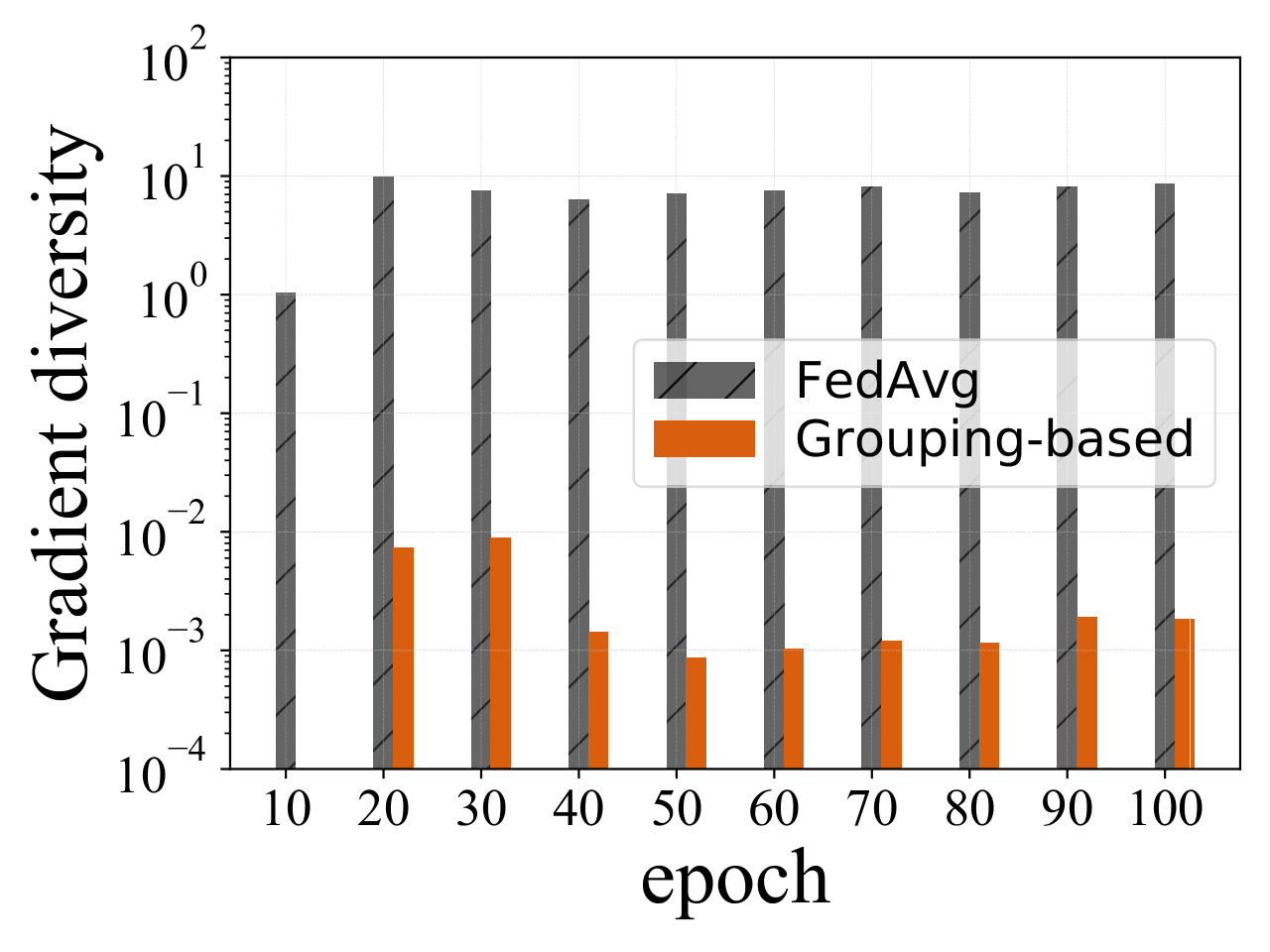}
    \includegraphics[width=0.32\linewidth]{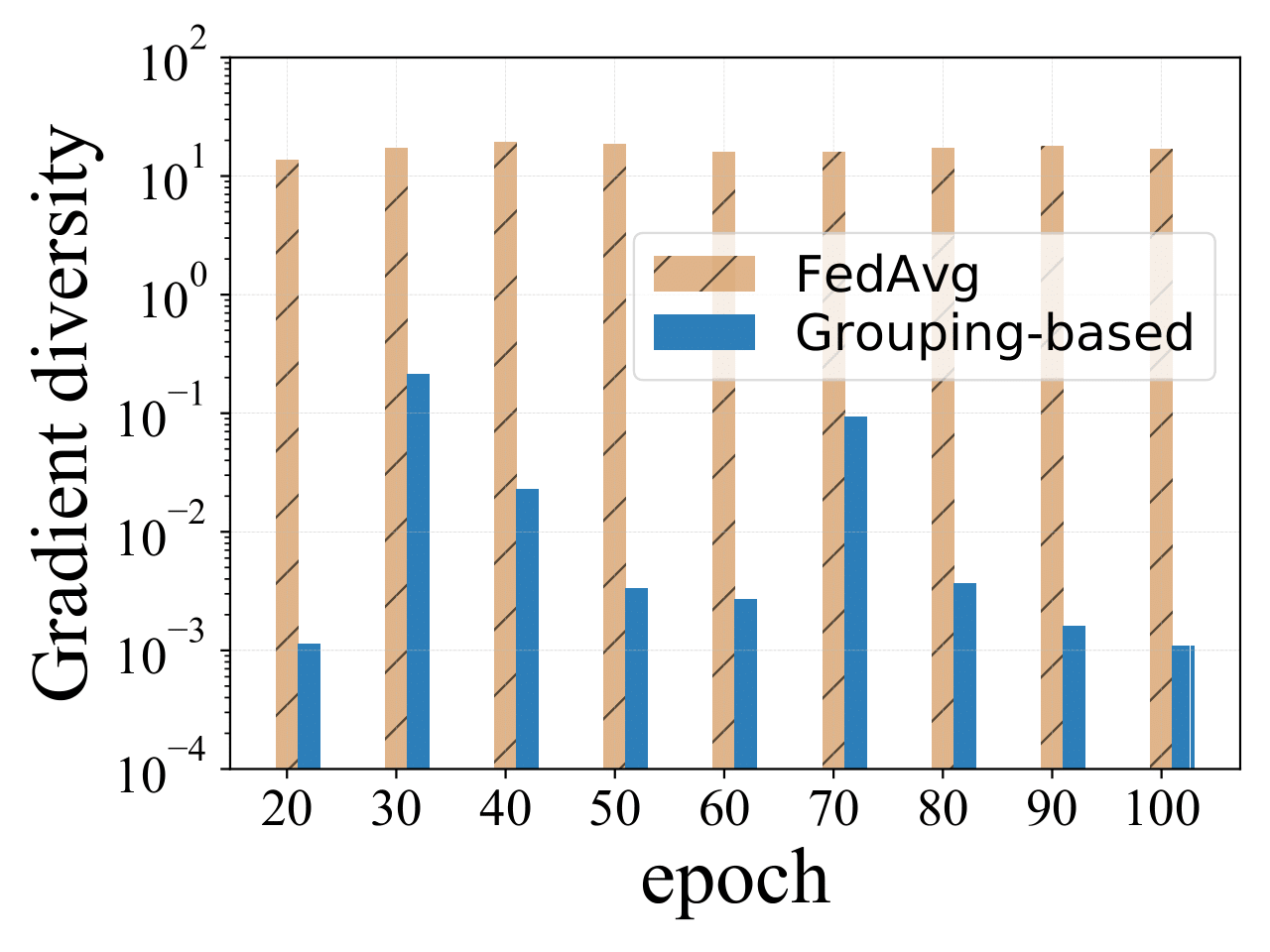}
    \includegraphics[width=0.32\linewidth]{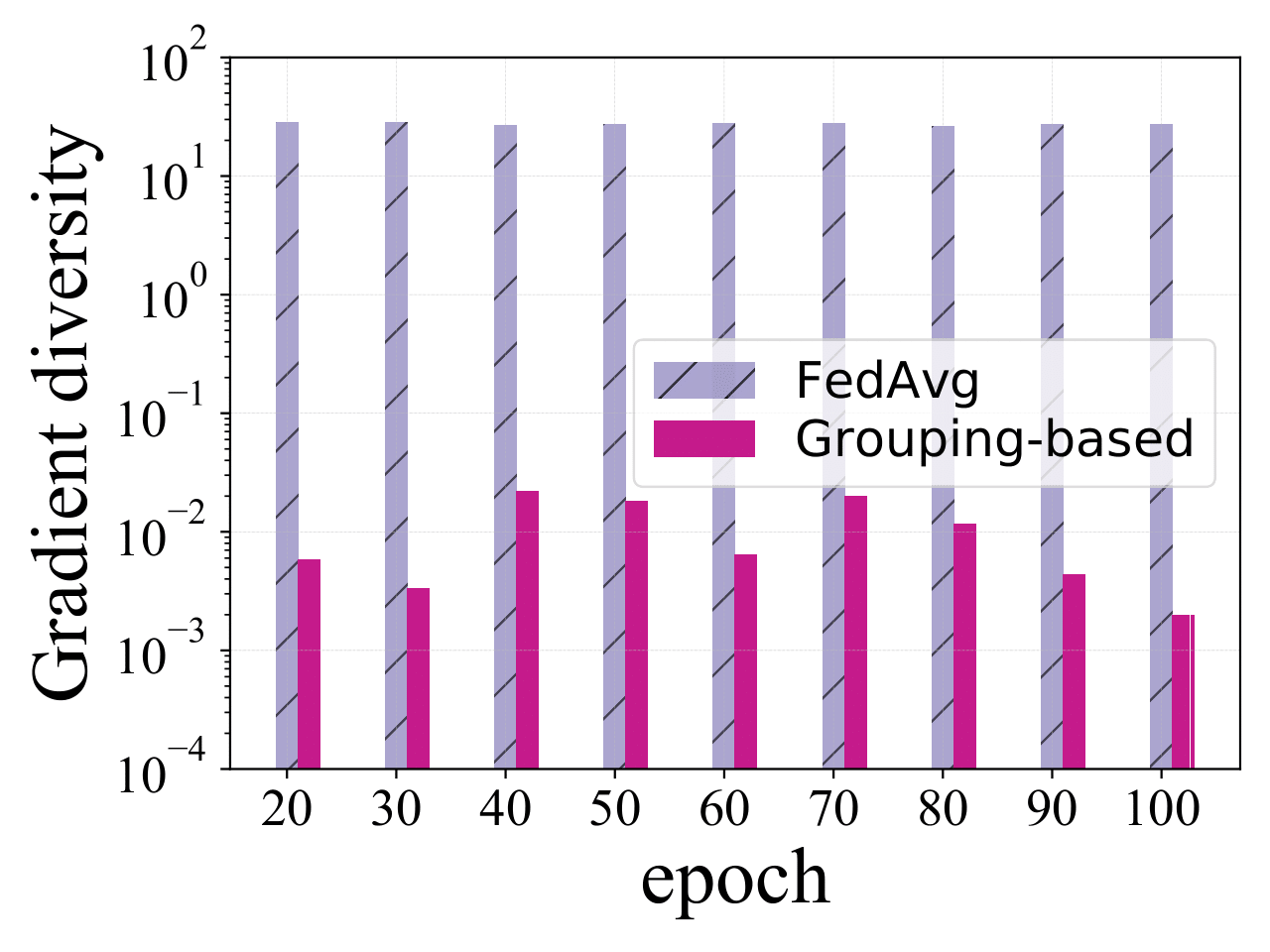}
    \includegraphics[width=0.32\linewidth]{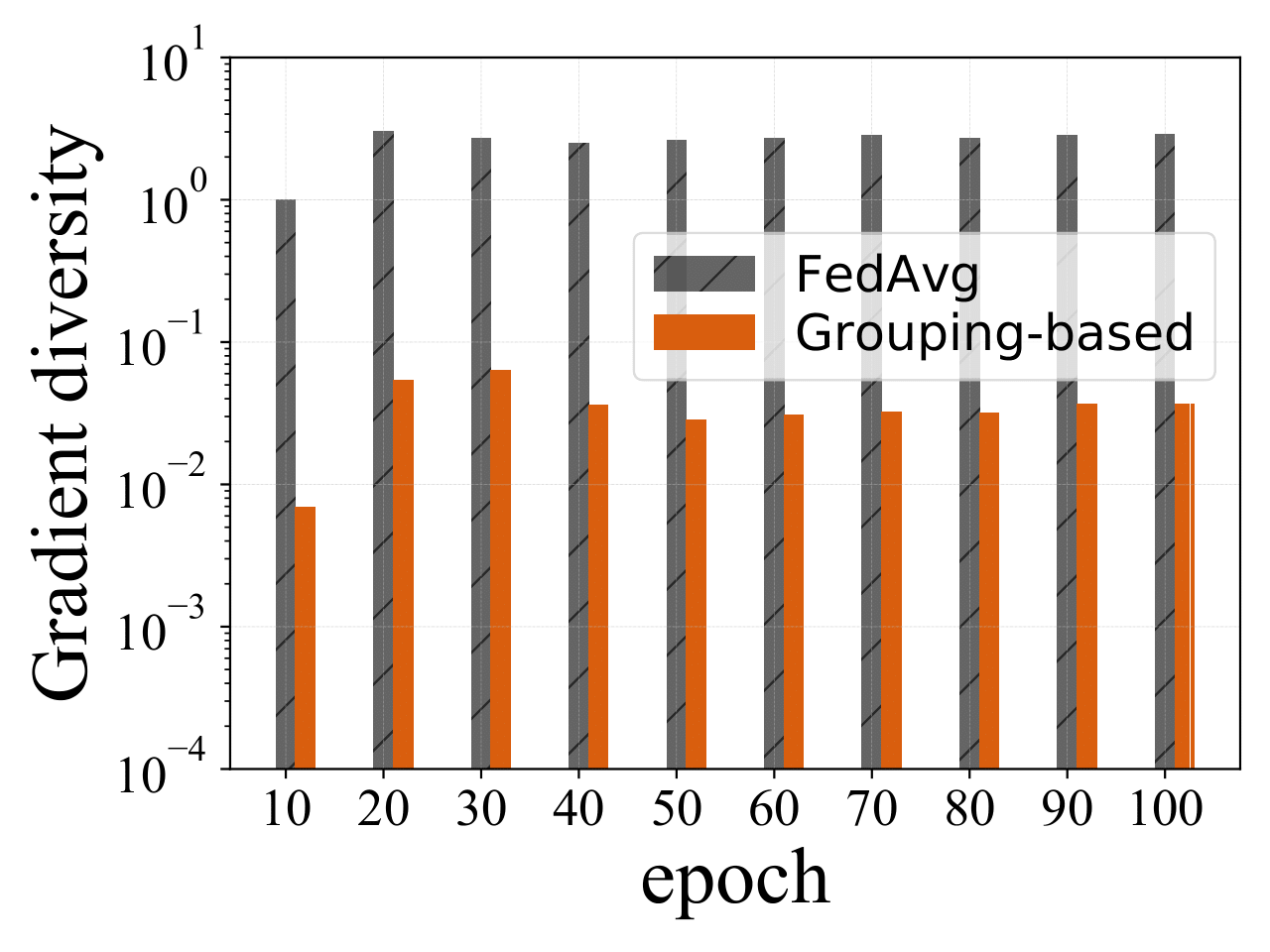}
    \includegraphics[width=0.32\linewidth]{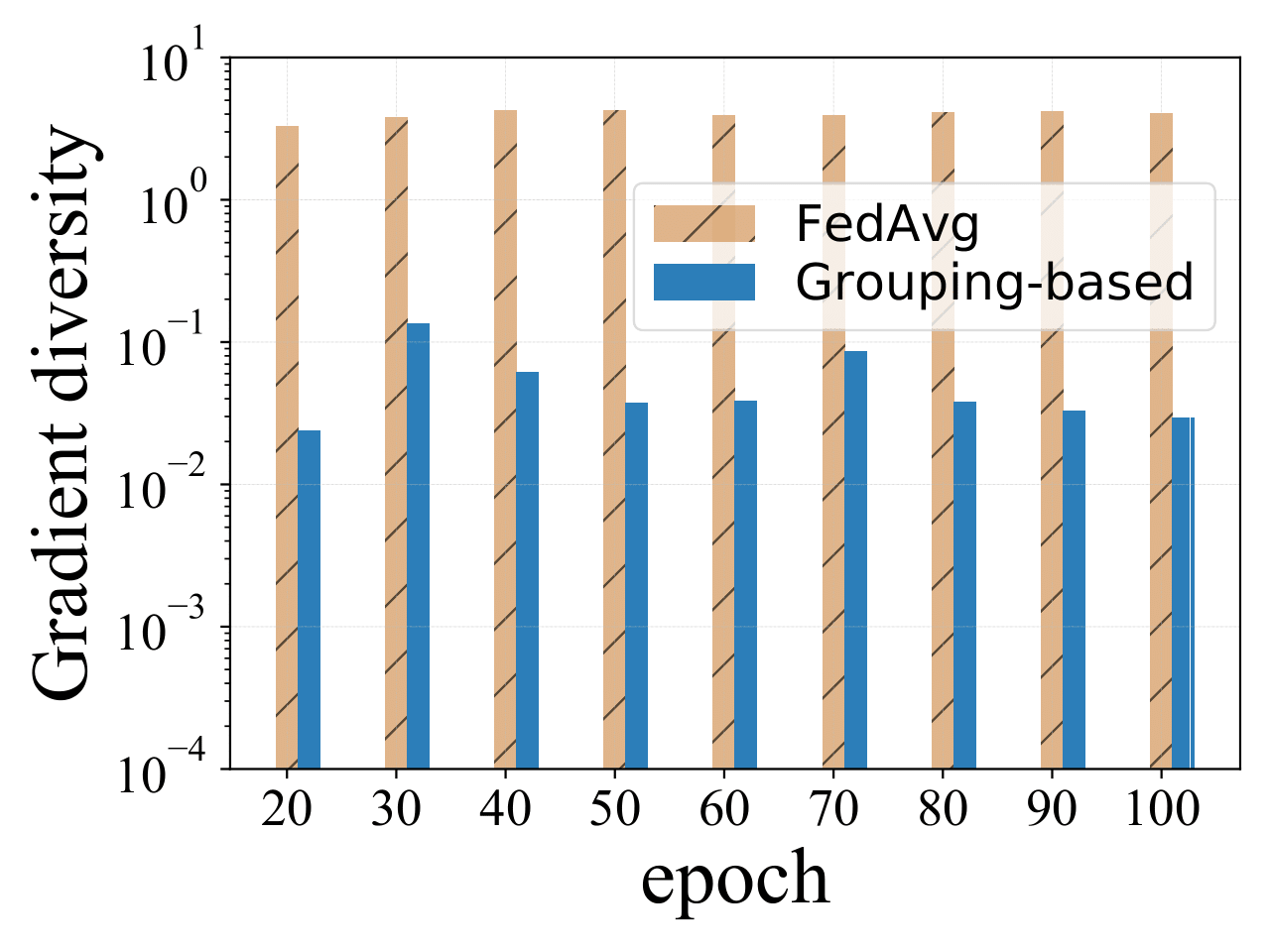}
    \includegraphics[width=0.32\linewidth]{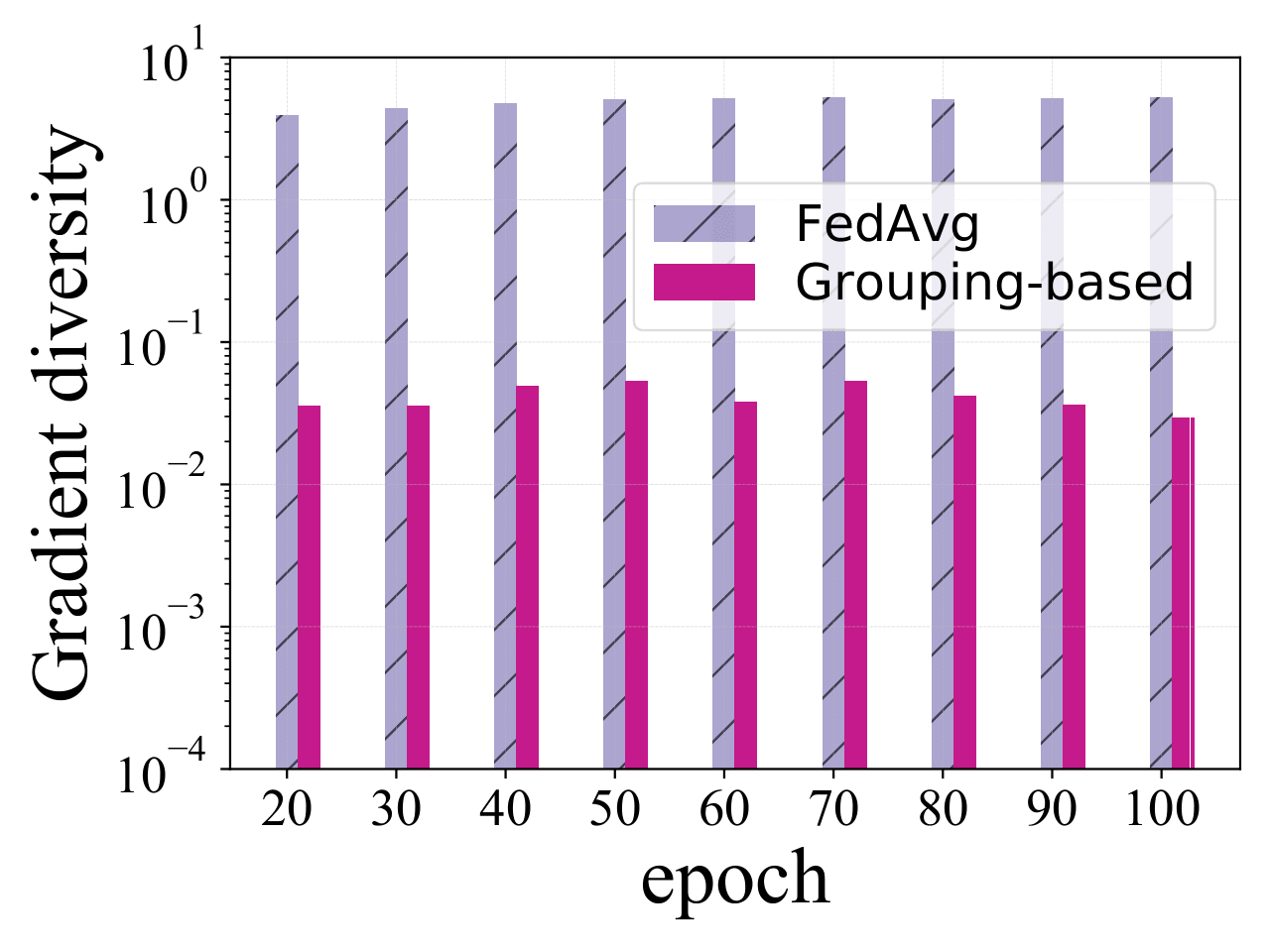}
    \includegraphics[width=0.32\linewidth]{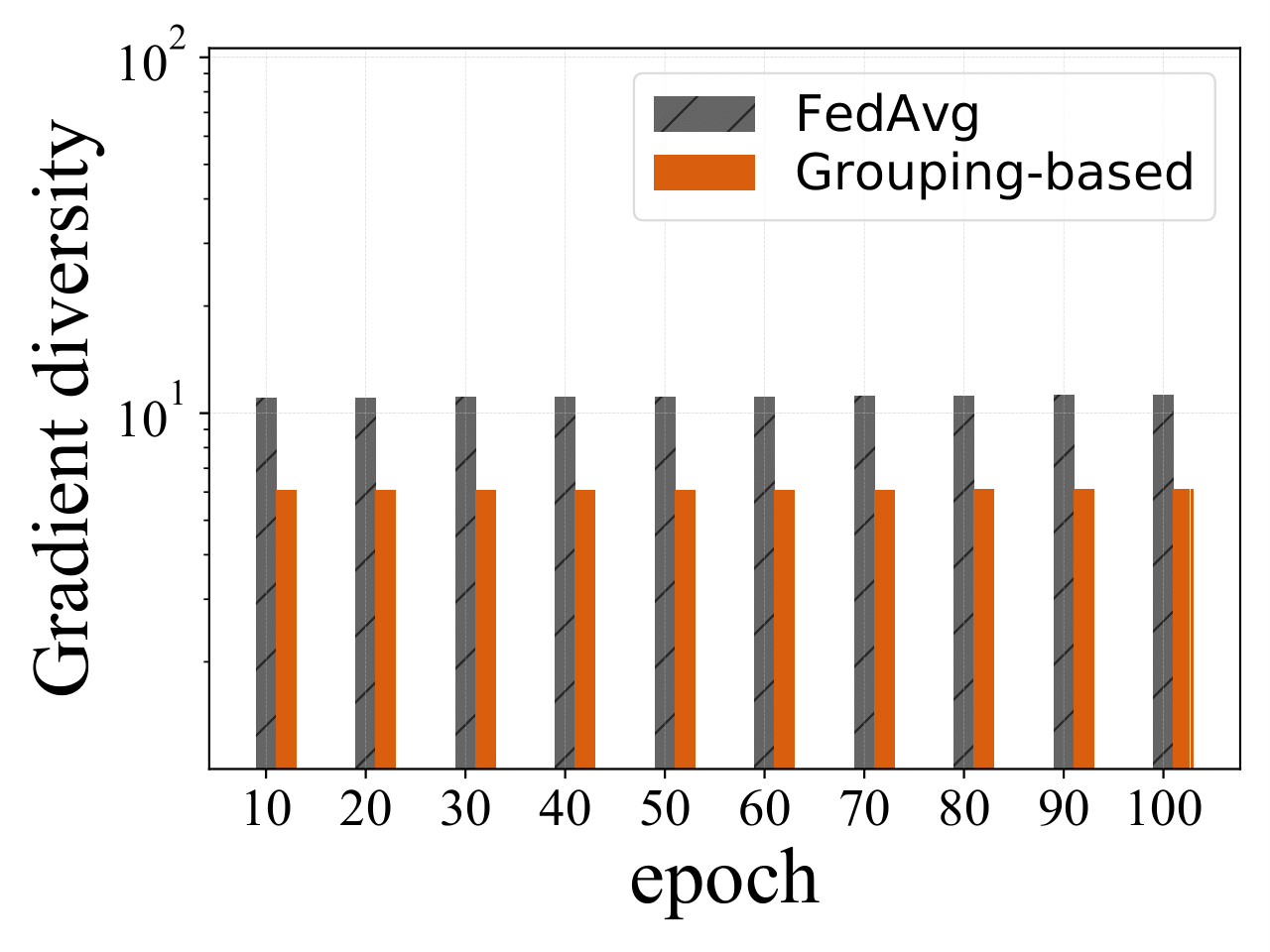}
    \includegraphics[width=0.32\linewidth]{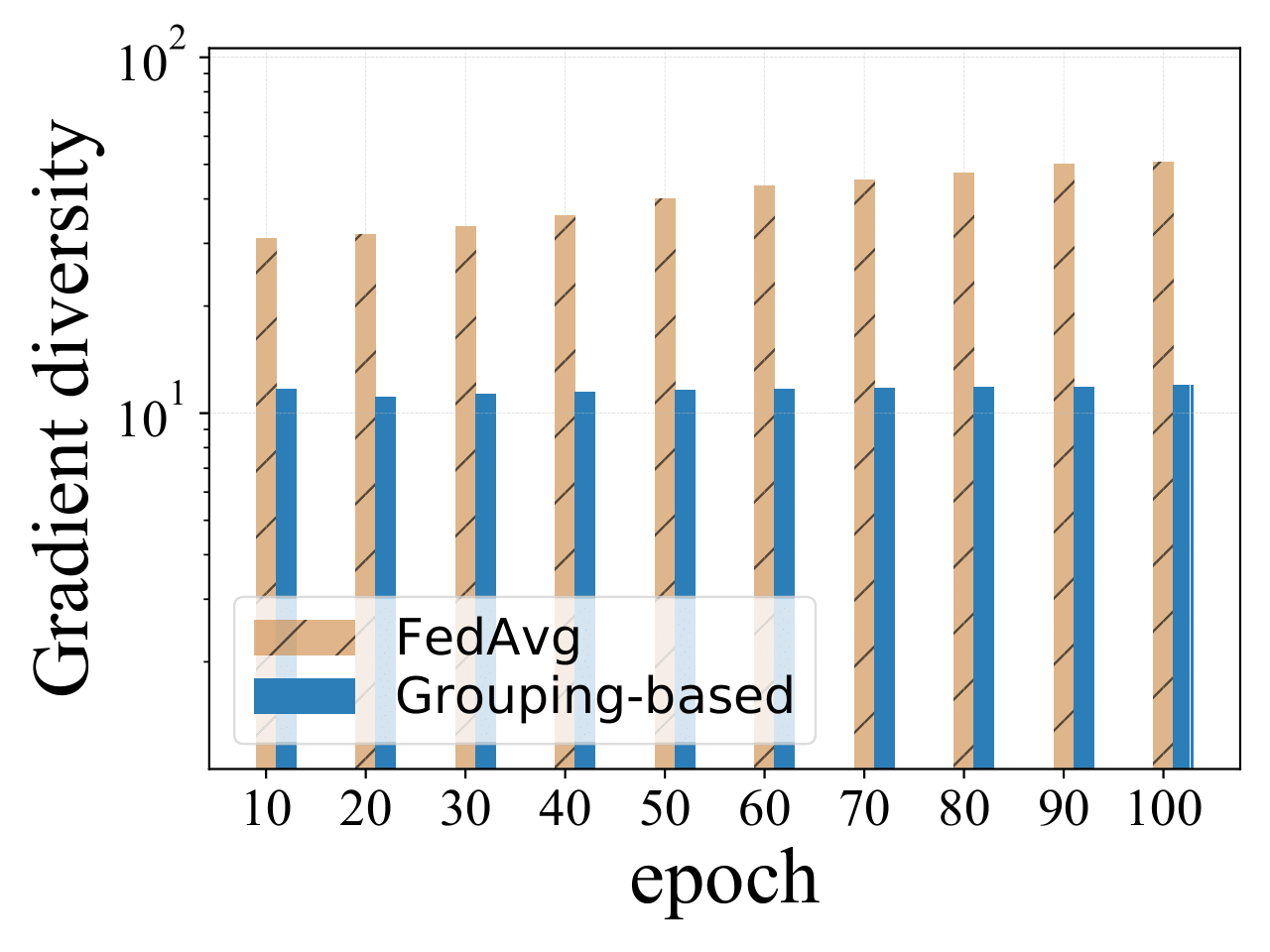}
    \includegraphics[width=0.32\linewidth]{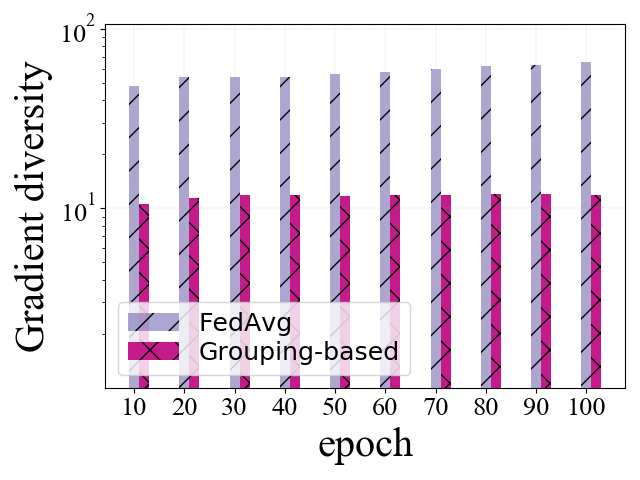}
    \includegraphics[width=0.32\linewidth]{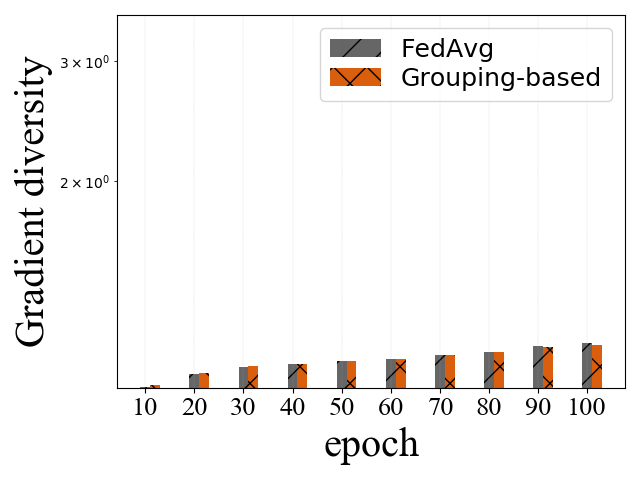}
    \includegraphics[width=0.32\linewidth]{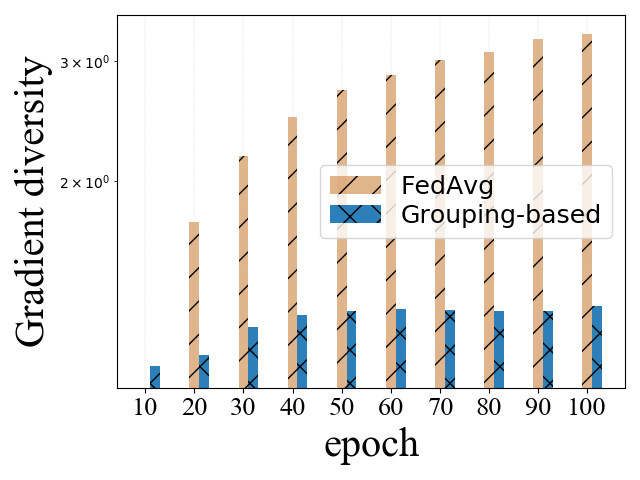}
    \includegraphics[width=0.32\linewidth]{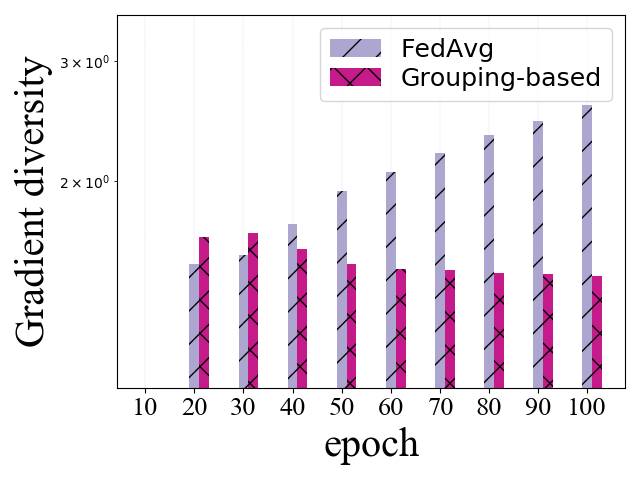}
    \caption{
    (Line 1) The convergence curves on EMNIST. (Line 2) Results on gradient diversity defined in \eref{def:weight_diversity}. 
    (Line 3) Results on gradient diversity defined in \eref{graddiversityType2}. 
    (Line 4) Results on gradient diversity defined in \eref{def:weight_diversity}, including both the users and the server. 
    (Line 5) Results on gradient diversity defined in \eref{graddiversityType2}, including both the users and the server. 
    }
    \label{fig:Groupingmethod}
\end{figure*}

\begin{figure*}
    \centering
    \includegraphics[width=0.32\linewidth]{figs/EMNIST_ACC_FedAvg_vs_Grouping_commUE10_deversity-1.png}
    \includegraphics[width=0.32\linewidth]{figs/EMNIST_ACC_FedAvg_vs_Grouping_commUE30_deversity-1.png}
    \includegraphics[width=0.32\linewidth]{figs/EMNIST_ACC_FedAvg_vs_Grouping_commUE47_deversity-1.png}
    \includegraphics[width=0.32\linewidth]{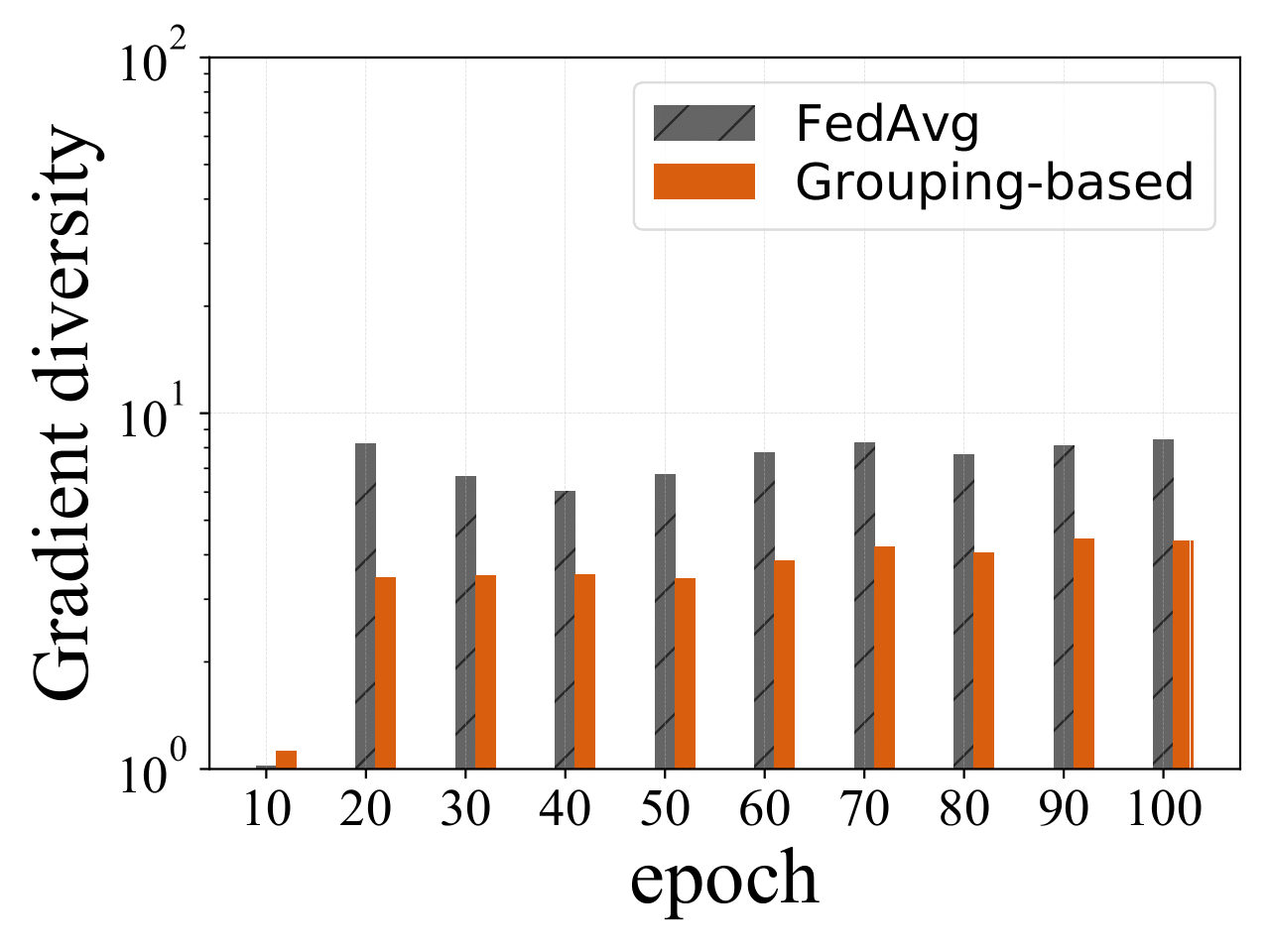}
    \includegraphics[width=0.32\linewidth]{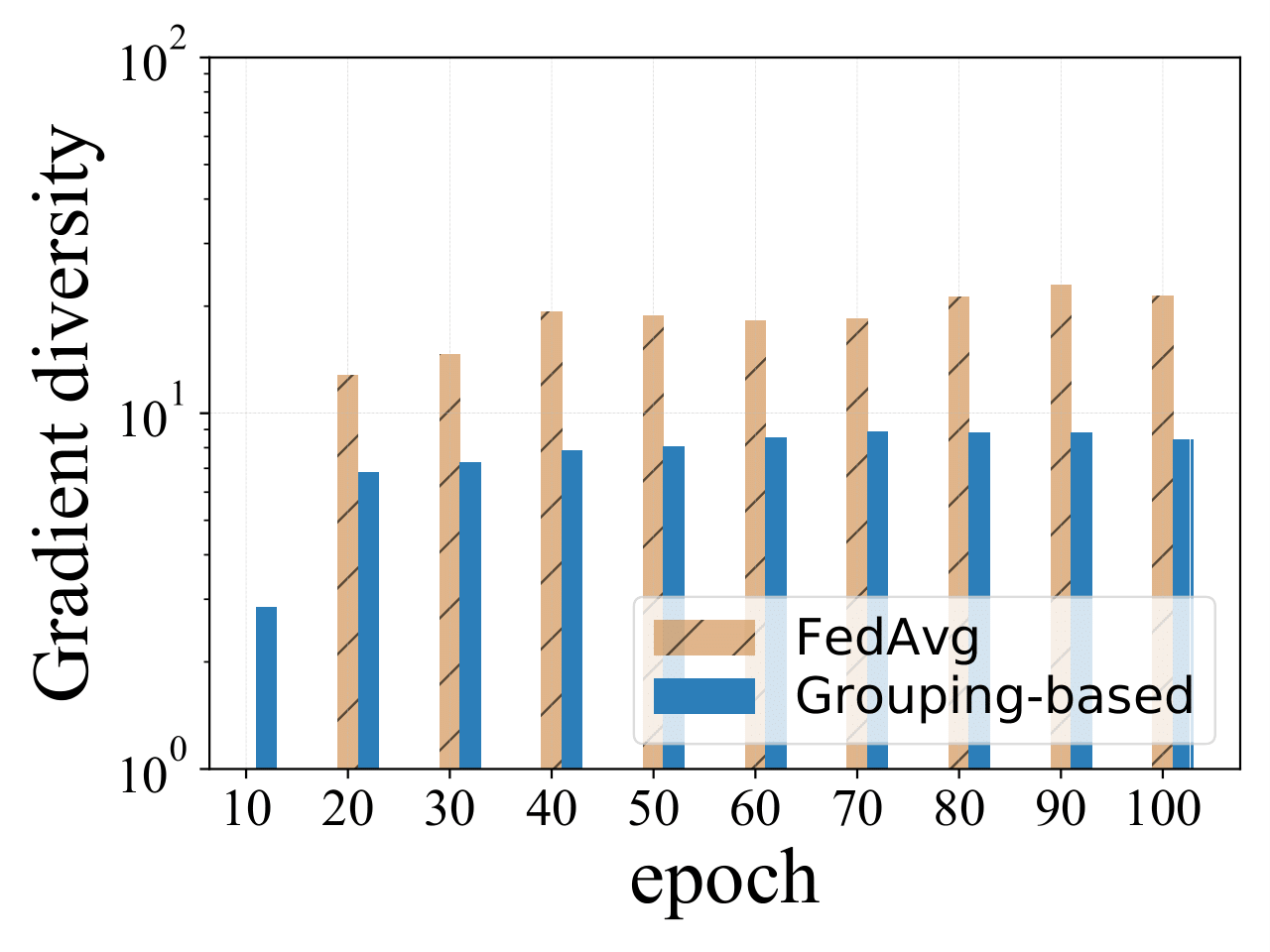}
    \includegraphics[width=0.32\linewidth]{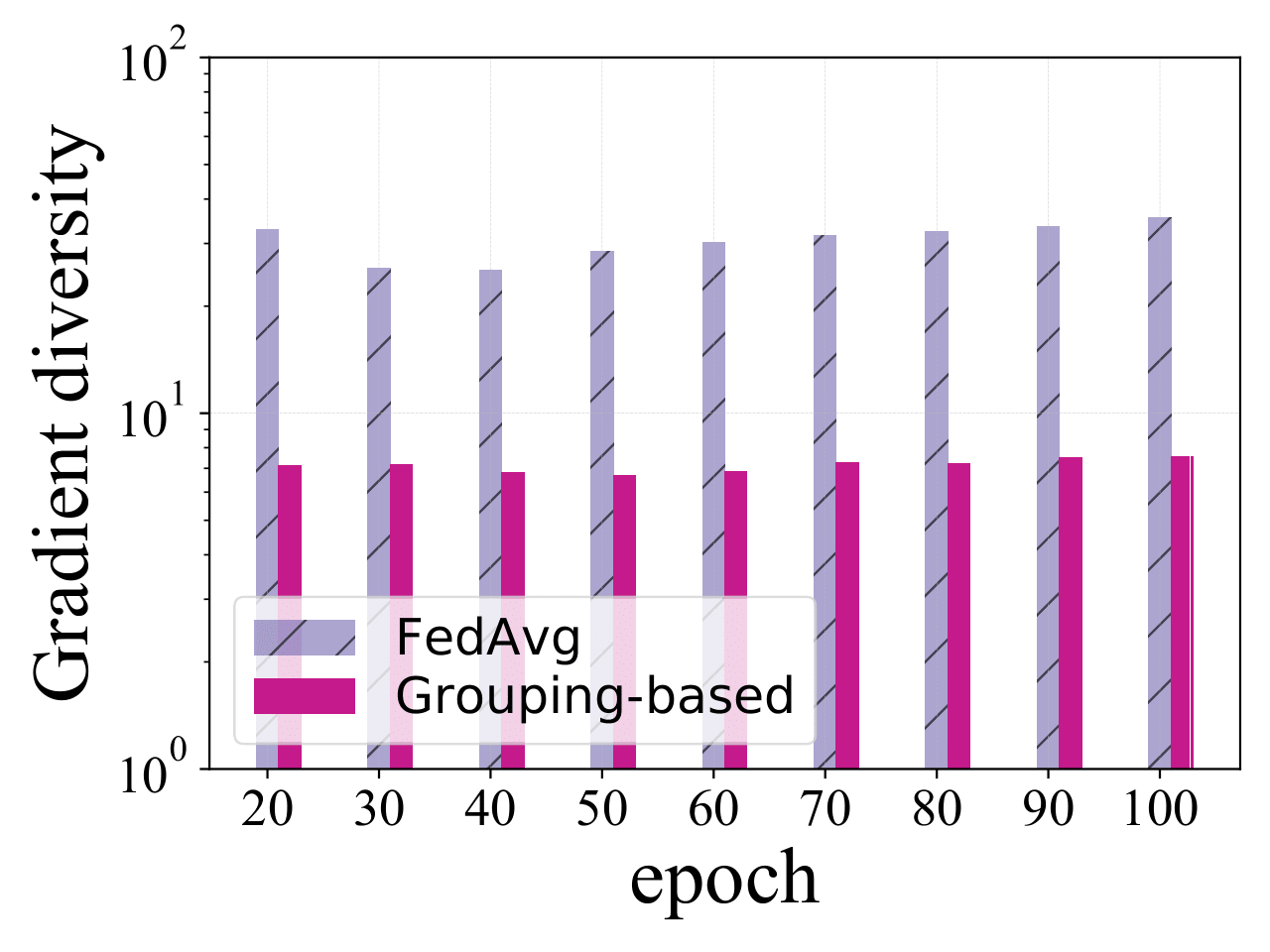}
    \includegraphics[width=0.32\linewidth]{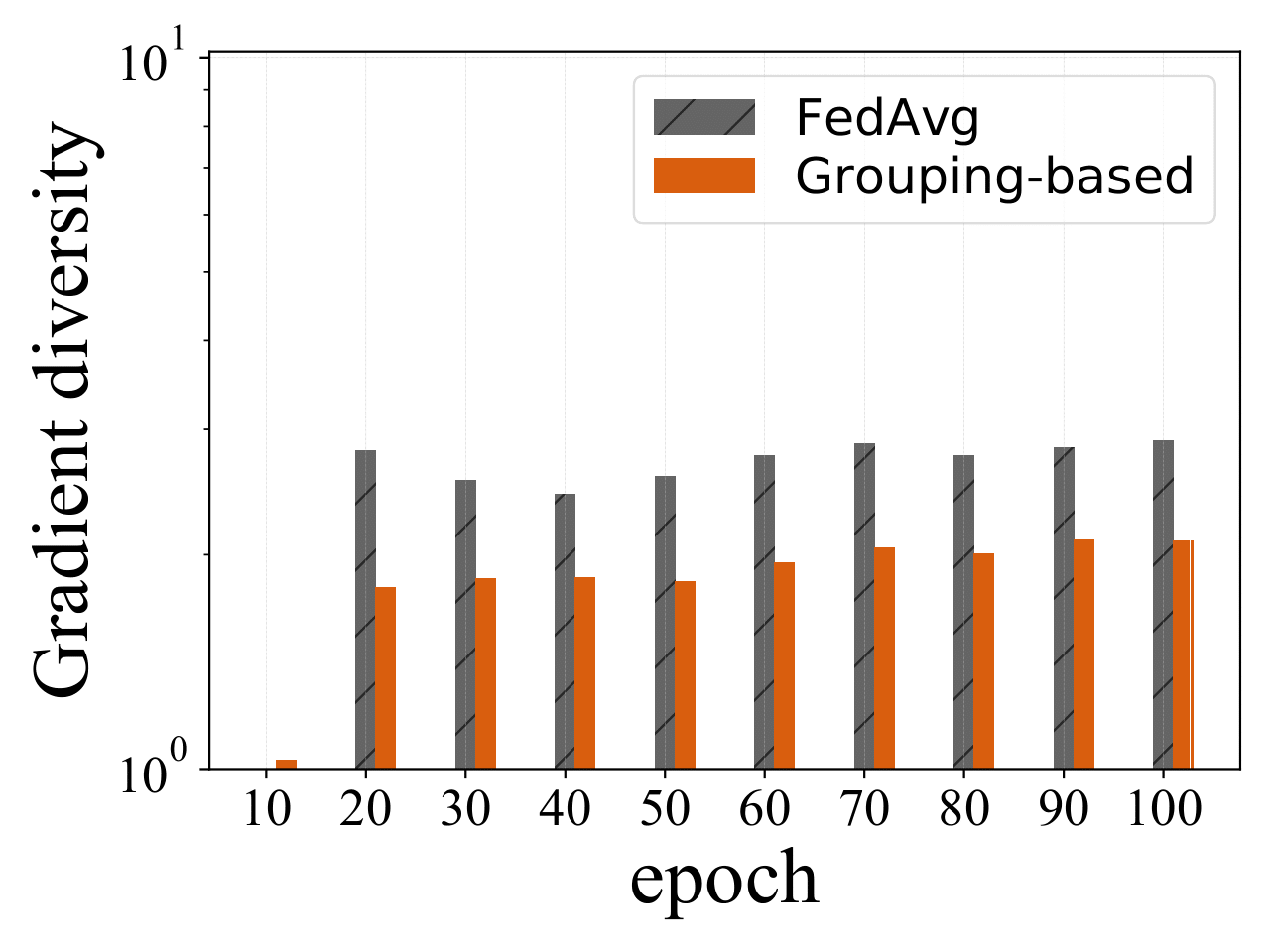}
    \includegraphics[width=0.32\linewidth]{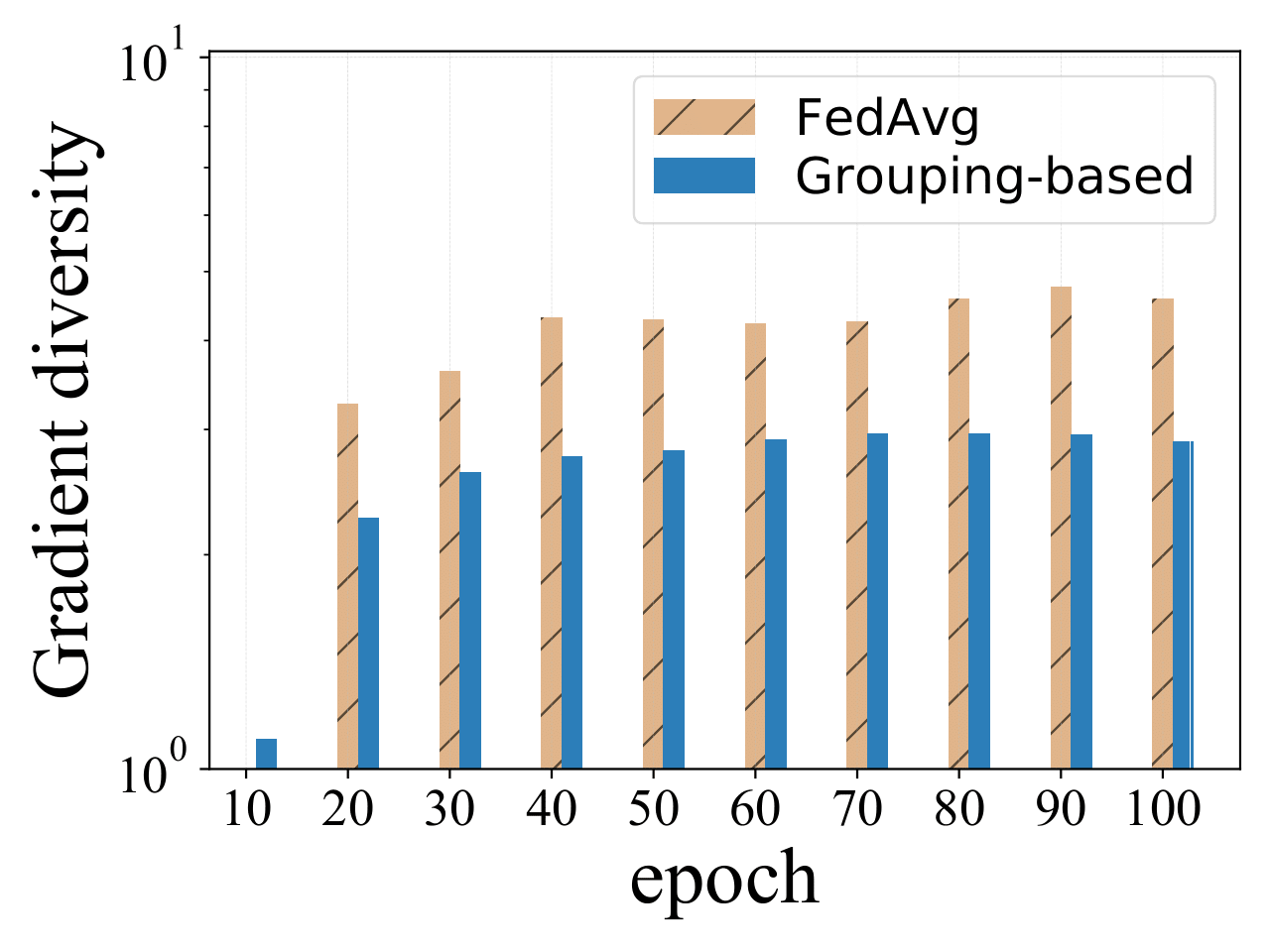}
    \includegraphics[width=0.32\linewidth]{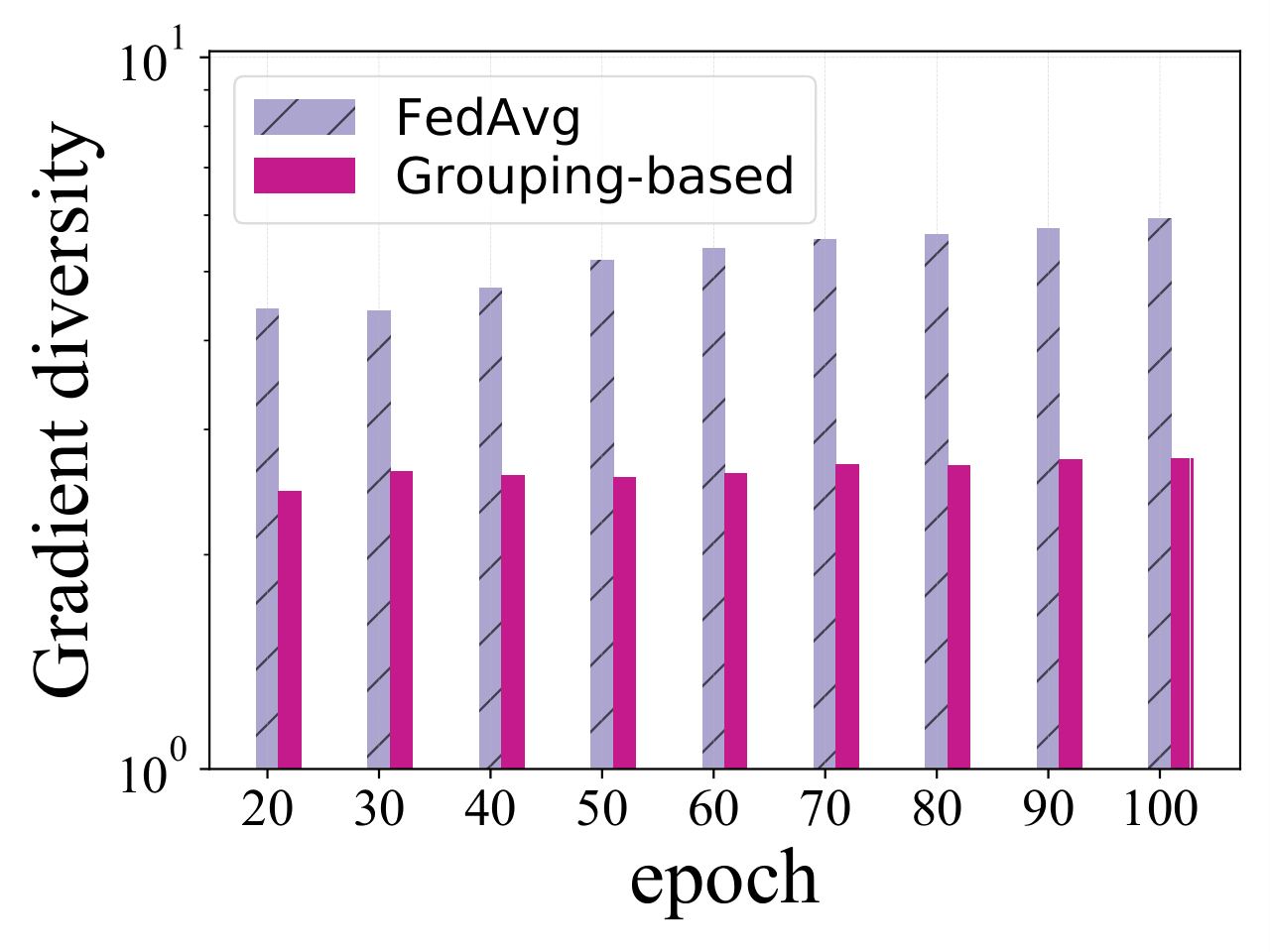}
    \includegraphics[width=0.32\linewidth]{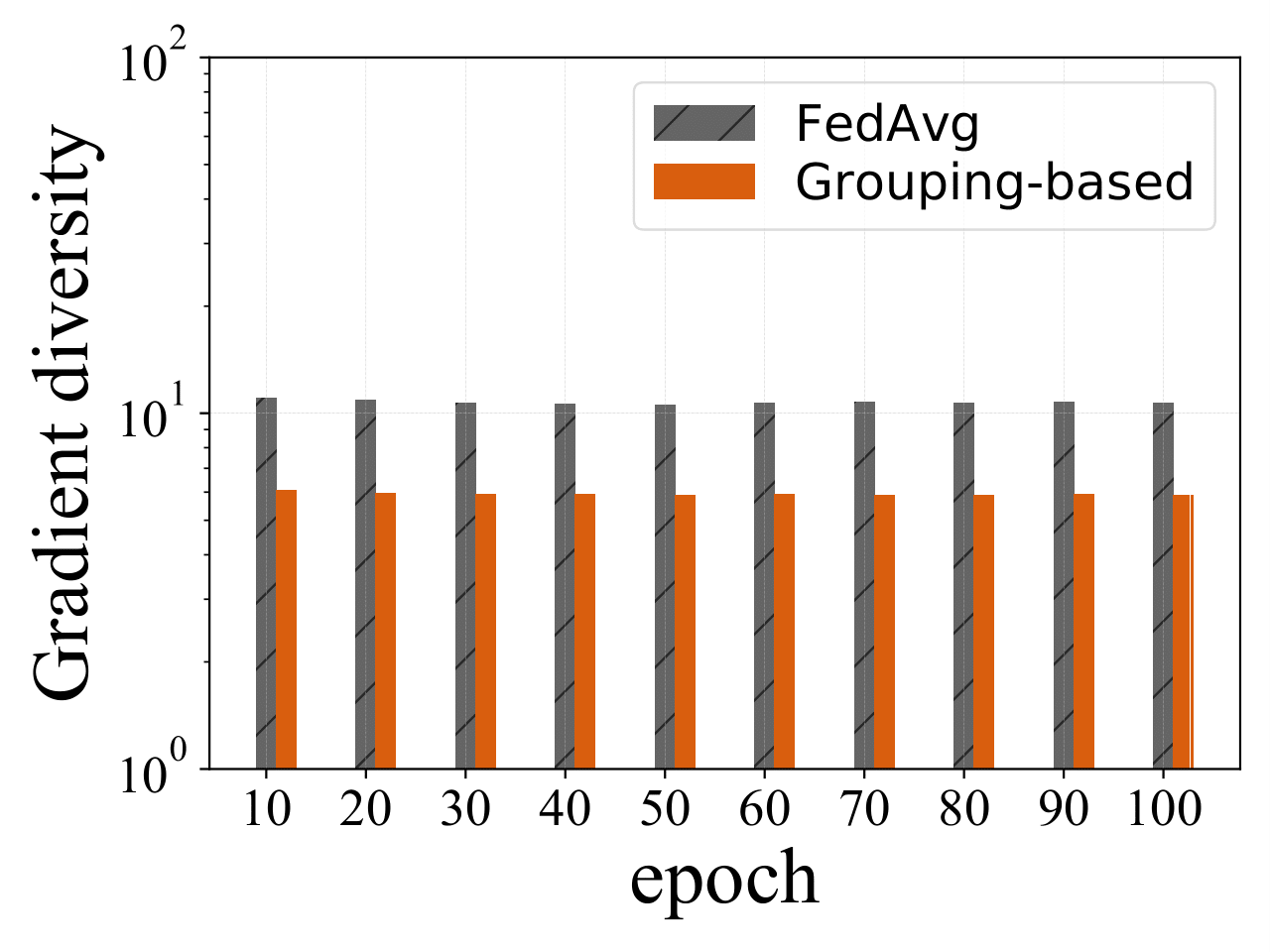}
    \includegraphics[width=0.32\linewidth]{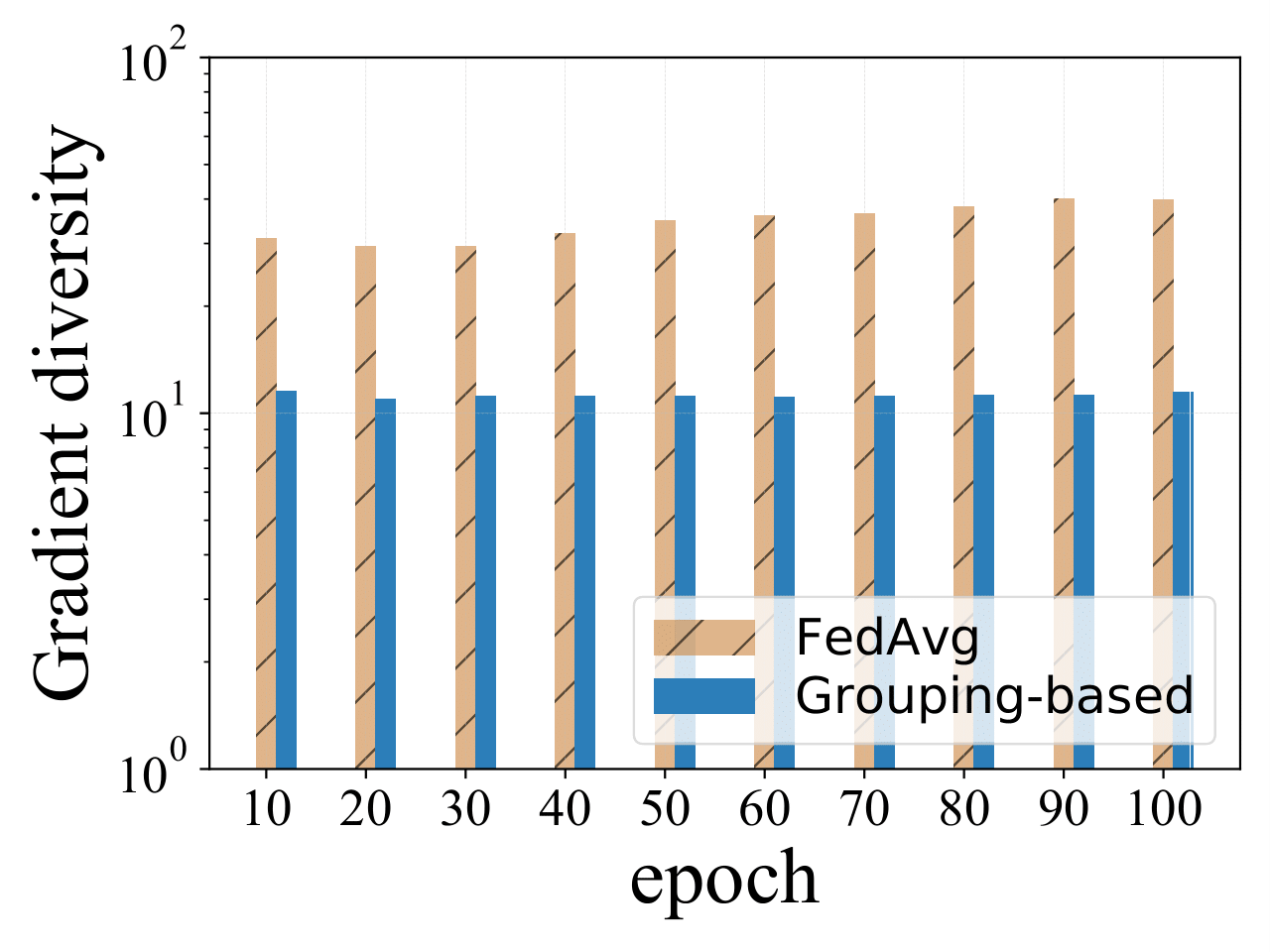}
    \includegraphics[width=0.32\linewidth]{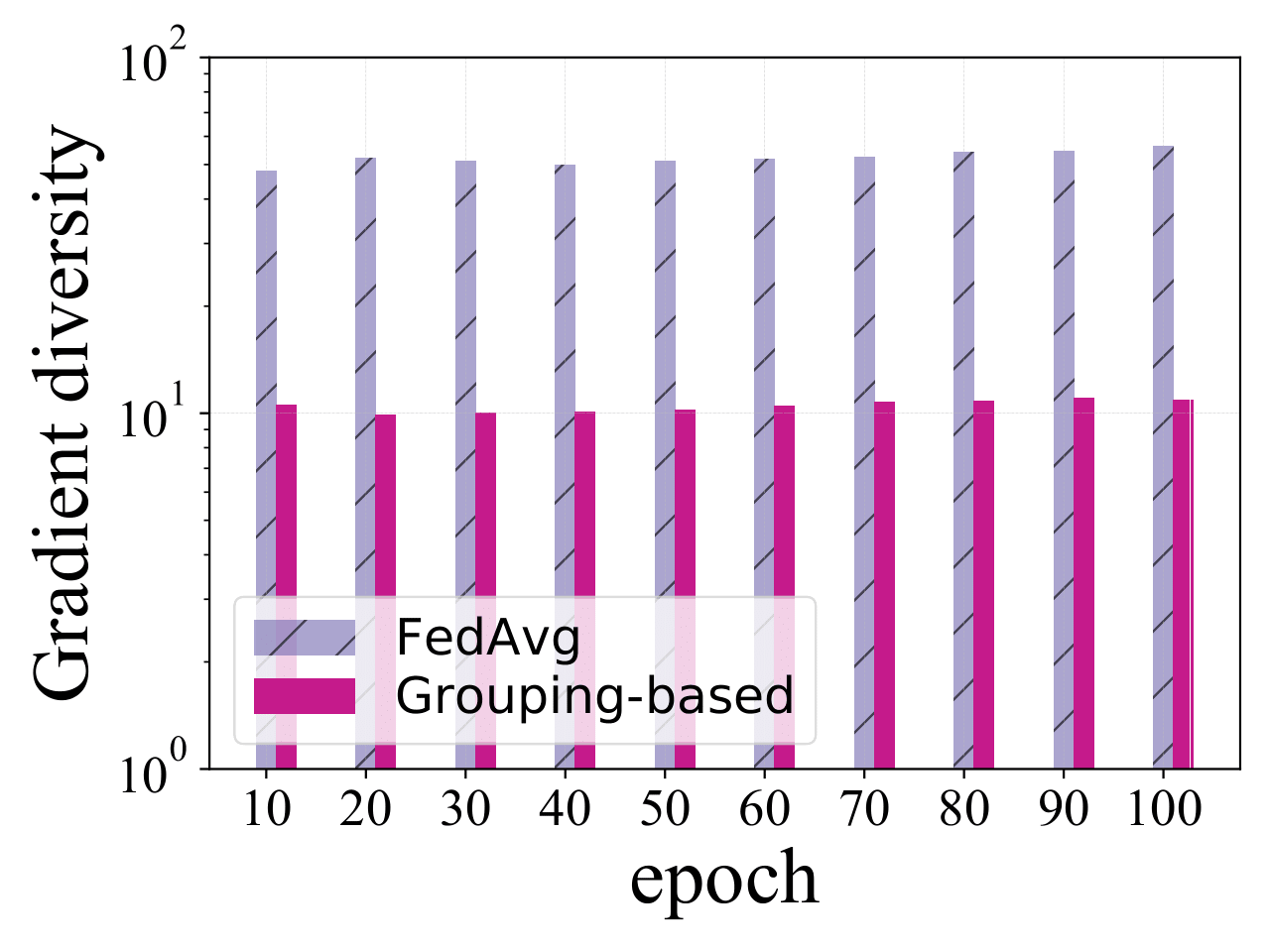}
    \includegraphics[width=0.32\linewidth]{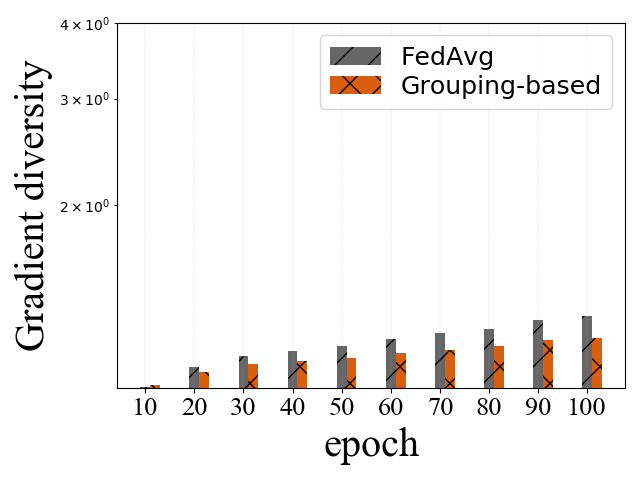}
    \includegraphics[width=0.32\linewidth]{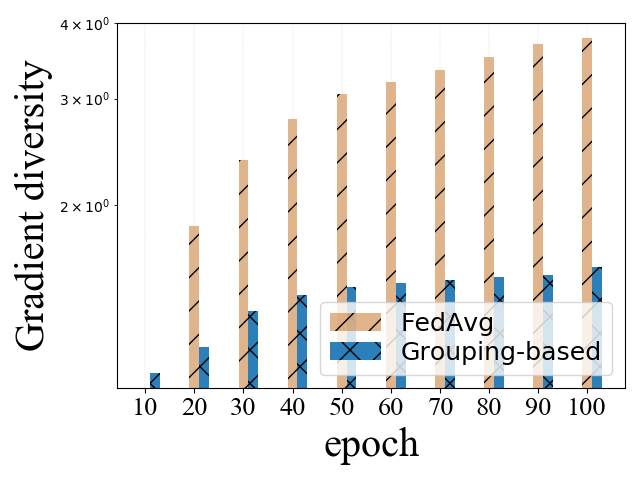}
    \includegraphics[width=0.32\linewidth]{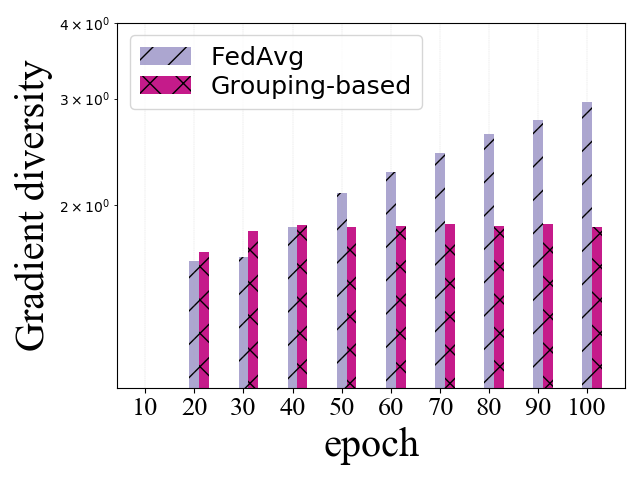}
    \caption{
    (Line 1) The convergence curves on EMNIST.
    (Line 2) Results on gradient diversity defined in \eref{graddiversityType3}. 
    (Line 3) Results on gradient diversity defined in \eref{graddiversityType4}. 
    (Line 4) Results on gradient diversity defined in \eref{graddiversityType3}, including both the users and the server. 
    (Line 5) Results on gradient diversity defined in \eref{graddiversityType4}, including both the users and the server. 
    }
    \label{fig:Groupingmethod_NormOrd1}
\end{figure*}

\begin{figure*}
    \centering
    \includegraphics[width=0.32\linewidth]{figs/EMNIST_ACC_FedAvg_vs_Grouping_commUE10_deversity-1.png}
    \includegraphics[width=0.32\linewidth]{figs/EMNIST_ACC_FedAvg_vs_Grouping_commUE30_deversity-1.png}
    \includegraphics[width=0.32\linewidth]{figs/EMNIST_ACC_FedAvg_vs_Grouping_commUE47_deversity-1.png}
    \includegraphics[width=0.32\linewidth]{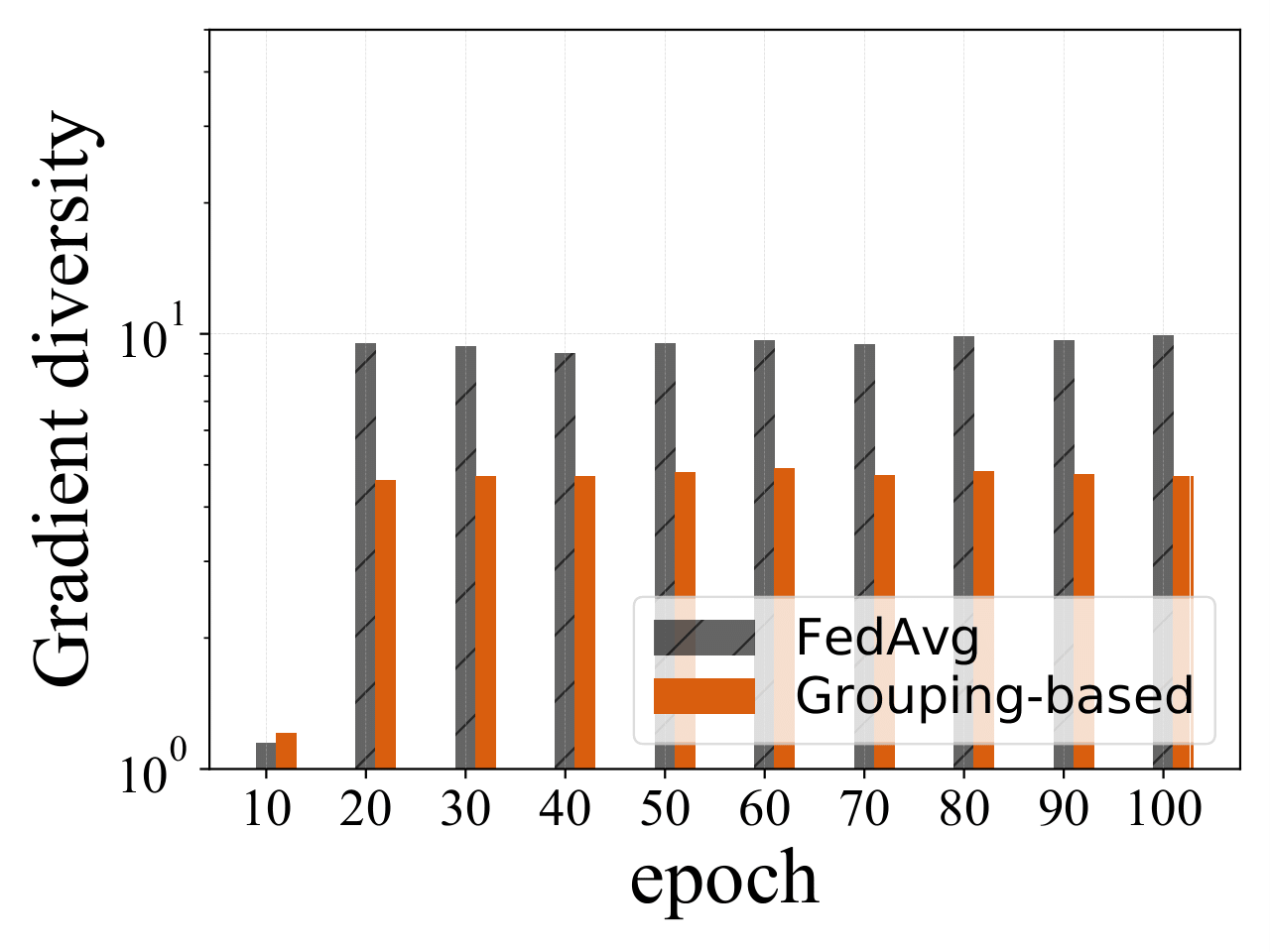}
    \includegraphics[width=0.32\linewidth]{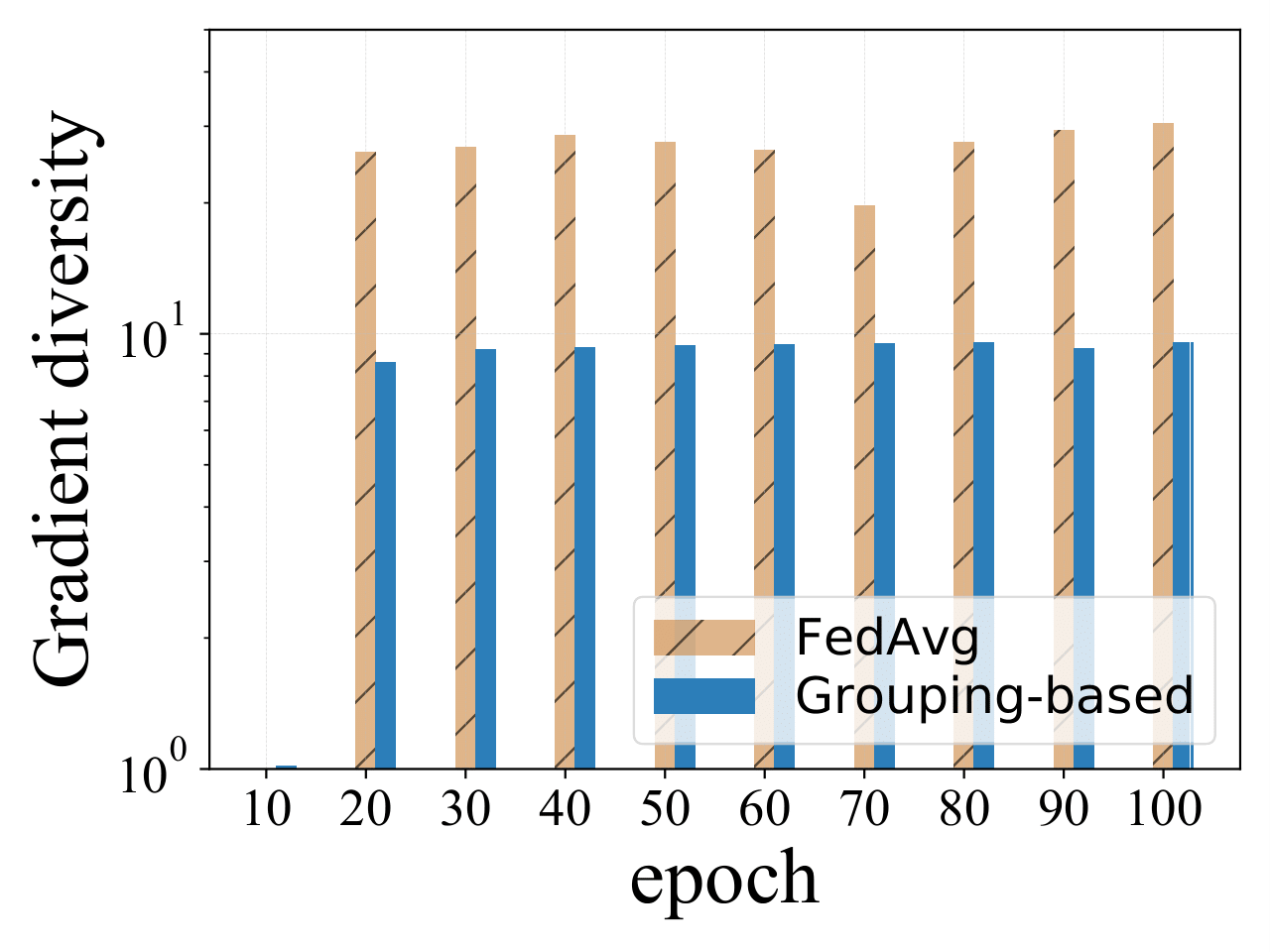}
    \includegraphics[width=0.32\linewidth]{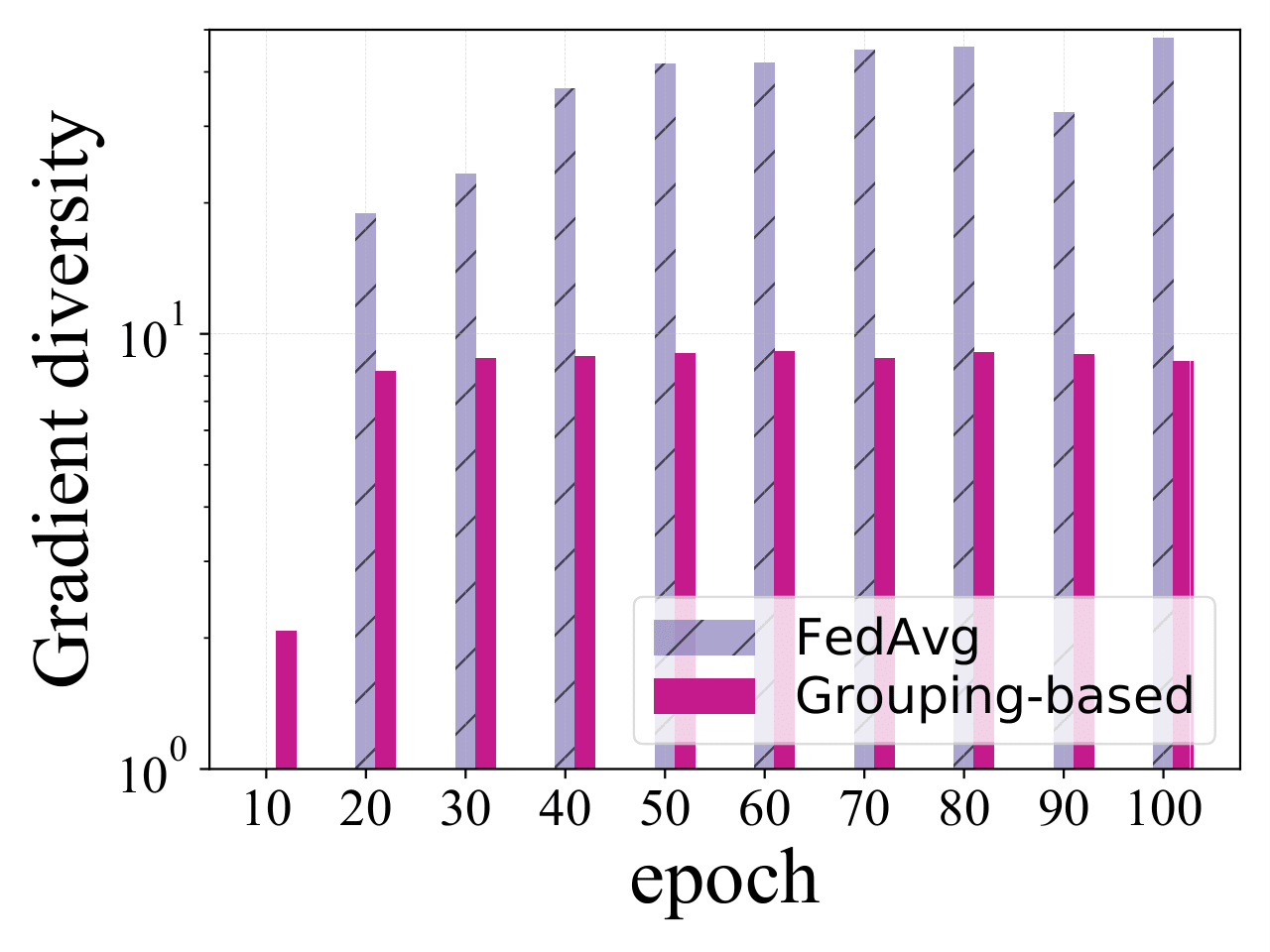}
    \includegraphics[width=0.32\linewidth]{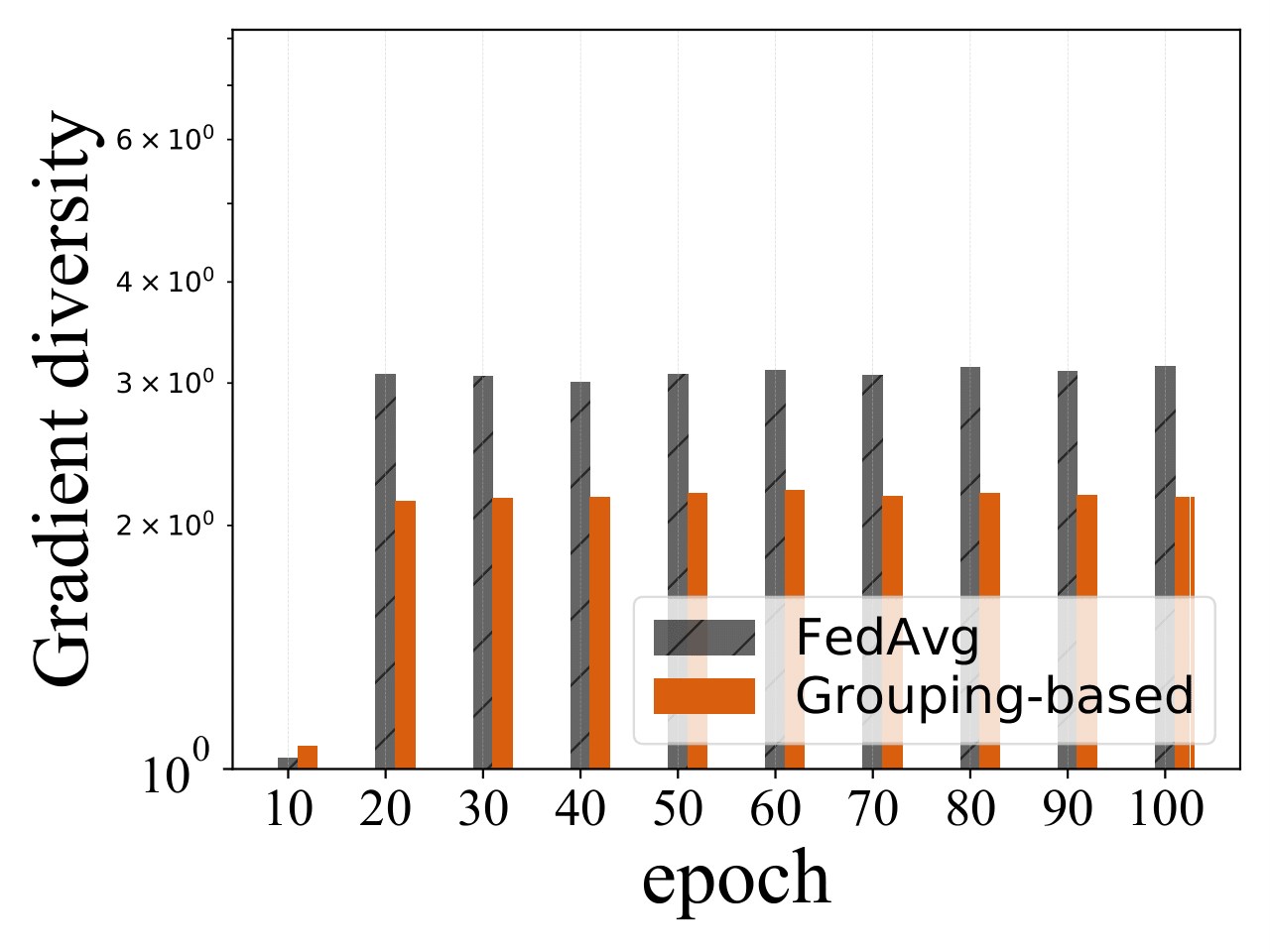}
    \includegraphics[width=0.32\linewidth]{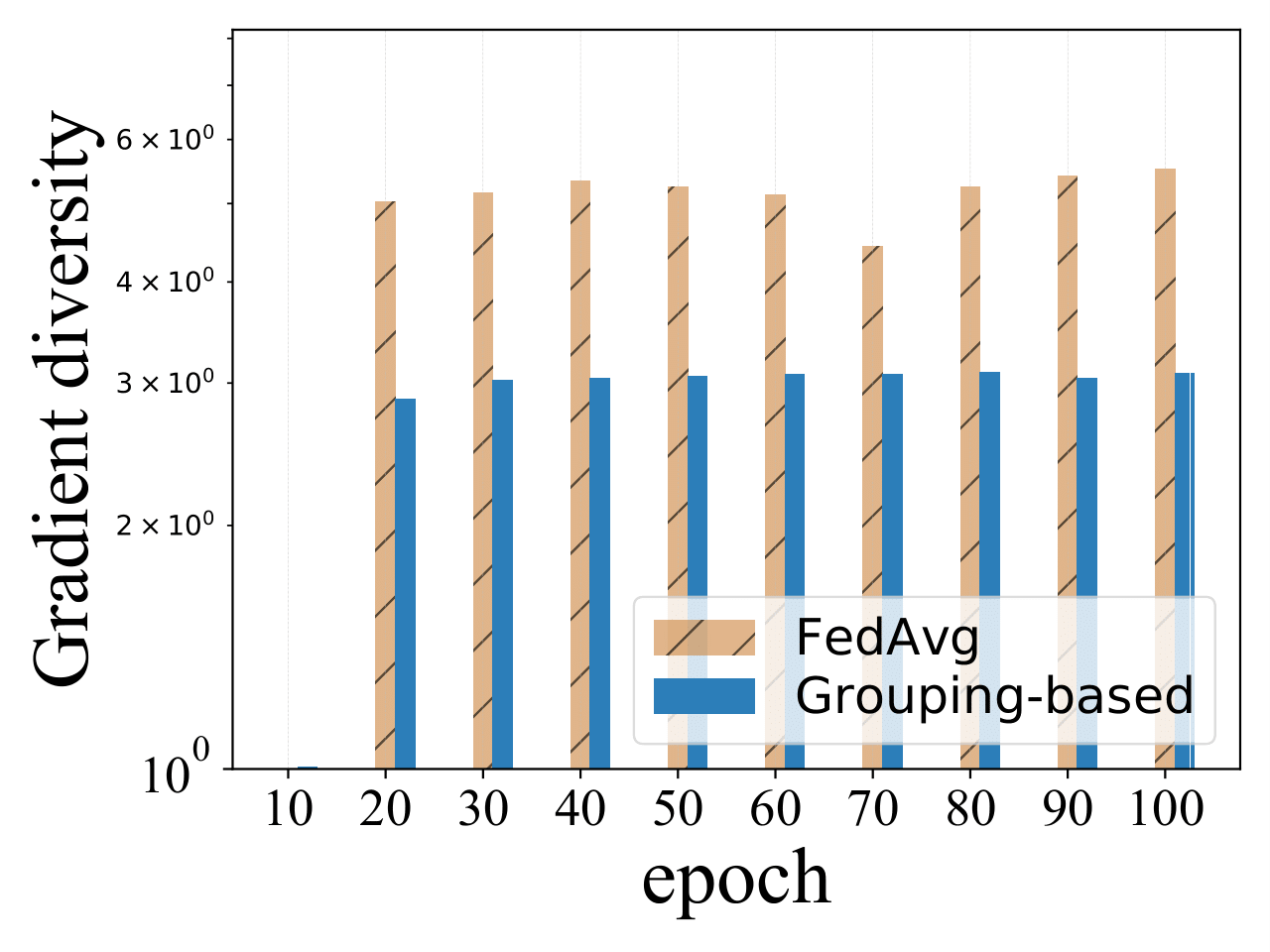}
    \includegraphics[width=0.32\linewidth]{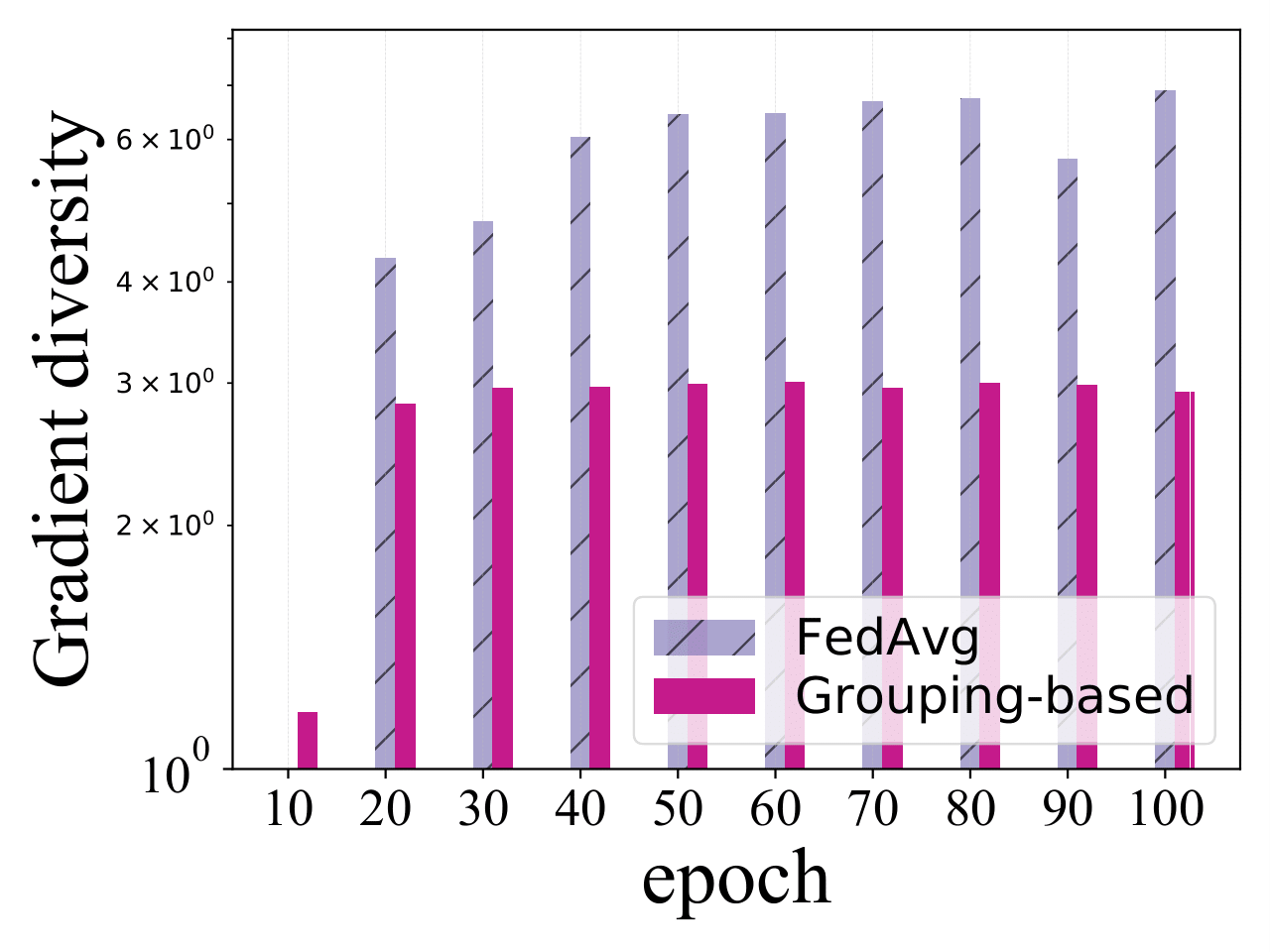}
    \includegraphics[width=0.32\linewidth]{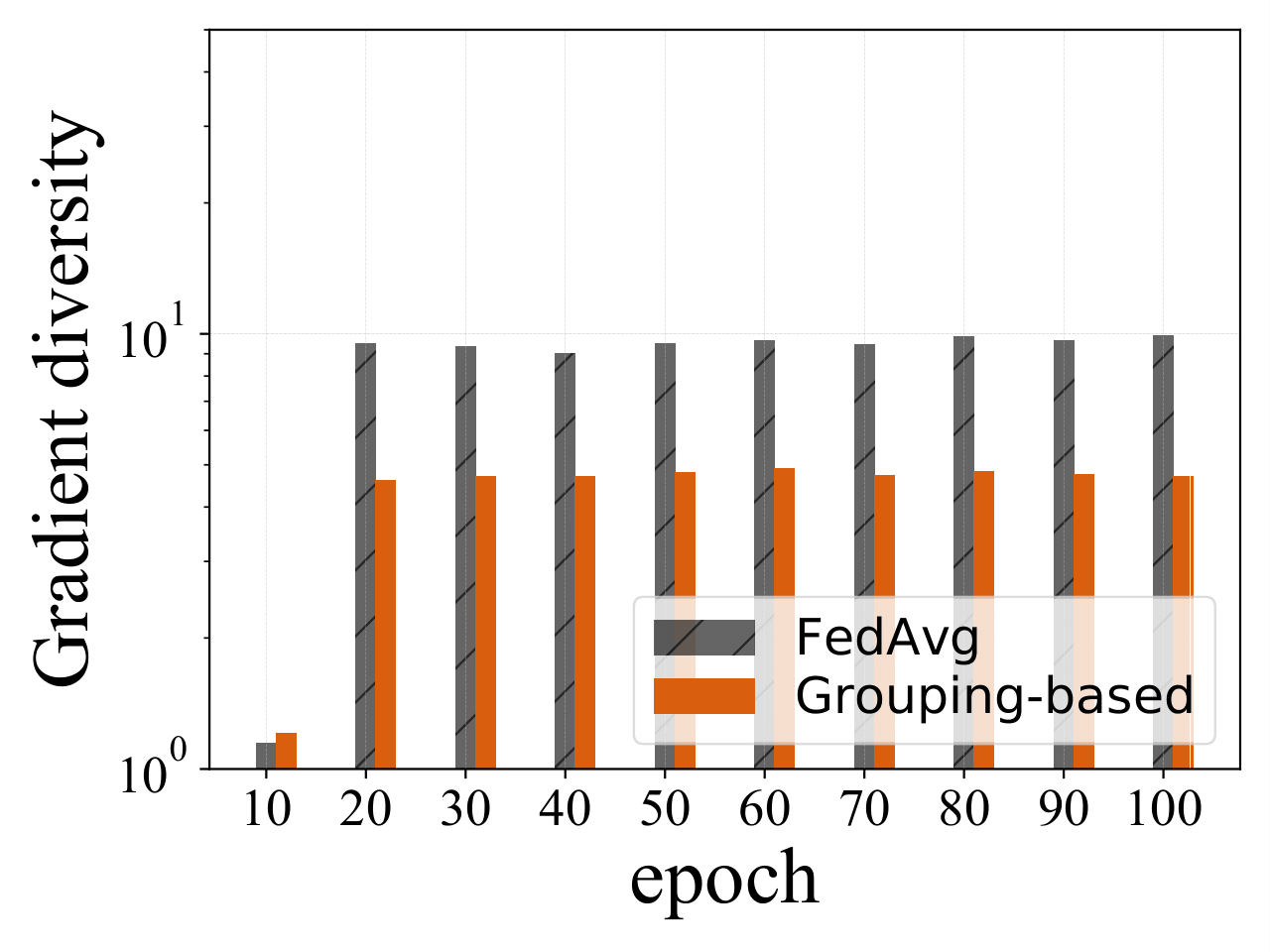}
    \includegraphics[width=0.32\linewidth]{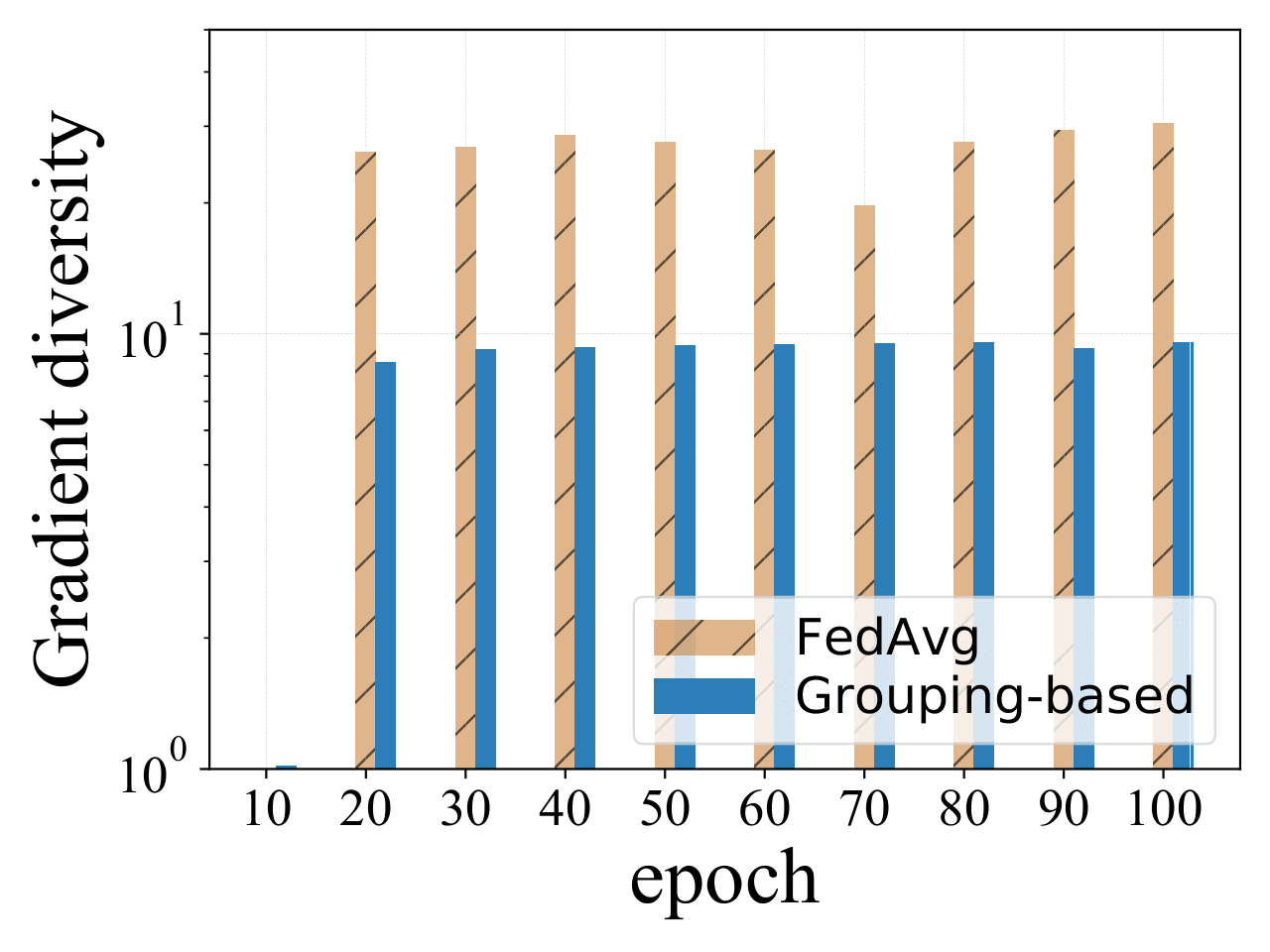}
    \includegraphics[width=0.32\linewidth]{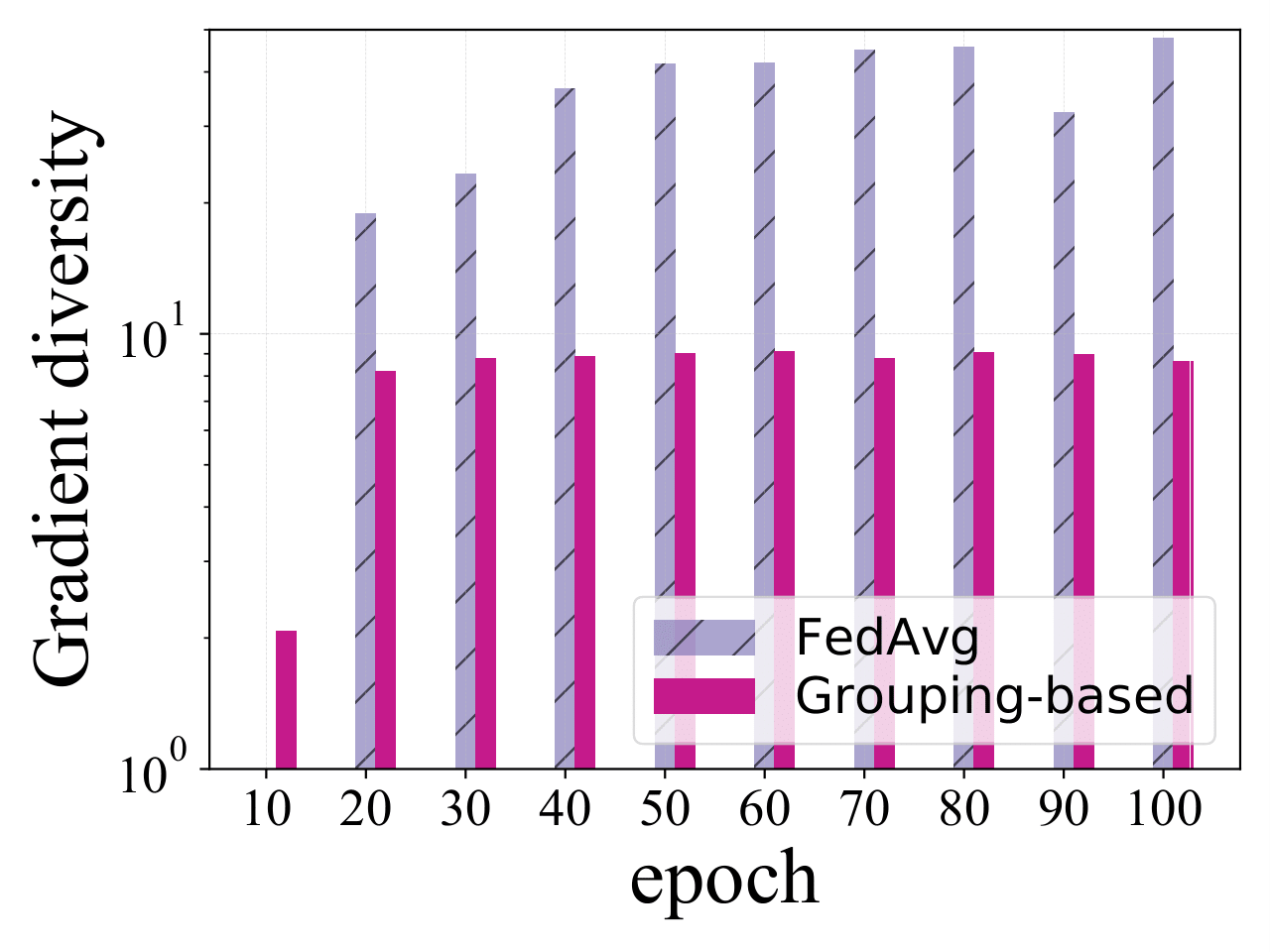}
    \includegraphics[width=0.32\linewidth]{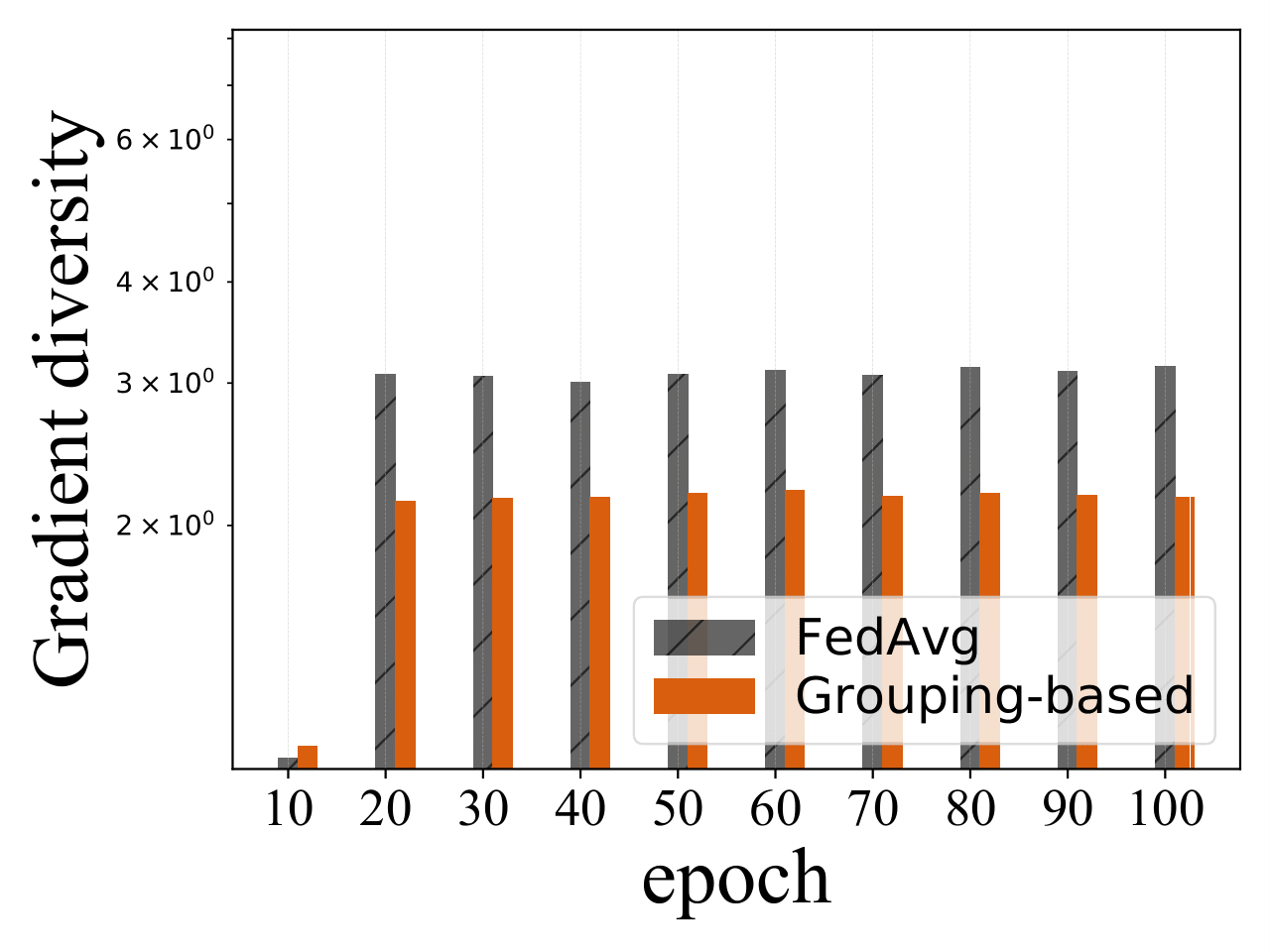}
    \includegraphics[width=0.32\linewidth]{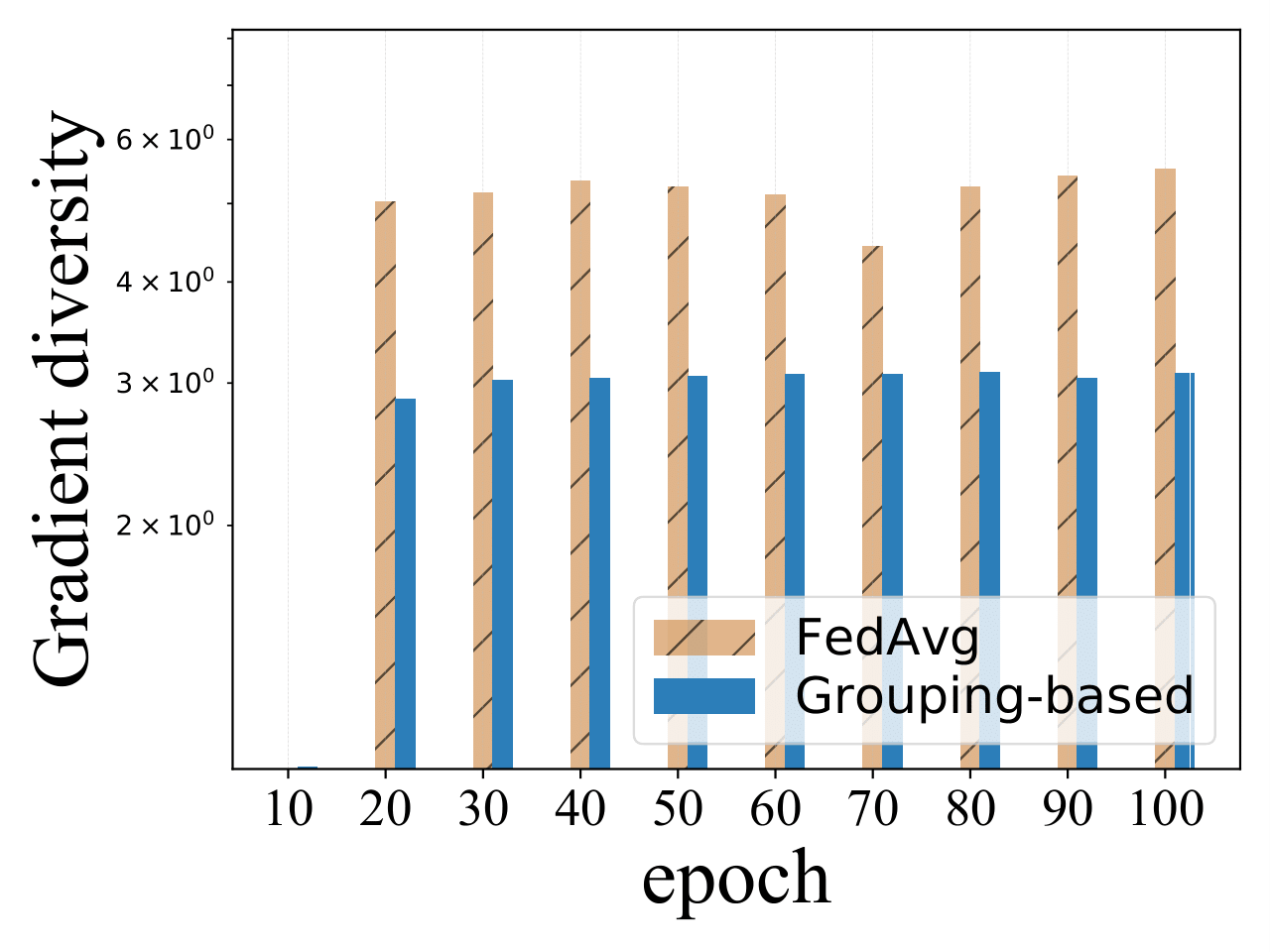}
    \includegraphics[width=0.32\linewidth]{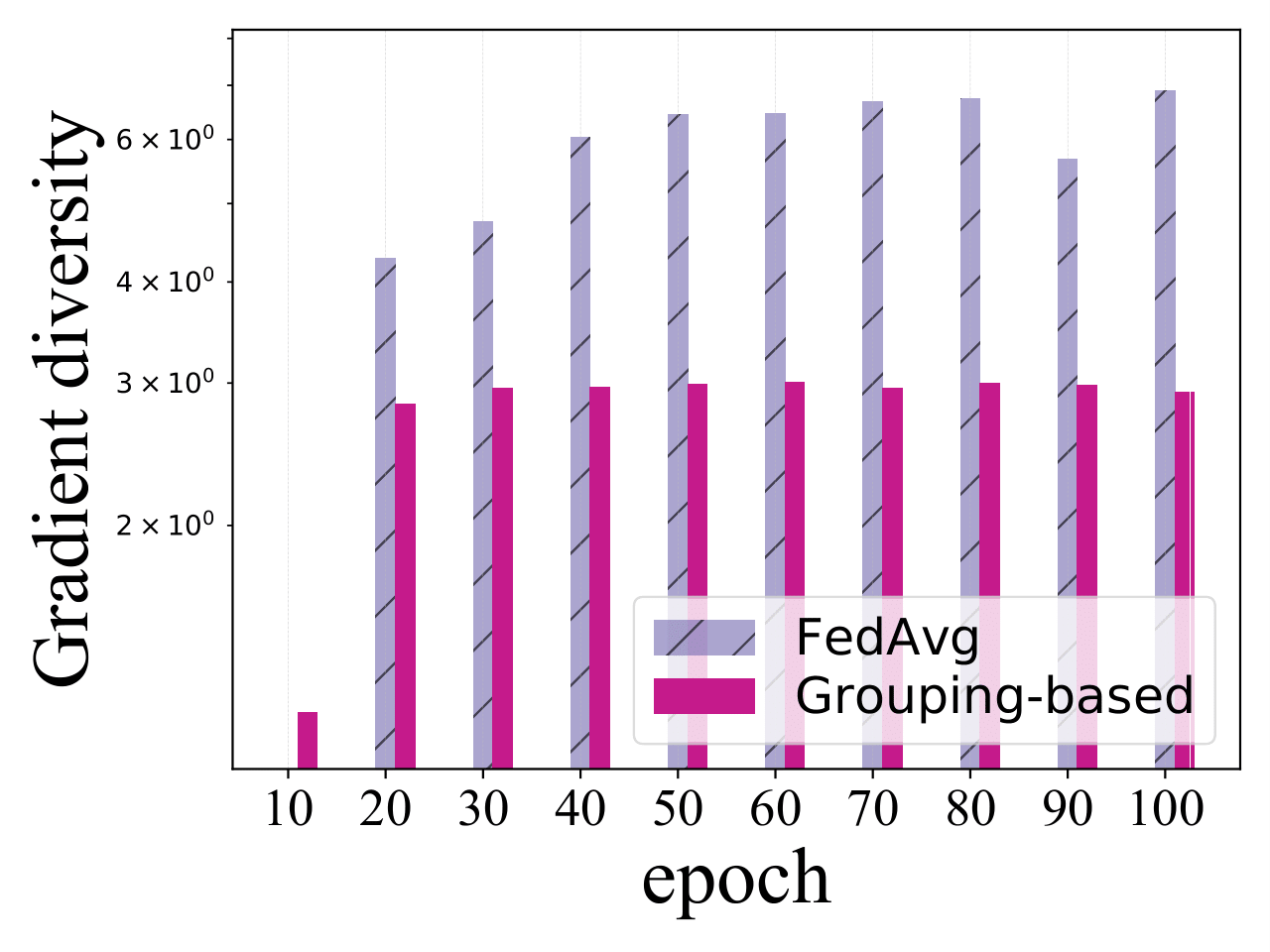}
    \caption{
    (Line 1) The convergence curves on EMNIST.
    (Line 2) Results on gradient diversity defined in \eref{def:weight_diversity} with $w_k^t$ defined in \eref{CumulativeGradient}. 
    (Line 3) Results on gradient diversity defined in \eref{graddiversityType2} with $w_k^t$ defined in \eref{CumulativeGradient}. 
    (Line 4) Results on gradient diversity defined in \eref{def:weight_diversity} with $w_k^t$ defined in \eref{CumulativeGradient}, including both the users and the server. 
    (Line 5) Results on gradient diversity defined in \eref{graddiversityType2} with $w_k^t$ defined in \eref{CumulativeGradient}, including both the users and the server. 
    }
    \label{fig:Groupingmethod_NormOrd2_with_FedGrad}
\end{figure*}

\begin{figure*}
    \centering
    \includegraphics[width=0.32\linewidth]{figs/EMNIST_ACC_FedAvg_vs_Grouping_commUE10_deversity-1.png}
    \includegraphics[width=0.32\linewidth]{figs/EMNIST_ACC_FedAvg_vs_Grouping_commUE30_deversity-1.png}
    \includegraphics[width=0.32\linewidth]{figs/EMNIST_ACC_FedAvg_vs_Grouping_commUE47_deversity-1.png}
    \includegraphics[width=0.32\linewidth]{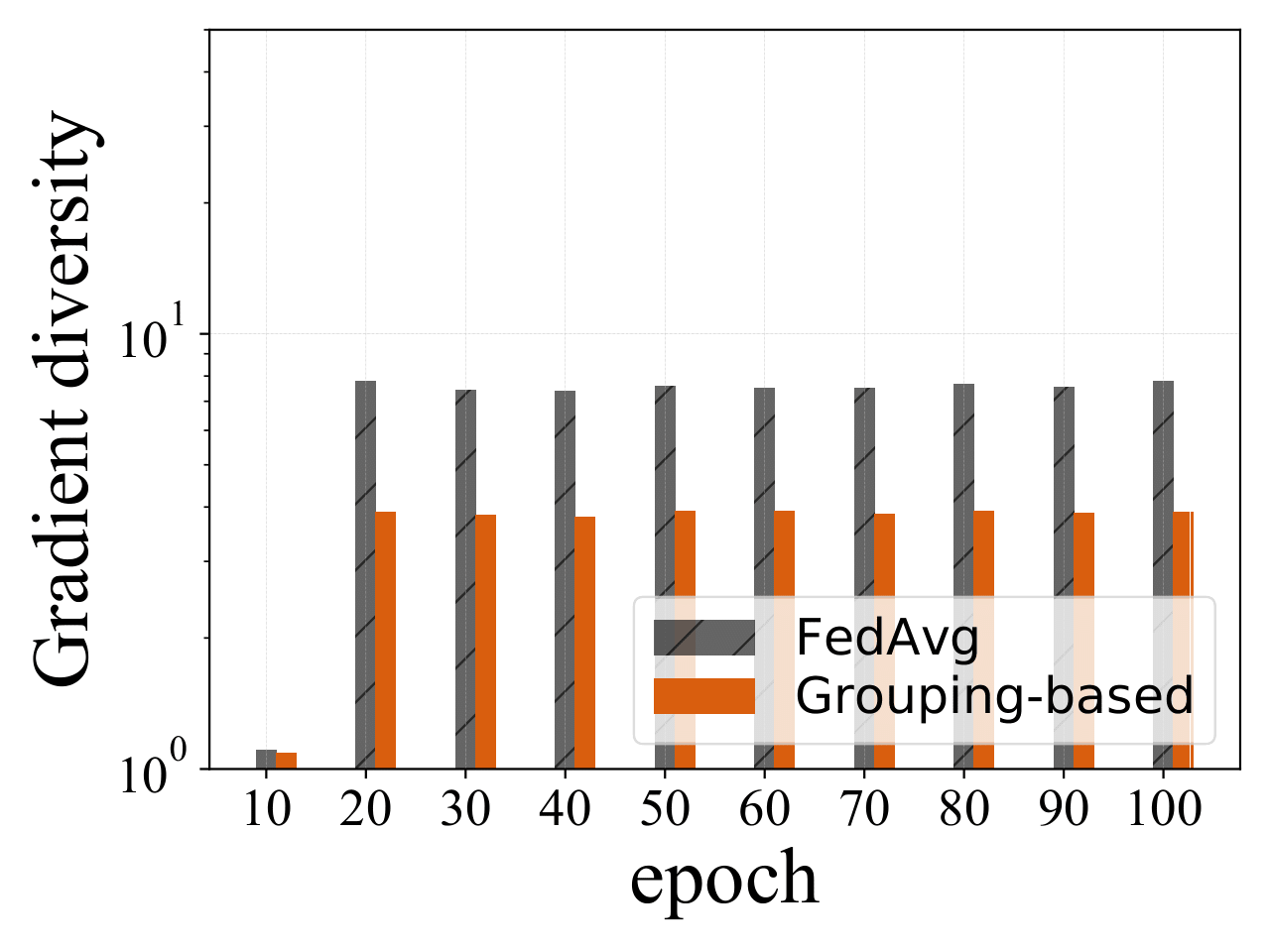}
    \includegraphics[width=0.32\linewidth]{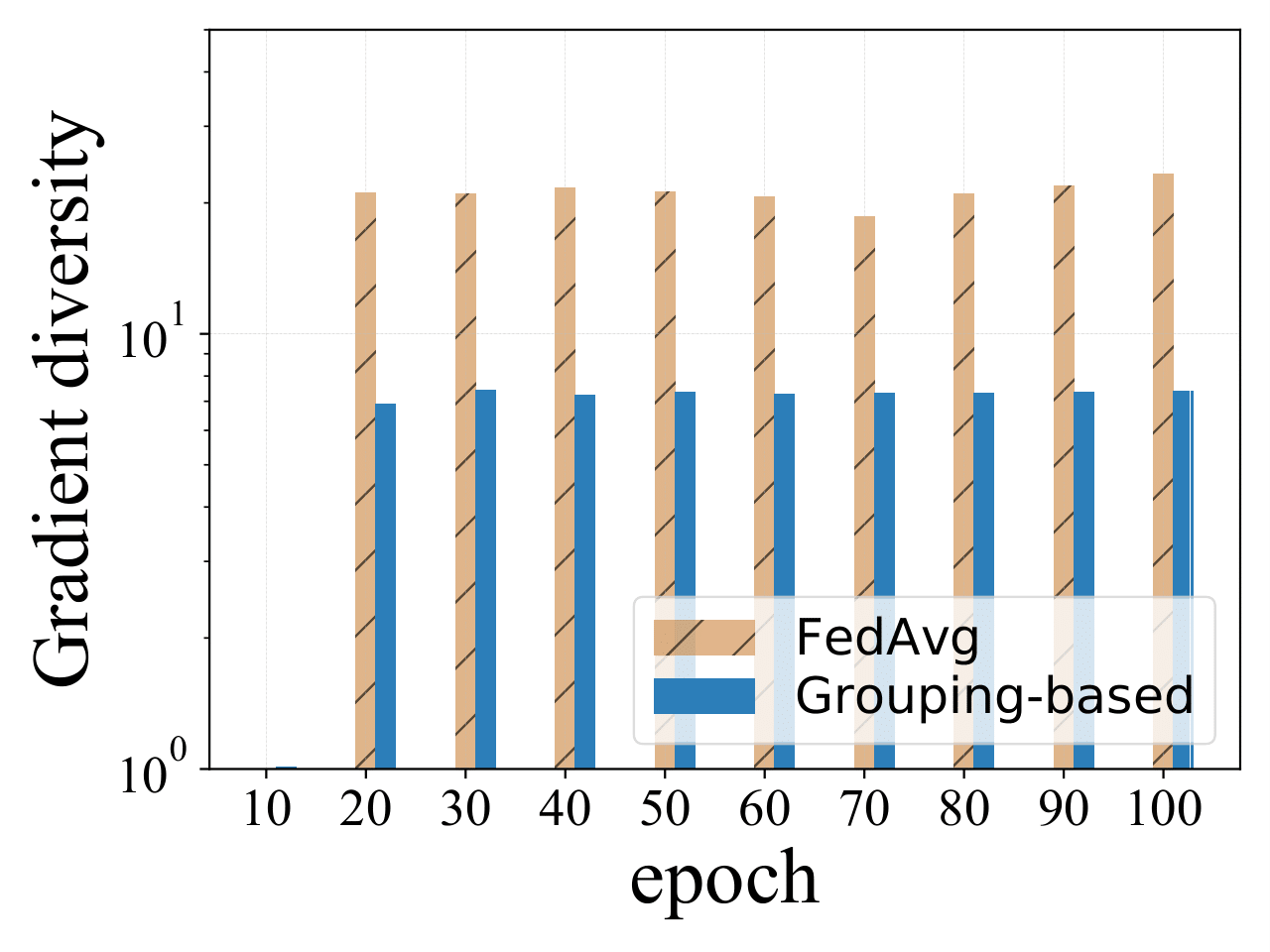}
    \includegraphics[width=0.32\linewidth]{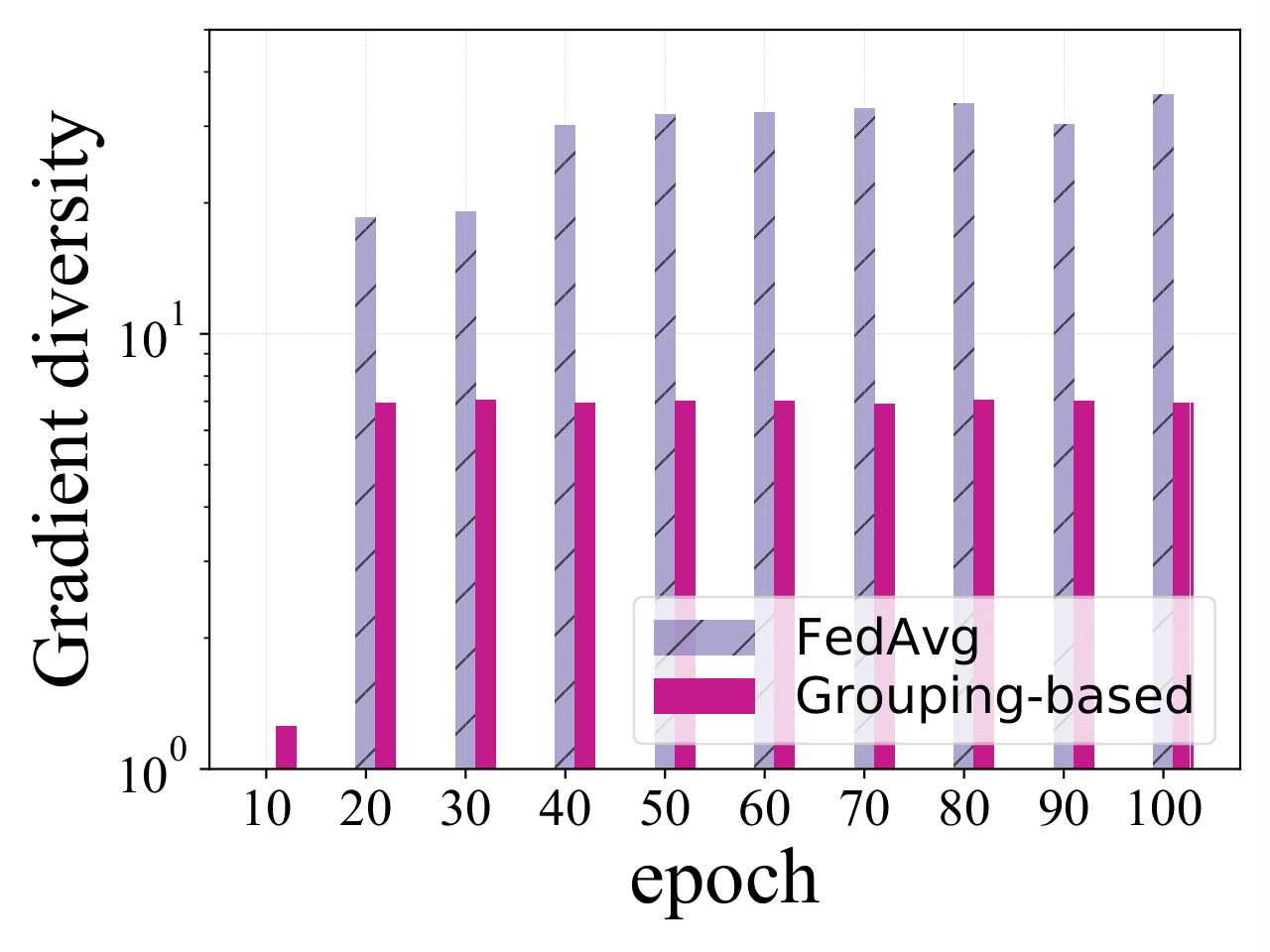}
    \includegraphics[width=0.32\linewidth]{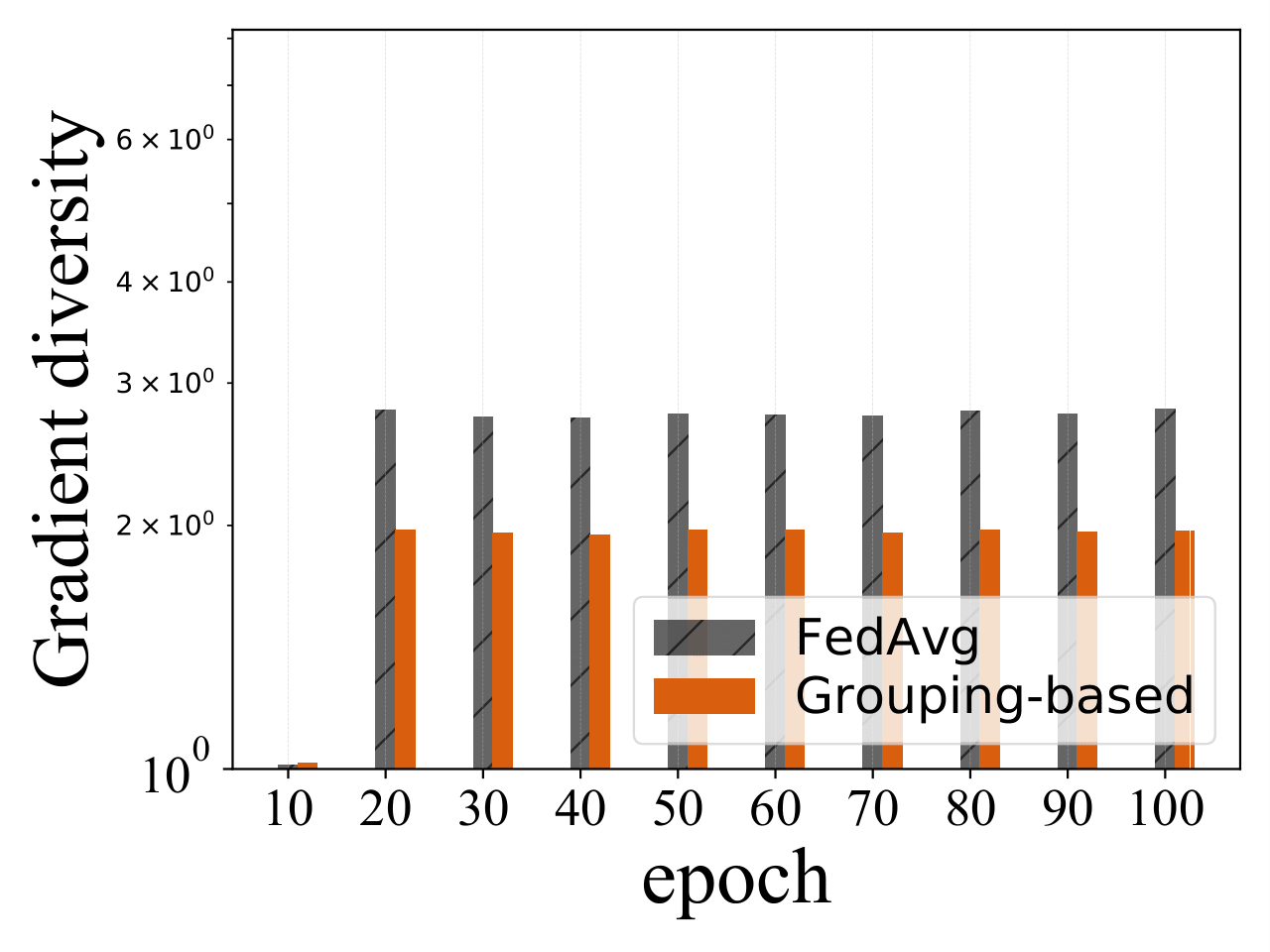}
    \includegraphics[width=0.32\linewidth]{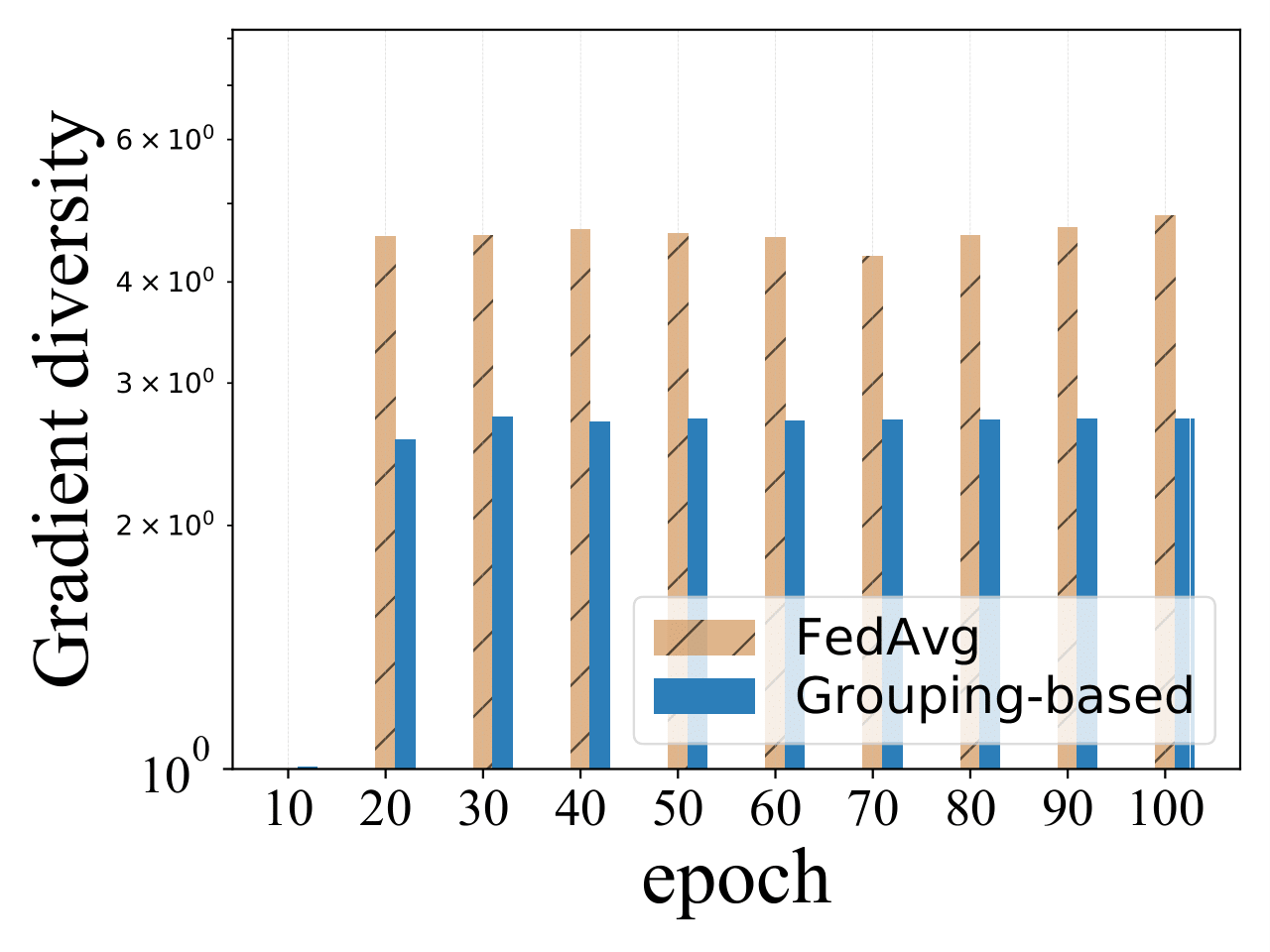}
    \includegraphics[width=0.32\linewidth]{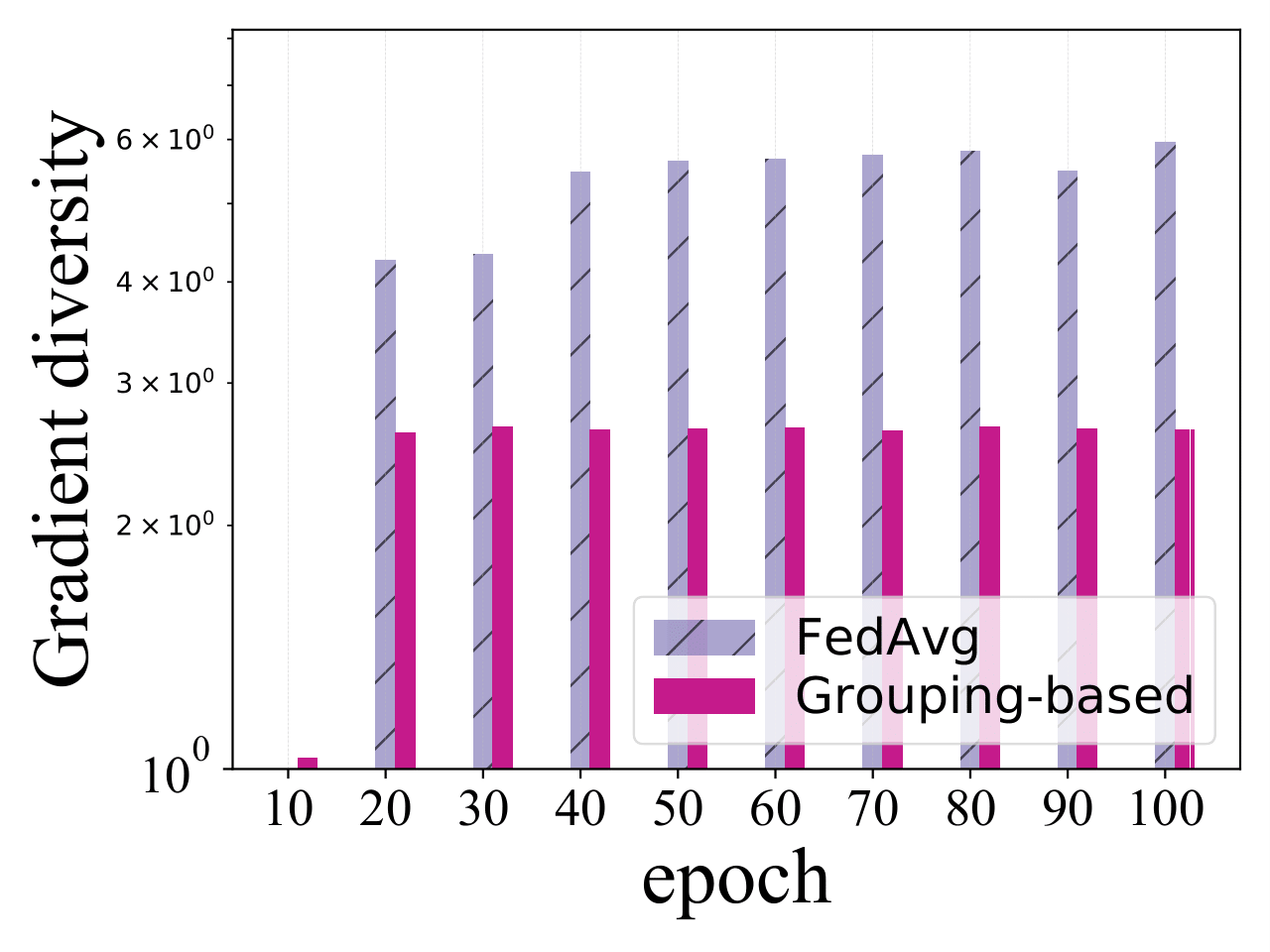}
    \includegraphics[width=0.32\linewidth]{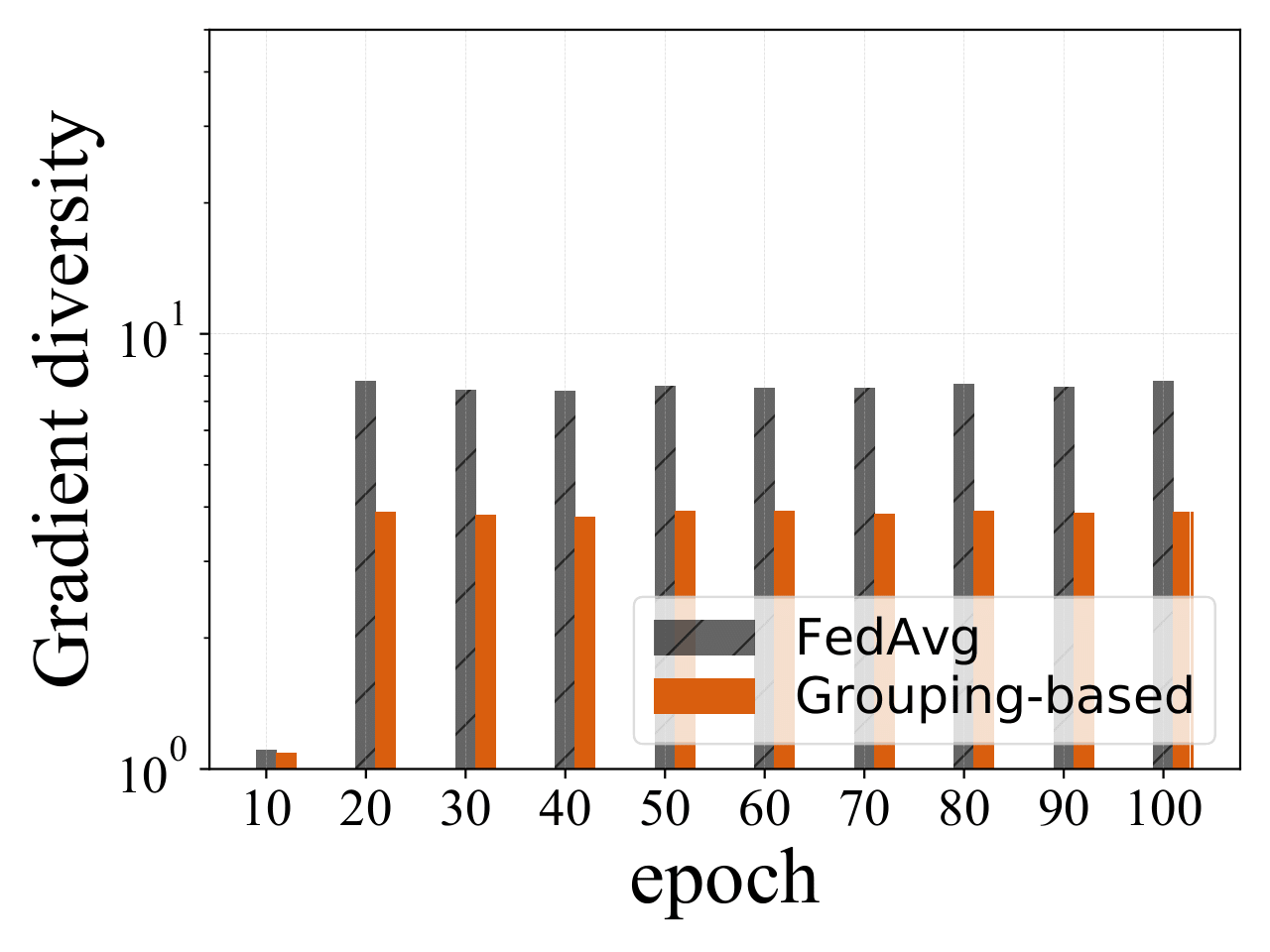}
    \includegraphics[width=0.32\linewidth]{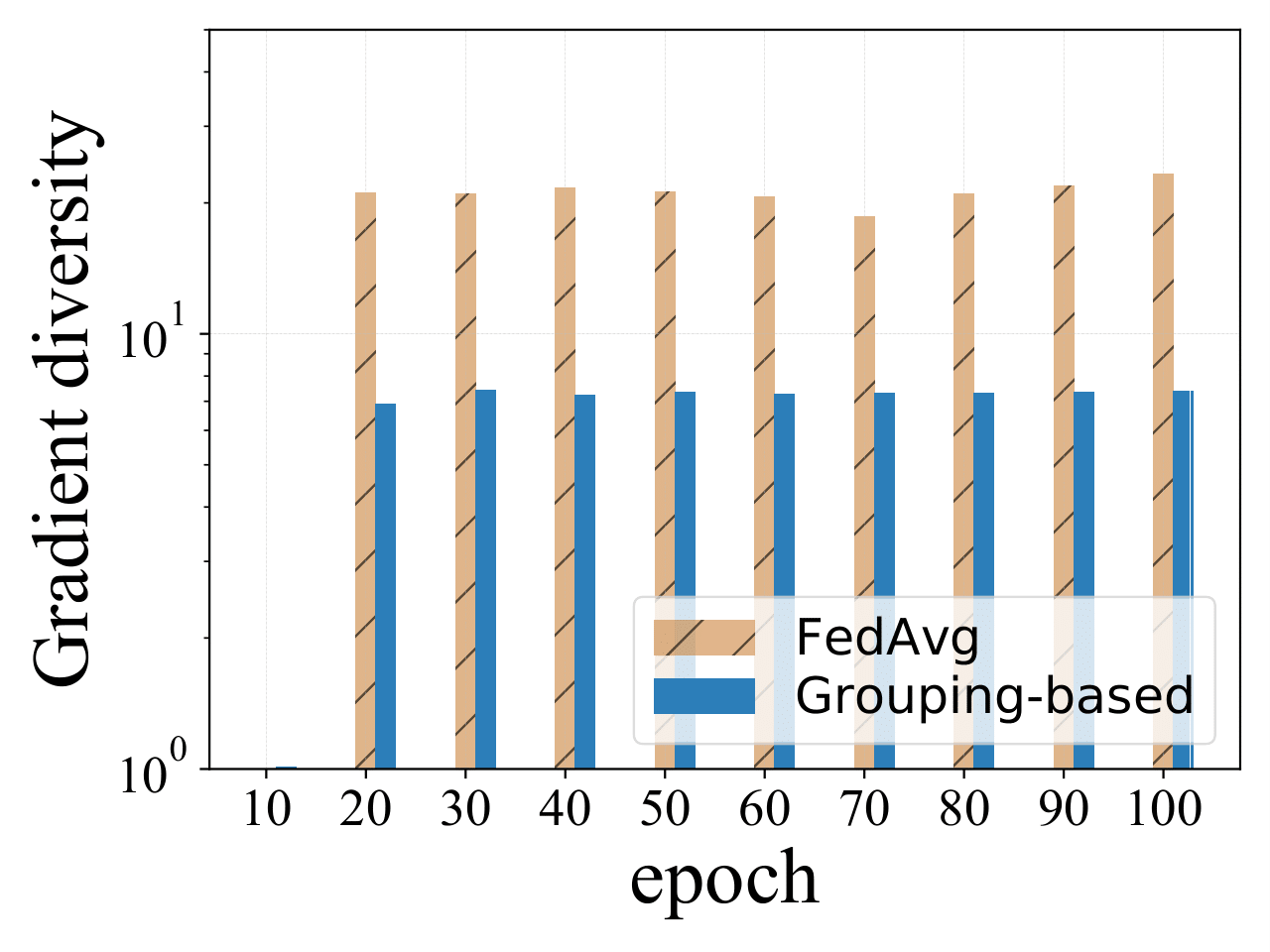}
    \includegraphics[width=0.32\linewidth]{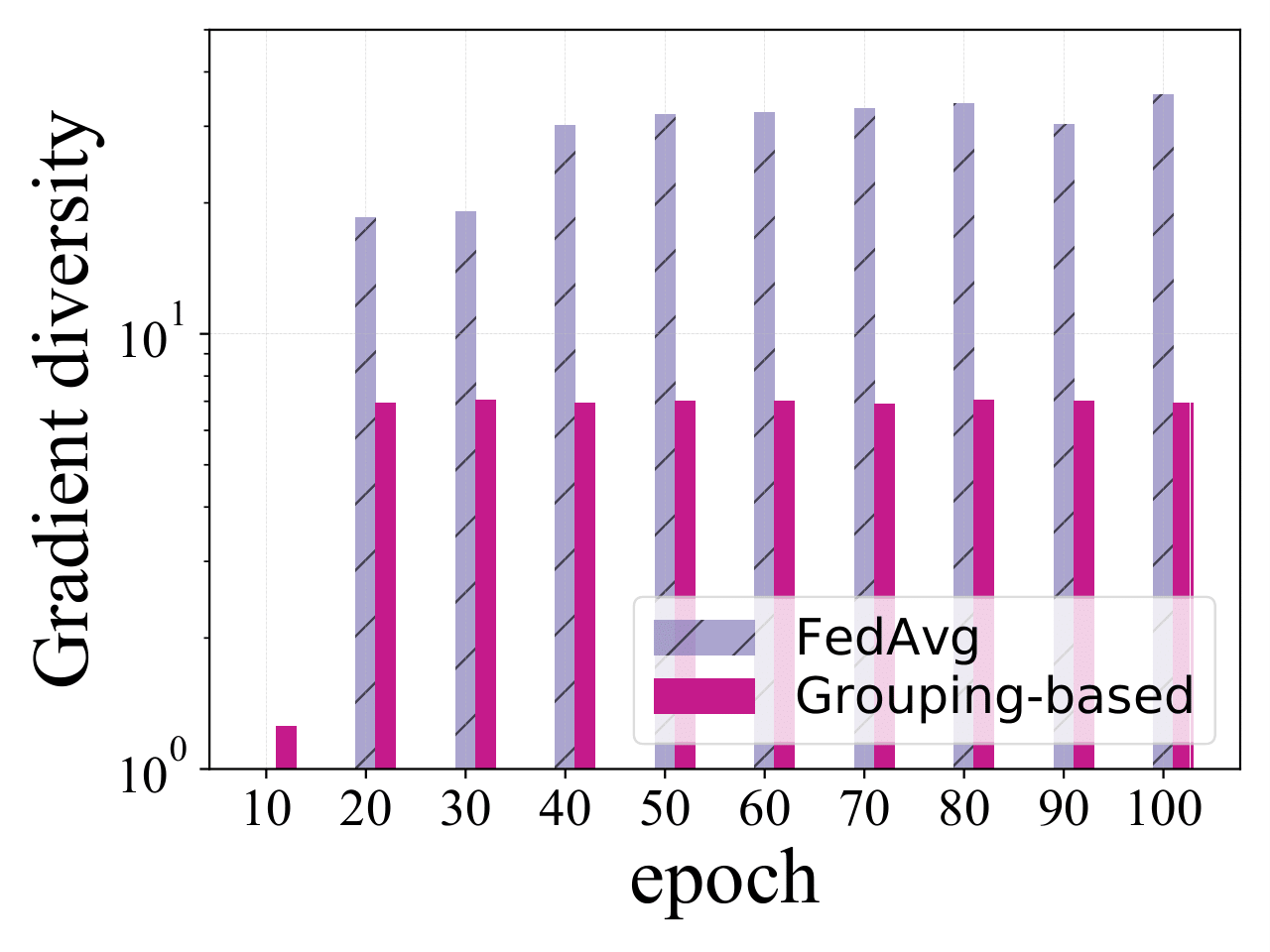}
    \includegraphics[width=0.32\linewidth]{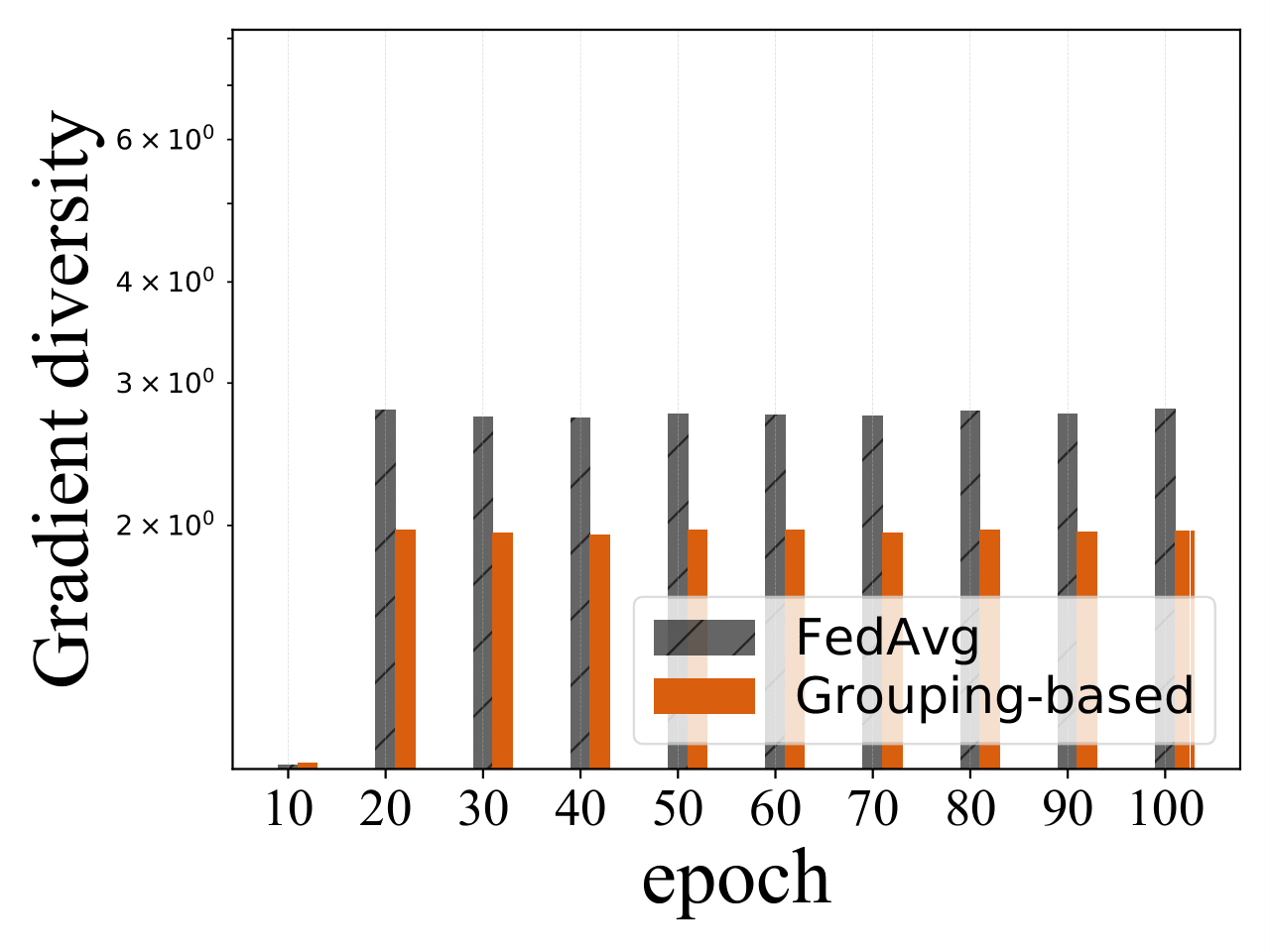}
    \includegraphics[width=0.32\linewidth]{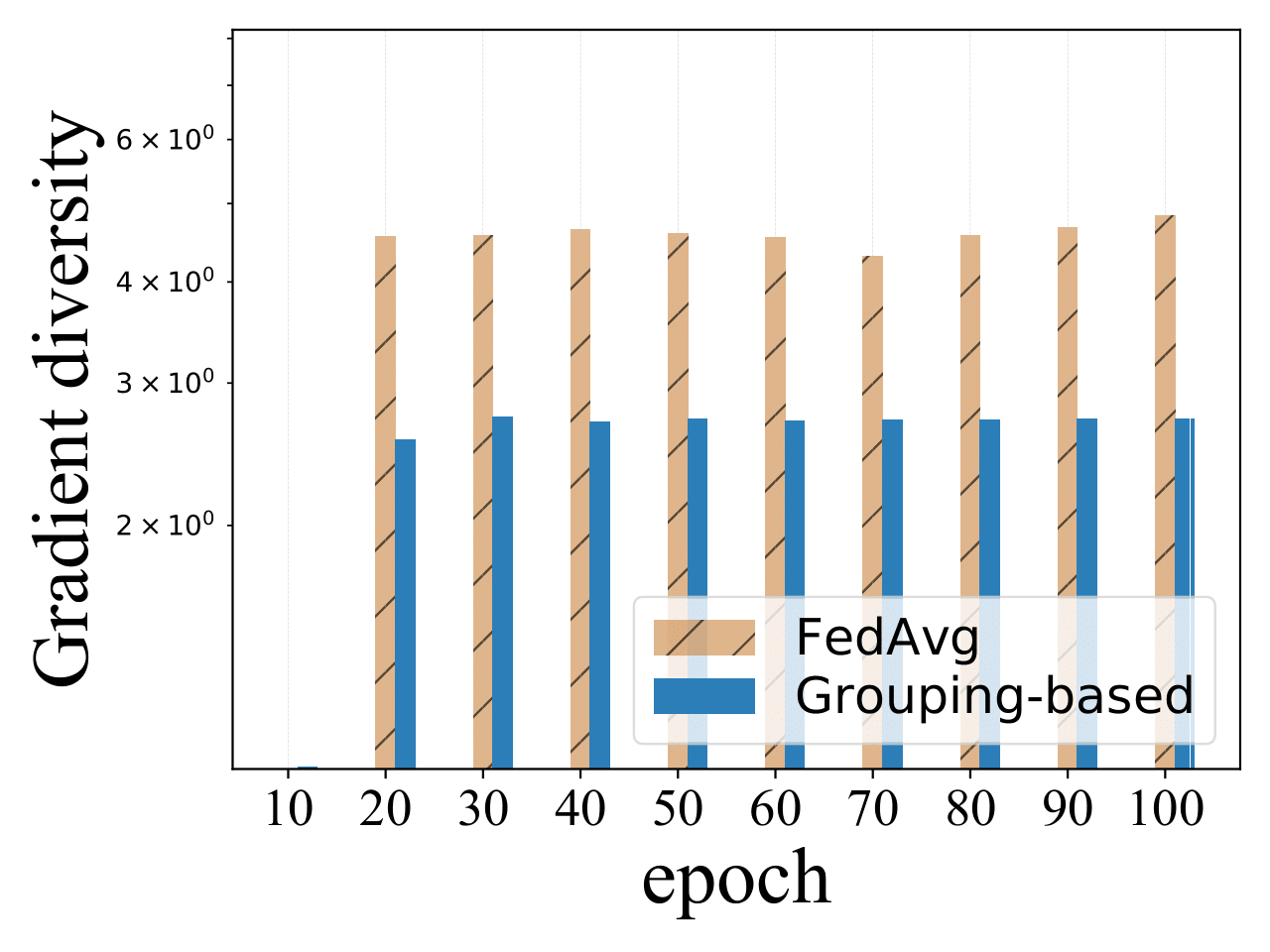}
    \includegraphics[width=0.32\linewidth]{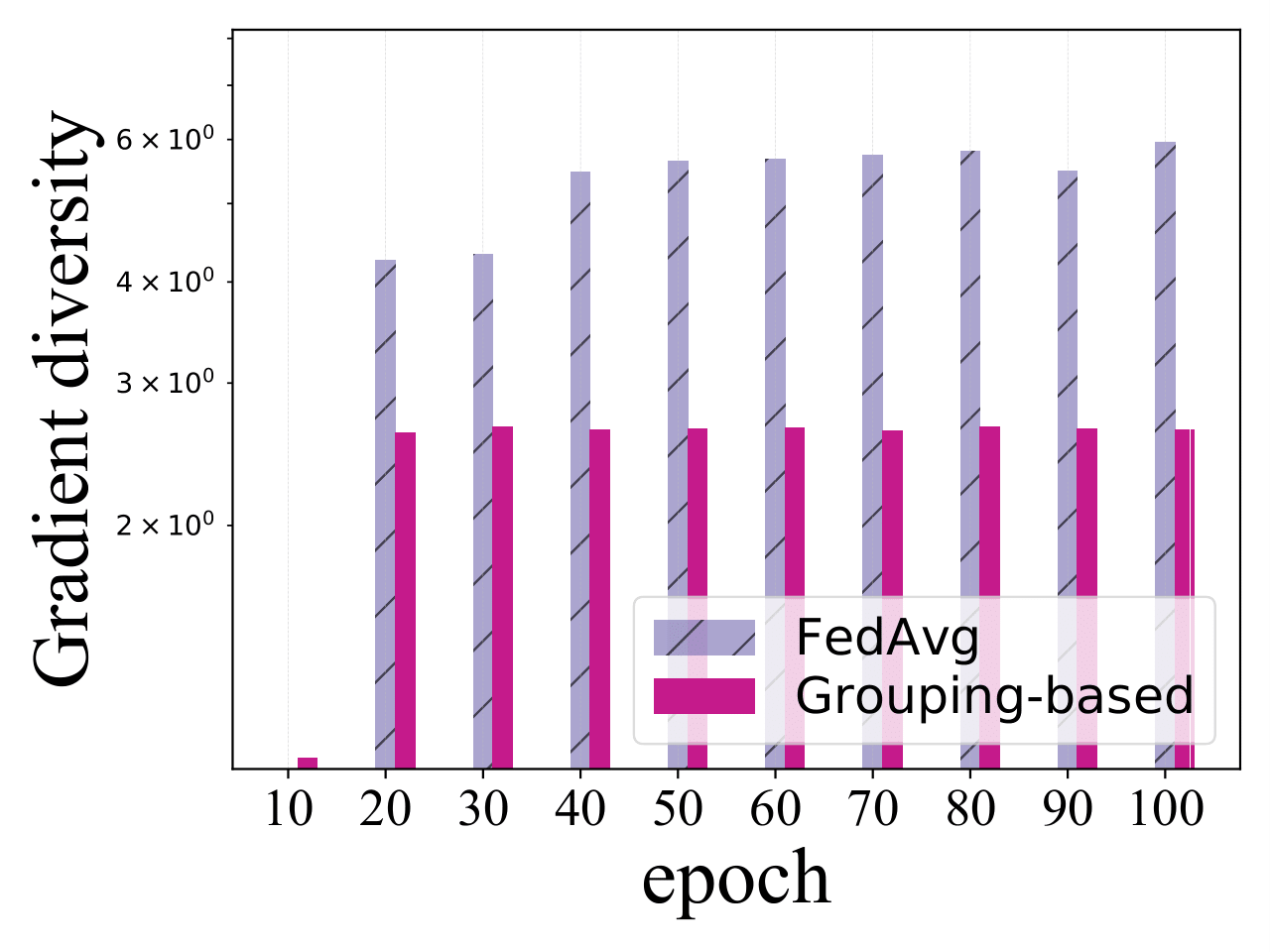}
    \caption{
    (Line 1) The convergence curves on EMNIST.
    (Line 2) Results on gradient diversity defined in \eref{graddiversityType3} with $w_k^t$ defined in \eref{CumulativeGradient}. 
    (Line 3) Results on gradient diversity defined in \eref{graddiversityType4} with $w_k^t$ defined in \eref{CumulativeGradient}. 
    (Line 4) Results on gradient diversity defined in \eref{graddiversityType3} with $w_k^t$ defined in \eref{CumulativeGradient}, including both the users and the server. 
    (Line 5) Results on gradient diversity defined in \eref{graddiversityType4} with $w_k^t$ defined in \eref{CumulativeGradient}, including both the users and the server. 
    }
    \label{fig:Groupingmethod_NormOrd1_with_FedGrad}
\end{figure*}

Therefore, we perform all the 16 different ways of calculating gradient diversity, and repeat the comparison between FedAvg and the grouping-based method under the same setting of the experiments on EMNIST shown in \sref{subs:FixCk}. The results are reported in \fref{fig:Groupingmethod} - \fref{fig:Groupingmethod_NormOrd1_with_FedGrad}.

We can see that the gradient diversity values of the grouping-based averaging method are consistently lower than FedAvg, and the corresponding test accuracy values are consistently higher than FedAvg.
More interestingly, we see that the grouping-based averaging method significantly accelerates the convergence speed compared to FedAvg. In other words, from~\fref{fig:Groupingmethod} - \fref{fig:Groupingmethod_NormOrd1_with_FedGrad}, a large gradient diversity value across different users can slow down the training process significantly.

\section{Impact of the ratio $\eta = C/K$}
\label{sxn:impact_c_over_k}

In this section, we study the impact of the user connection ratio because in real-life FL scenarios, the number of connected users can vary during training.
Here, we define the ratio of the connected users as $\eta=C/K$, and we study the impact of $\eta$ on Cifar-10 and SVHN.

We set $T=16$, $R=0.4$, and $N_s=1000$ for both Cifar-10 and SVHN. 
We use ResNet-18~\cite{ResNet} as the model for training, and we train 300 epochs and 40 epochs respectively on Cifar-10 and SVHN. The experiment parameters are shown in rows 30-31 of~\tref{tab:ExperimentParameters}.
The results of our grouping-based method on Cifar-10 and SVHN are shown in~\tref{tab:DifferentProportions}. 
It can be found that as $\eta$ increases from $\frac{1}{10}$ to $\frac{1}{3}$, the performance of our grouping-based method gradually improves. 
This result is intuitive since a smaller $\eta$ means fewer models participate in the model averaging, and only a small ratio of the models can be utilized. 
It should also be noted that, as shown in~\sref{subs:FixCk}, when $\eta$ increases to a large value (e.g. $\eta=1$), the diversity of models across users becomes too large, and the performance decreases.
Therefore, properly increasing $\eta$ can improve the performance but increasing $\eta$ too much can also decrease the performance. 
However, more work is needed to determine the most effective ratios for different datasets.

\begin{table*}
\newcommand{\tabincell}[2]{\begin{tabular}{@{}#1@{}}#2\end{tabular}}
\caption{
Test accuracy versus the ratio of communicating users $\eta=C/K$ on Cifar-10 and SVHN}
  \centering
  \begin{tabular}{lccccccccccccc}
\toprule
 & \tabincell{c}{$\eta=1/10$\\($K=30$, $C=3$)} & \tabincell{c}{$\eta=1/5$\\($K=30$, $C=6$)} & \tabincell{c}{$\eta=1/3$\\($K=30$, $C=10$)} &
 \tabincell{c}{$\eta=1$\\($K=30$, $C=30$)}\\
 \midrule
\hc  Cifar-10 (Grouping-based) & 82.48$\%$ & 92.08$\%$ & 92.84$\%$ & 92.12$\%$\\
  SVHN  (Grouping-based) & 92.83$\%$ & 93.42$\%$ & 93.56$\%$ & 78.77$\%$\\
\bottomrule
  \end{tabular}
  \label{tab:DifferentProportions}
\end{table*}

\section{Grouping-based averaging in fully supervised FL}
\label{sec:GroupingbasedinSFL}

\begin{table}
\caption{
The accuracy comparison of FedAvg and the grouping-based average for supervised FL on EMNSIT.
}
\label{tab:ablation_init} 
\centering
\begin{adjustbox}{width=0.45\textwidth} 
\begin{tabular}[t]{lcccccccccccccccccccccccccccc}
\toprule
User number & FedAvg  & Grouping-based &  \\
\midrule
   $K=47$  & 84.71$\%$ & \bf{84.97}$\%$ \\
\hc   $K=20$ & 86.14$\%$ & \bf{86.27}$\%$  \\
   $K=10$  &  86.19$\%$  &  \bf{86.29}$\%$\\
\bottomrule
\end{tabular}
\label{SFLComparison}
\end{adjustbox}
\end{table}

In this section, we study whether the grouping-based averaging can be extended to supervised FL (SFL). We want to see whether this particular way of averaging is more suitable for the semi-supervised setup or the supervised setup. 
We conduct experiments on EMNIST using SFL with three different settings with different number of users $K\in\{47,20,10\}$. In these three settings, we let $C=K$.
The other environmental factors are shown in rows 32-34 of~\tref{tab:ExperimentParameters}. 
We set the group number $S=5$, $S=2$ and $S=2$ for the setting of $K=47$, $K=20$ and $K=10$, respectively.
See Table~\ref{SFLComparison}.
The results show that the performance of the grouping method is only slightly better than that of FedAvg. 
Thus, the performance gain of the grouping-based averaging method for SFL is much less than that of \ssfl. This mean that grouping-based averaging is more suitable for the semi-supervised setup than the supervised setup.

\section{Comparison of grouping-based solution and a centralized scheme}
\label{sec:GroupingbasedandFixMatch}

In this section, we compare our grouping-based solution, which works in a distributed setting, to FixMatch~\cite{Fixmatch} which is originally proposed for the centralized semi-supervised setting. 
In particular, we compare our method in the \ssfl setting ($T=16$, $R=0.4$) and FixMatch (with centralized training on a single machine, $T=1$, $R=0$). 
For Cifar-10/SVHN/EMNIST, the environmental factors for both our grouping-based solution and FixMatch are shown in rows 35-40 of~\tref{tab:ExperimentParameters}. From~\tref{FixMatchComparison}, we see that the results are comparable to FixMatch even if FixMatch uses centralized training.

\begin{table}[H]
\caption{
Comparison with FixMatch.
}
\centering
\begin{tabular}[t]{lcccccccccccccccccc}
\toprule
Dataset  & FixMatch  & Grouping-based&  \\
\midrule
  Cifar10 & 95.74$\%$ & 92.86$\%$ \\
\hc  SVHN& 97.72$\%$ & 95.49$\%$  \\
  EMNSIT  &  83.70$\%$  &  81.63$\%$\\
\bottomrule
\end{tabular}
\label{FixMatchComparison}
\end{table}

\section{\ssfl for the label-at-client scenario}
\label{sec:UersSideSemi}

In this section, we discuss a different scenario of \ssfl when users have both unlabeled data and limited labeled data. 
Although in this paper, we have mainly focused on the case when labeled data is held by the server, we also think it is reasonable to assume the users have both unlabeled data and limited labeled data.
Thus, in this section, we study this ``label-at-client'' case.
Note that in this case, the server does not have any data, and only aggregates model weights.
This different scenario is reasonable when users have access to interactive applications to provide labels to the local data \cite{kairouz2019advances}.

All of our solutions discussed in the main paper apply directly to this different scenario. More specifically, we can simply apply the same server loss function \eref{eq:serverloss} to the labeled data at the users.

Now, we conduct an experiment on EMNIST to test the grouping-based solution, and the experiment results on Cifar-10 can been seen from~\tref{JeongCifar10}.
Here, we set $K=C=47$, $T=16$ and $N_s=4700$, which has the same environmental factors as reported in \tref{tab:CommunicationVolumeResults_on_emnist}
for the the grouping-based solution.
See \tref{UersSideSemi} for the results. From the results, we see that for $R=0.4$, the obtained accuracy $81.88\%$. This result indicates that our method can still apply to this alternative setting where users have both labeled and unlabeled data. 

\begin{table}
\caption{
Performance of the label-at-client setting.
}
\centering
\begin{tabular}[t]{lcccccccccccccccccc}
\toprule
      & $R=0.4$ & $R=0.6$ \\
    \midrule
    \hc  EMNSIT   &  81.88$\%$  &  81.40$\%$\\
    \bottomrule
\end{tabular}
\label{UersSideSemi}
\end{table}

\section{Performance on STL-10}
\label{sec:PerformanceSTL10}

In this section, we test all of our solutions on STL-10~\cite{coates2011analysis}, which is a dataset created specifically for semi-supervised learning. 
One thing worth noting is that we can only test the iid setting ($R=0$) on STL-10. This is because for the non-iid setting, we need to synthesize datasets with a particular non-iid level $R$, but STL-10 does not have full labeling information for us to synthesize the data distribution procedure (see \sref{subsec:DataDistribute}).
For STL-10, we set $K=10$, $C=2$, $T=16$ and $N_s=1000$. We use VGG11 as our model. The results are shown in~\tref{STL10}.
The self-training solution achieves 74.25\%.
The CRL with BN uses \eref{eq:Fixmatchuserloss} and \eref{eq:serverloss}, which is introduced in Section 3.1, achieves 78.96\%.
Then, when we change BN to GN, we achieve 81.71\%. 
When we further change FedAvg to grouping-based averaging, we achieve 82.81\%.

\begin{table}
\begin{minipage}[t]{1.0\linewidth}
\caption{
Performance on STL-10.
}
\centering
\begin{adjustbox}{width=\tabsize\textwidth,center} 
\begin{tabular}[t]{lcccccccccccccccccc}
\toprule
Self-training & CRL with BN & CRL with GN & Grouping-based\\
\midrule
\hc  74.25\%  &  78.96\% & 81.71\% &  82.81\% \\
      
\bottomrule
\end{tabular}
\label{STL10}
\end{adjustbox}
\end{minipage}\hfill
\end{table}

\section{Performance on EMNIST with a large user number}
\label{sec:EMNISTLargeK}

\begin{table}[H]
\caption{
The accuracy comparison of FedAvg and grouping-based average with large user number on EMNIST.
}
\centering
\begin{tabular}[t]{lcccccccccccccccccccccccccccc}
\toprule
 & FedAvg  & Grouping-based &  \\
\midrule
   $K=470, C=10$  & 83.69$\%$ & \bf{83.94}$\%$ \\
\hc    $K=470, C=20$ & 82.36$\%$ & \bf{83.66}$\%$  \\
   $K=470, C=30$  &  79.41$\%$  &  \bf{81.31}$\%$\\
\bottomrule
\end{tabular}
\label{EMNSITLargeUserNumber}
\end{table}

In this section, we study the performance of the FedAvg solution and the grouping-based averaging with a large user number, to see if this particular way of averaging is still useful when the number of users is particularly large (which is closer to the practical scenario). 
We conduct experiments on EMNIST with $K=470$,  $C=\{10,20,30\}$ and the number of groups $S=2$. We notice that this result represents the case when we have a relatively small user connection ratio $\eta=C/K$.

The results in Table~\ref{EMNSITLargeUserNumber} show that the performance of the grouping-based averaging is better than that of FedAvg even with 470 users. One can also observe that as the number of communicating users $C$ increases, the performance of the FedAvg decreases, which is consistent with the experimental phenomenon observed in~\sref{subs:FixCk}.

\end{document}